\documentclass[11pt]{article}
\usepackage{times,latexsym}
\usepackage{url}
\usepackage[T1]{fontenc}
\usepackage{soul}

\usepackage[utf8]{inputenc}
\usepackage{microtype}
\usepackage{inconsolata}
\usepackage[most]{tcolorbox}
\usepackage{float}

\tcbset{
  enhanced,
  frame hidden,
  size=tight,
  left=1pt,
  right=1pt,
  top=1pt,
  bottom=1pt
}

\usepackage{graphicx}
\usepackage{booktabs}
\usepackage{multirow}
\usepackage{xcolor,colortbl}
\usepackage[dvipsnames]{xcolor}
\usepackage{subfigure}
\usepackage{makecell}
\usepackage{float}
\newcommand{\socio}[1]{\small \color{black}{ (+socio)}\color{black}}

\usepackage[final]{acl}

\title{Improving the Distributional Alignment of LLMs using Supervision}

\author{
  Gauri Kambhatla,
  Sanjana Gautam,
  Angela Zhang,
  Alex Liu, \\
  \textbf{Ravi Srinivasan},
  \textbf{Junyi Jessy Li},
  \textbf{Matthew Lease}
  \\
   \\
  The University of Texas at Austin ~~~
  \\
  \texttt{gkambhat@utexas.edu}
}

\begin{document}
\maketitle
\begin{abstract}
The ability to accurately align LLMs with diverse population groups on subjective questions would have great value. In this work, we show that adding simple supervision can more \textit{consistently} improve the alignment of LLM-generated distributions with diverse population groups, as measured across three datasets spanning public health, public opinion, and values and beliefs. Beyond evaluating average alignment, we also report how alignment varies across specific groups. Our broad findings provide insights into the \textit{distributional} alignment of LLM generations with diverse populations. By conducting evaluation over many LLMs and prompting strategies, we provide a benchmark to stimulate future research.\footnote{Our data and code will be available at \url{https://github.com/GauriKambhatla/supervised-llm-alignment}}
\end{abstract}

\section{Introduction}

As LLMs are increasingly incorporated into real-world systems, it is more and more important to evaluate their alignment with human values and opinions. A growing body of work seeks to align LLMs with particular human population groups, with the goal of simulating their opinions or behaviors. This could be for applications such as psychology and economic experiments, piloting social science surveys (as part of AI for science), or seeking subjective data annotations \cite{pmlr-v202-aher23a, DILLION2023597, sun2024randomsiliconsamplingsimulating, Argyle_Busby_Fulda_Gubler_Rytting_Wingate_2023, JANSEN2023100020, hayati2024far, cegin-etal-2023-chatgpt}. One common method to do so is \textit{persona prompting}, which refers to defining a persona with demographic (\textit{sociodemographic prompting}) or description variables within a prompt, with the goal of modeling the behavior of that group using an LLM \cite{hu-collier-2024-quantifying, beck-etal-2024-sensitivity}. 

\begin{figure}[h]
    \centering
    \includegraphics[width=0.8\linewidth]{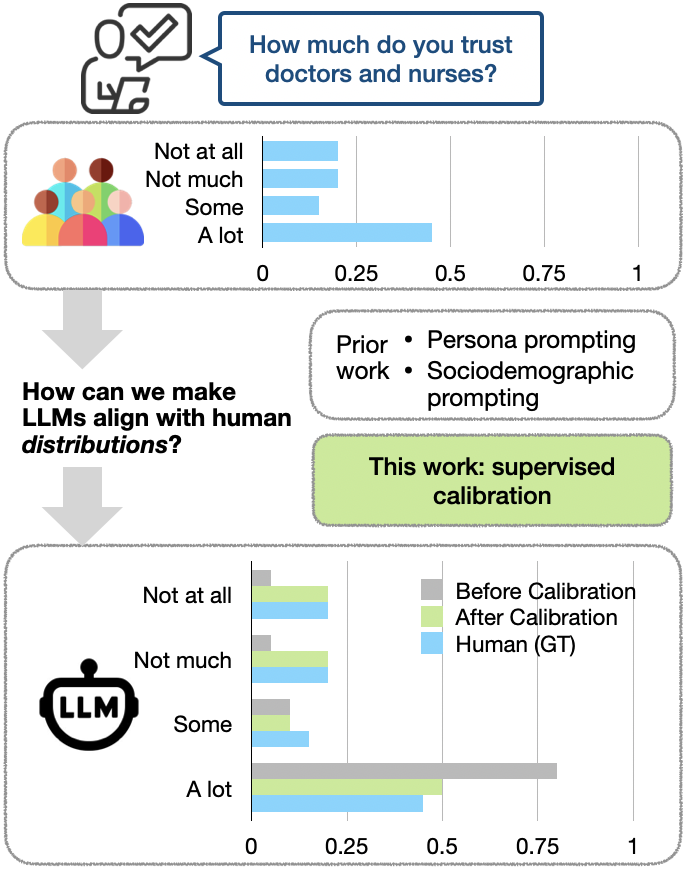}
    \vspace{-0.6em}
    \caption{Prior work studies using persona/sociodemographic prompting to align LLMs with humans for subjective questions.
    In this work, we elicit \textit{distributions} from LLMs and \textit{calibrate} them to better align with human response distributions.}
    \vspace{-1em}
    \label{fig:description_fig}
\end{figure}

Recent work in NLP has sought to evaluate sociodemographic (SD) prompting by studying its performance on large subjective surveys with thousands of human responses \cite{santurkar2023whose, durmus2024towards, masoud-etal-2025-cultural, alkhamissi-etal-2024-investigating}. However, evaluation of SD prompting has typically suffered from a key shortcoming: many studies assume a single majority answer to each question per sociodemographic group in modeling and/or evaluation \cite{hwang-etal-2023-aligning, hu-collier-2024-quantifying, sun2023aligning, mukherjee-etal-2024-cultural, durmus2024towards, masoud-etal-2025-cultural, alkhamissi-etal-2024-investigating}. The fallacy in such a framing is clear: because members of a given group do not all share the same beliefs, accurate modeling and evaluation must incorporate intra-group disagreement. Prompting for a distribution, rather than a representative response, could potentially also make a model less susceptible to caricature and stereotyping \cite{cheng-etal-2023-marked, Wang2025-cx, cheng-etal-2023-compost}, better aligning with the goals of pluralistic alignment \cite{SorensenMFGMRYJ24}. Prior studies that do model distributional beliefs tend to evaluate only a single method of extracting an LLM distribution \cite{santurkar2023whose, sun2024randomsiliconsamplingsimulating}, or evaluate LLM distributions 
in an SD-agnostic manner \cite{beck-etal-2024-sensitivity}. 

In this work, we evaluate SD prompting in a \textit{distributional} manner across three survey datasets using a variety of methods to extract distributions from LLMs. We focus on methods that broadly support use with open-source, open-model, and API-access LLMs. We apply \textit{supervised calibration} (described in Section \ref{sec:calibration}) to better align (see Section \ref{sec:metric}) LLM-generated distributions with human response distributions, with the intuition that LLM distributions might be directionally correct, but simply \textit{uncalibrated}; e.g., LLM distributions might exaggerate differences between groups. Our method calibrates output distributions for collective opinion estimation, as opposed to internal weight alignment. 

Concretely, we explore distributional opinion alignment via four {\bf research questions}: \textbf{(RQ 1)} Does SD prompting generate distributions that are more aligned with human opinion? \textbf{(RQ 2)} Can LLM-generated distributions be calibrated to better align with humans using supervised data? \textbf{(RQ 3)} Can LLM distributions be more easily aligned with some SD groups? \textbf{(RQ 4)} How many supervised training examples does calibration need? 

We evaluate the alignment of LLM-generated distributions to human opinion across 
3 distribution elicitation methods and 15 models of varying degrees of openness (open-source, open-model, API-access only), size, pre-training data, and post-training methods. We first show that baseline SD prompting techniques do not lead to consistent improvements in alignment across settings (RQ1). In contrast, our approach to supervised calibration of LLM-generated distributions improves alignment by 16.3\% on average across prompt methods, elicitation methods, datasets, and LLMs (RQ2), though it improves for some SD groups and degrades for others (RQ3). Moreover, as few as 1-10 gold examples can be used to calibrate distributions (RQ4). 

Overall, we find that applying calibration to LLM distributions reduces the importance of model and elicitation method choice by reliably improving alignment. However, we emphasize caution given that alignment success varies across SD groups and evaluating with SDs risks essentializing demographic categories.

\section{Related Work}

\paragraph{Social alignment of LLMs} 
As we seek to do in our work, many methods have been proposed to better {align} LLMs with human values and opinions \cite{ma-etal-2024-potential}. Persona prompting---defining a persona with description variables within a prompt to simulate or model groups---is one such method \cite{hu-collier-2024-quantifying}. One form, 
sociodemographic (SD) prompting, defines the persona with SD variables \cite{beck-etal-2024-sensitivity}. A variety of other methods for aligning LLMs includes: using knowledge of a user's previous opinions to model their future opinions \cite{hwang-etal-2023-aligning, gordon2022jury}, using psychological scaffolds to improve model rationalization \cite{joshi2025improving}, self-alignment via the simulation of social scenes \cite{pmlr-v235-pang24a}, extracting persona vectors to mitigate certain character traits \cite{chen2025personavectorsmonitoringcontrolling}, creating more descriptive value profiles \cite{sorensen-value-profiles-2025}, using in-context learning \cite{choenni2025selfalignmentimprovingalignmentcultural}, and fine-tuning LLMs \cite{orlikowski-etal-2025-beyond, krsteski2025validsurveysimulationslimited}. Evaluation frameworks have also been proposed for assessing alignment and social simulation \cite{kang2025deep, shi2025impersona, lutz2025promptmakespersonasystematic}.

\paragraph{Pitfalls of social alignment}
However, several works report inadequacies in these methods in accurately aligning models with humans, 
such as LLMs being unreliable as personalized annotators \cite{dong-etal-2024-llm}, unable to use different languages to steer generation towards groups \cite{kwok2024evaluating}, biased towards some groups \cite{sun-etal-2025-sociodemographic}, lacking stability and steerability under prompt variations \cite{khan-2025-randomness}, and lacking the reasoning depth to simulate humans in social experiments \cite{gao2025cautionusingllmshuman}.

Prior work evaluating SD prompting has also reported mixed results in
accurately simulating social groups \cite{beck-etal-2024-sensitivity, santurkar2023whose, hu-collier-2024-quantifying, kaiser-2025-simulating, zheng-etal-2024-helpful}. While some works show that SD prompting can improve performance, results vary greatly by prompt, model, and task \cite{beck-etal-2024-sensitivity, mukherjee-etal-2024-cultural, Bisbee_Clinton_Dorff_Kenkel_Larson_2024}. 
Other work warns against risks of  
misportrayal, othering, and exoticization of identities \cite{cheng-etal-2023-marked, Wang2025-cx, cheng-etal-2023-compost, kitadai-2024-examining} and have shown biases in language models that may be perpetuated with SD prompting \cite{nadeem-etal-2021-stereoset, gallegos2024bias, gupta2023bias}. 

\paragraph{Distributional alignment}
Our work follows a body of work that has advocated for a shift to pluralistic alignment due to the ecological fallacy of assuming all members of a group share the same values, beliefs, or opinions \cite{gordon2022jury, SorensenMFGMRYJ24, feng-etal-2024-modular, kirk2024the, xie2025a}. Similar to our work, prior works have studied aligning LLM \textit{distributions} with human response distributions on subjective questions. However, some of this work focuses on the systematic biases with LLM-generated distributions \cite{NEURIPS2024_515c6280} or on fine-tuning approaches \cite{cao-etal-2025-specializing}. \citet{santurkar2023whose} evaluate alignment with human opinion via a large US opinion poll. We draw inspiration from this work, though we evaluate multiple methods of extracting distributions from LLMs across multiple datasets. Closest to our work, \citet{meister-etal-2025-benchmarking} also benchmarks the distributional alignment of LLMs with various distribution elicitation methods and several datasets. However, we introduce supervised calibration as a method for improving distributional alignment.

\section{Approach}

\subsection{Datasets}
\label{sec:datasets}

We consider three survey datasets: Wellcome Global Monitor 2018 (WGM) \cite{Wellcome-Trust2019-dm}, 
OpinionQA (OQA) \cite{santurkar2023whose}, and the World Values Survey (WVS) \cite{Haerpfer2020-do}. OQA and WVS have been used to study alignment in prior work \cite{durmus2024towards, masoud-etal-2025-cultural, alkhamissi-etal-2024-investigating}. These three datasets cover diverse subjective topics, including global perceptions of science and public health (WGM), US public opinion (OQA), and various moral opinions and values (WVS). We include all WGM ordinal questions and two questions per category from OQA and WVS, yielding 92 questions across WGM (14), OQA (38), and WVS (40) and a total of 4,500 human response distributions over all datasets, questions, and demographics.

Table \ref{tab:example_dataset_questions} shows example questions and demographics for each dataset. For each question, model, and elicitation method (Section \ref{sec:distribution_elicitation_methods}), we use an LLM to generate distributions over answer choices, yielding 220,500 SD-specific response distributions and 4,500 SD-agnostic response distributions.

\subsection{Distribution elicitation}
\label{sec:distribution_elicitation_methods}

From ground truth human response data, we infer reference probability distributions over answer choices for each question by relative frequency, specific to each SD group. To generate distributions via LLM, we apply methods inspired from model confidence literature. 
We investigate techniques that work most broadly, supporting use with open-source, open-model, and API-access LLMs. That said, Appendix \ref{sec:appendix_log_probs} reports a smaller scale study with log probabilities for comparison. Appendix \ref{sec:prompts} shows prompts used for all methods. 

\textbf{Verbalized} directly prompts the model to output a probability distribution \cite{geng-etal-2024-survey, tian-etal-2023-just}. For example, given the question \textit{How much do you trust vaccines?}, with answer choices \textit{a lot, somewhat, not much, and not at all}, we prompt the model to output a distribution over all answer choices, such as [0.7, 0.2, 0.05, 0.05]. 
We sample $n=3$ times and average. We re-normalize any improper distributions and discard invalid outputs.

\textbf{Self-random} prompts the model to output a single answer choice, samples $n=5$ times with temperature 0.7 \cite{xiong2024can}, then creates a distribution over the $n$ responses. For example, given the question \textit{How much do you trust vaccines?}, with answer choices \textit{a lot, somewhat, not much, and not at all}, we prompt the model to output a single number that correlates with an answer choice (e.g., 4 for \textit{a lot} or 1 for \textit{not at all}).  

\textbf{Paraphrase} prompts the model to output a single answer choice but uses different paraphrases of the prompt instead of sampling responses for the same prompt \cite{xiong2024can}. We use $n=5$ paraphrases for each prompt.

\subsection{Prompts}
\label{sec:prompts_approach}
We use two categories of prompts, with three variations. For each, we prompt with the 3 different elicitation methods. See Appendix \ref{sec:prompts} for the prompts. 

\textbf{Base}. Base prompts exclude any sociodemographic information, asking questions from the original dataset with formatting slightly changed for LLMs. See Appendix Table \ref{tab:prompts_base} for full prompts.

\textbf{Sociodemographic (SD)}. These use a form similar to \textit{``Imagine you are \{d\}. Question: ''}, where \textit{d} is a demographic value. The exact prompt varies with demographic and elicitation method. See Appendix Table \ref{tab:prompts_sd} for full prompts.

\textbf{Prompt variations}. We study three prompt variations, standard prompts, few shot prompts (Section \ref{sec:few_shot}), and chain of thought (CoT) prompts \cite{kojima-cot-2022, wei-cot-2022} (Section \ref{sec:cot}). See Appendix Table \ref{tab:prompts_variations} for the variations.

\subsection{Metric: Opinion Alignment}
\label{sec:metric}
We use the following opinion alignment metric from \citet{santurkar2023whose} to measure similarity between an elicited distribution ($D_1$) and ground truth distribution ($D_2$) for the set of questions $Q$: % for all results:
$$A(D_1, D_2; Q) = \frac{1}{|Q|}\sum_{q \in Q}{\frac{1 - WD(D_1(q), D_2(q))}{N - 1}}$$
where $WD$ refers to Wasserstein distance, or earth-mover's distance, and $N-1$ is a normalization factor. \citet{santurkar2023whose} note this metric accounts for the ordinal nature of the survey questions, as opposed to other distribution divergence metrics like Kullback-Leibler or Jensen-Shannon. The opinion alignment metric ranges in $[0,1]$, but we show the value as a percentage in our results.

\subsection{Models}
We evaluate 15 models ranging in openness (open-source, open-weight, black box), size, modality, and post-training. Model families include Claude, Llama, Mistral, OLMo-2, and Qwen. Our main results report on only the most powerful model in each family. Others are reported in Appendix \ref{sec:appendix_all_models}.

\begin{table*}[h]
    \centering
    \footnotesize
    \addtolength{\tabcolsep}{-0.3em}
\begin{tabular}{cc|rrrrrr|rrrrrr}
\toprule
& \multirow{2}{*}{Model} &  \multicolumn{6}{c}{Base prompt} & \multicolumn{6}{c}{Sociodemographic prompt} \\
& & $P$ & \!$P_C$ & $S$ & \!$S_C$ & $V$ & \!$V_C$ & $P$ & \!$P_C$ & $S$ & \!$S_C$ & $V$ & \!$V_C$ \\
\midrule
\multirow{5}{*}{\rotatebox[origin=c]{90}{WGM}} & Claude-3.5-v2 & 66.2 & \textbf{85.2} & 59.5 & \textbf{85.1} & \cellcolor{yellow!30}89.3 & 87.7 & 65.4 & \textbf{84.4} & 61.4 & \textbf{84.2} & 89.0 & \cellcolor{yellow!30}89.3\\
& Llama-3.2-90B & 68.1 & \textbf{84.6} & 73.0 & \textbf{86.2} & 84.8 & \textbf{89.0} & 70.6 & \textbf{86.0} & 67.6 & \textbf{86.1} & 85.0 & \cellcolor{yellow!30}\textbf{89.8} \\
& Mistral-large & 62.4 & \textbf{84.7} & 72.0 & \textbf{83.9} & \cellcolor{yellow!30}89.4 & 88.9 & 68.4 & \textbf{84.7} & 63.0 & \textbf{84.9} & 87.3 & 88.4 \\
& OLMo-2-7B-I  & 59.6 & \textbf{81.5} & 67.7 & \textbf{80.4} & 59.8 & \cellcolor{yellow!30}\textbf{85.1} & 62.3 & \textbf{82.5} & 64.5 & \textbf{82.7} & 69.8 & \textbf{84.6} \\
& Qwen-2.5-72B & 57.7 & \textbf{83.1} & 53.3 & \textbf{83.2} & 88.2 & 87.1 & 66.0 & \textbf{84.1} & 63.4 & \textbf{85.2} & 89.1 & \cellcolor{yellow!30}89.4 \\
\cline{2-14}
& Average & 62.8 & \textbf{83.8} & 65.1 & \textbf{83.8} & 82.3 & \textbf{87.6} & 66.5 & \textbf{84.3} & 64.0 & \textbf{84.6} & 84.0 & \cellcolor{cyan!30}\textbf{88.3}\\
\bottomrule
\multirow{5}{*}{\rotatebox[origin=c]{90}{OQA}} & Claude-3.5-v2 & 70.5 & \textbf{88.8} & 72.5 & \textbf{90.4} & 91.7 & 91.7 & 76.1 & \textbf{89.9} & 73.3 & \textbf{89.6} & \cellcolor{yellow!30}91.9 & 91.6 \\
& Llama-3.2-90B & 79.3 & \textbf{89.6} & 75.3 & \textbf{89.4} & 86.8 & 87.9 & 79.2 & \cellcolor{yellow!30}\textbf{90.1} & 76.1 & \textbf{90.0} & 83.4 & \textbf{85.9} \\
& Mistral-large & 79.5 & \cellcolor{yellow!30}\textbf{89.9} & 75.3 & \textbf{88.3} & 85.0 & 86.2 & 75.8 & \textbf{89.2} & 72.4 & \textbf{89.5} & 83.8 & 84.7 \\
& OLMo-2-7B-I & 72.6 & \cellcolor{yellow!30}\textbf{89.0} & 72.3 & \textbf{88.4} & 65.4 & \textbf{79.9} & 72.8 & \textbf{88.4} & 70.5 & \textbf{88.6} & 68.5 & \textbf{81.5} \\
& Qwen-2.5-72B & 73.9 & \textbf{89.7} & 67.0 & \textbf{88.8} & 88.4 & 87.9 & 74.4 & \cellcolor{yellow!30}\textbf{90.0} & 71.3 & \textbf{89.5} & 89.2 & 88.6 \\
\cline{2-14}
& Average & 75.2 & \textbf{89.4} & 72.5 & \textbf{89.1} & 83.5 & \textbf{86.7} & 75.7 & \cellcolor{cyan!30}\textbf{89.5} & 72.7 & \textbf{89.4} & 83.4 & \textbf{86.5} \\
\bottomrule
\multirow{5}{*}{\rotatebox[origin=c]{90}{WVS}} & Claude-3.5-v2 & 46.5 & \textbf{80.3} & 51.0 & \textbf{80.3} & 75.2 & \textbf{80.3} & 61.0 & \textbf{80.4} & 56.8 & \textbf{80.4} & 75.6 & \cellcolor{yellow!30}81.7 \\
& Llama-3.2-90B &61.6 & \textbf{79.9} & 59.1 & \textbf{80.3} & 64.6 & \textbf{80.3} & 62.1 & \textbf{80.8} & 59.5 & \textbf{81.5} & 67.7 & \cellcolor{yellow!30}\textbf{82.7}  \\
& Mistral-large &  48.8 & \textbf{80.3} & 44.0 & \textbf{82.2} & 72.8 & \textbf{80.3} & 54.3 & \textbf{80.4} & 51.5 & \textbf{80.3} & 76.6 & \cellcolor{yellow!30}\textbf{83.8} \\
& OLMo-2-7B-I & 75.3 & \textbf{80.3} & 74.9 & \textbf{80.3} & 86.6 & 86.5 & 58.3 & \textbf{79.2} & 60.0 & \textbf{77.2} & 86.0 & \cellcolor{yellow!30}\textbf{89.8} \\
& Qwen-2.5-72B &  39.6 & \textbf{82.0} & 42.4 & \textbf{82.3} & 74.0 & 77.9 & 49.1 & \cellcolor{yellow!30}\textbf{82.4} & 49.4 & \textbf{81.5} & 73.4 & \textbf{81.7} \\
\cline{2-14}
& Average & 54.4 & \textbf{80.6} & 54.3 & \textbf{81.1} & 74.6 & \textbf{81.1} & 57.0 & \textbf{80.6} & 55.4 & \textbf{80.2} & 75.9 & \cellcolor{cyan!30}\textbf{83.9} \\
\bottomrule
\multirow{5}{*}{\rotatebox[origin=c]{90}{Average}} & Claude-3.5-v2 & 61.1 & \textbf{84.8} & 61.0 & \textbf{85.3} & 85.4 & \textbf{86.6} & 67.5 & \textbf{84.9} & 63.8 & \textbf{84.7} & 85.5 & \cellcolor{cyan!30}\textbf{87.5} \\
& Llama-3.2-90B & 69.7 & \textbf{84.7} & 69.1 & \textbf{85.3} & 78.7 & \textbf{85.7} & 70.6 & \textbf{85.6} & 67.7 & \textbf{85.9} & 78.7 & \cellcolor{cyan!30}\textbf{86.1} \\
& Mistral-large & 63.6 & \textbf{85.0} & 63.8 & \textbf{84.8} & 82.4 & \textbf{85.1} & 66.2 & \textbf{84.8} & 62.3 & \textbf{84.9 }& 82.6 & \cellcolor{cyan!30}\textbf{85.6} \\
& OLMo-2-7B-I & 69.2 & \textbf{83.6} & 71.6 & \textbf{83.0} & 70.6 & \textbf{83.8} & 64.5 & \textbf{83.4} & 65.0 & \textbf{82.8} & 74.8 & \cellcolor{cyan!30}\textbf{85.3} \\
& Qwen-2.5-72B & 57.1 & \textbf{84.9} & 54.2 & \textbf{84.8} & 83.5 & \textbf{84.3} & 63.2 & \textbf{85.5} & 61.4 & \textbf{85.4} & 83.9 & \cellcolor{cyan!30}\textbf{86.6}\\
\cline{2-14}
& Average & 64.1 & \textbf{84.6} & 64.0 & \textbf{84.6} & 80.1 & \textbf{85.1} & 66.4 & \textbf{84.8} & 64.0 & \textbf{84.7} & 81.1 & \cellcolor{cyan!30}\textbf{86.2} \\
\bottomrule
\end{tabular}
\caption{Opinion alignment before and after calibration for each dataset, LLM, and elicitation method. Each pair of columns compares the base-generated or SD-generated distributions to the calibrated distributions ($C$) for each elicitation method: paraphrase ($P$), self-random ($S$), and verbalized ($V$). \textbf{Bolded} values are significant between each pair (calculated via paired t-test and Bonferroni correction, see Appendix \ref{sec:appendix_stat_sig}). The best performance for each model is highlighted in \tcboxmath[colback=yellow!30]{$yellow$}. The best averages across models and datasets are highlighted in \tcboxmath[colback=cyan!30]{$blue$}. Results for all LLMs are shown in Appendix Table \ref{tab:regression_full}. 
\textbf{Adding SD information does not consistently improve alignment, but calibration more consistently improves alignment on average and in most settings.}}
\label{tab:regression_all_datasets}
\vspace{-1em}
\end{table*}

\subsection{Calibration}
\label{sec:calibration}

We apply supervised \textit{regression} to transform LLM-generated distributions to more closely align them with the ground truth human distributions. For example, for the question, \textit{How much do you trust doctors and nurses? A lot, some, not much, or not at all?}, the LLM might generate a distribution such as $[0.6, 0.2, 0.2, 0]$, while the human response distribution might be $[0.25, 0.5, 0.1, 0.15]$. We seek to learn a transformation from the LLM-generated distributions to the human distributions that might be applied to all LLM-generated distributions. More formally, for a  regression model $R$ and LLM distribution $\mathbf{D}$ parameterized by the values for each answer choice $[D_a, D_b, D_c, ..., D_k]$ (where $k$ is the number of answer choices), we learn a regression such that each value is transformed using supervision from ground truth values for each answer choice $[G_a, G_b, G_c, ..., G_k]$. We then re-normalize the transformed distribution.

\textbf{Training examples.} The examples for the regression model are the distributions split by answer choices. For example, the LLM and human distributions given above, $[0.6, 0.2, 0.2, 0]$ and $[0.25, 0.5, 0.1, 0.15]$, would be split into supervised pairs of $(0.6, 0.25)$, $(0.2, 0.5)$, $(0.2, 0.1)$, and $(0, 0.15)$. We thus train $R$ on $(X, y)$ pairs of $(D_a, G_a), (D_b, G_b),...$ to learn transformed values $[D_{a'}, D_{b'}, D_{c'}, ..., D_{k'}]$ on held-out test questions. 

\textbf{Regression model.} We learn a regression model for each \textit{dataset-LLM-elicitation-prompt} setting (270 regression models for 3 datasets, 15 models, 3 elicitation methods, and 2 prompt categories). We perform grid search for regressor model selection and corresponding hyperparameters (see Appendix \ref{sec:appendix_hyperparameters} for details).
We use the scikit-learn \cite{Pedregosa2011-rk} implementation for each regressor.

\textbf{Training.} The total number of \textit{examples} for each regression model is $|C|$ x $|Q|$ x $|A|$, where $C$ is the class within each sociodemographic group (e.g., female and male for sex), $Q$ is the question (e.g., how much do you trust doctors?), and $A$ is the set of answer choices per question. These values differ for each setting (see Appendix Table \ref{tab:example_dataset_questions}). We will refer to \textit{full examples} as the \textit{full} distributions not split by answer choices (total of $|C|$ x $|Q|$).
We were unable to extract distributions for all questions from some of the smaller LLMs (see Appendix Table \ref{tab:num_extracted_distributions}), affecting the number of training examples.  We split the LLM distributions into train, development, and test sets by questions (80\%-10\%-10\%). All of our reported results use the test set.

For minimal supervision (Section \ref{sec:minimal_supervision}), we train on a random sample of $n$ \textit{full examples} ($n$ = [1, 5, 10, 50, 100, 200, 500, 1000, 1200]) from the training set. We repeat this procedure 10 times for different random splits and average. 

\textbf{Evaluation.} We optimize the regression models with Mean Squared Error (MSE). 
We re-normalize output values to sum to 1 (for a proper probability distribution) before calculating alignment with ground truth distributions.
For example, if the output of $R$ for the distribution $[0.6, 0.2, 0.2, 0]$ is $[0.3, 0.7, 0.1, 0.1]$, we would normalize these values to $[0.25, 0.59, 0.08, 0.08]$.

\section{Aligning LLMs with human opinions}
\label{sec:aligning_llms}

We compare two methods: adding SD information to prompts vs.\ calibrating LLM-generated distributions with human ground truth data for supervision. We evaluate the effectiveness of these two methods in aligning with opinion distributions of those SD groups across models, datasets, and probability elicitation methods. Our main results are shown in Table \ref{tab:regression_all_datasets}. The alignment values for all models are shown in Appendix Table \ref{tab:regression_full} for all the results in this section. All significance results are calculated via paired t-test with Bonferroni correction.

\subsection{Does SD prompting improve alignment?}
\label{sec:sd_vs_no_sd}

For RQ1, we compare SD vs.\ base (SD-agnostic) prompts to study the effect of adding SD information on opinion alignment. While prior work studies this question only for majority-voted responses \cite{hu-collier-2024-quantifying, sun2023aligning, mukherjee-etal-2024-cultural, masoud-etal-2025-cultural, alkhamissi-etal-2024-investigating}, we investigate the effect of adding SD information when comparing \textit{distributions}. We study whether this is consistent across distribution elicitation methods, models, and datasets.

For base and SD prompts, LLM distributions are generally most aligned with verbalized elicitation (Table \ref{tab:regression_all_datasets}), with exceptions (e.g., on the OQA dataset). As shown in the table, prompting with SD information often leads to comparable or even \textit{lower} opinion alignment than prompting without any SD information, although this is model, dataset, and elicitation method dependent.

\subsection{Can we calibrate LLM distributions?}
\label{sec:llm_calibration}
We have seen that prompting LLMs with SD information does not consistently improve opinion alignment with human responses. 
We next investigate for RQ2 whether  \textit{calibrating} these distributions can more accurately and consistently align them with human response distributions.

The results after regression, compared to pre-regression results, are shown in Table \ref{tab:regression_all_datasets}. Calibration increases opinion alignment in 94.8\% of dataset-LLM-elicitation method settings, improving by an average of 16.3\%, where we average the delta between uncalibrated and calibrated opinion alignment across dataset-LLM-elicitation method settings. We find that in the 5.2\% of settings where it does not increase opinion alignment, the original alignment is relatively high and the decrease in alignment with calibration is low: alignment in these settings decreases by an average of  4.1\%. These results suggest that LLM-generated distributions are somewhat uncalibrated, and simple supervised regression can improve alignment. 

\begin{figure}
    \vspace{-1em}
    \centering
    \includegraphics[width=1\linewidth]{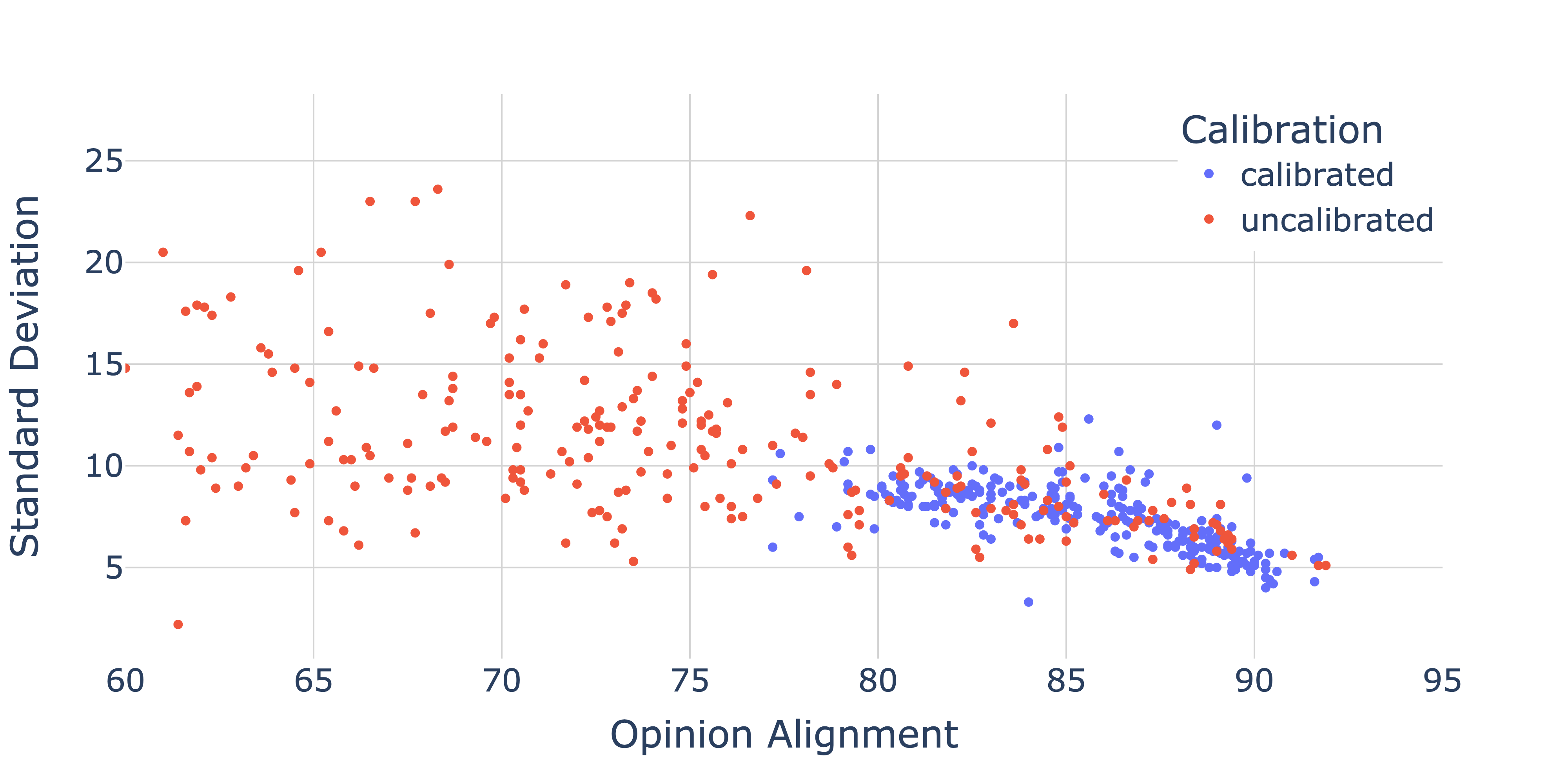}
    \caption{Standard deviation vs.\ opinion alignment. Each point represents the average alignment for each dataset, LLM, and elicitation method setting. For visual clarity, we omit 43/270 uncalibrated points having opinion alignment below 60.    \textbf{Calibration decreases variance in opinion alignment and standard deviation \textit{between} settings.}}
    \label{fig:std_wasserstein}
    \vspace{-1em}
\end{figure}

After calibration, opinion alignment in each setting also has much lower variance. Calibrated distributions have a lower standard deviation in 87.2\% of dataset-LLM-elicitation settings, and are 1.62 times lower on average. That lower standard deviation leads to higher opinion alignment provides further evidence that LLMs may exaggerate differences in opinion distributions between demographic groups \cite{cheng-etal-2023-marked, cheng-etal-2023-compost}, and that calibration could help mitigate this.

We also find that calibration reduces variance \textit{across} settings. Figure \ref{fig:std_wasserstein} plots standard deviation vs.\ opinion alignment, showing standard deviation across datasets, LLMs and elicitation methods before and after calibration.
Standard deviation is over 3 times smaller across all settings, and up to 5 times smaller per dataset. Although there are some settings where uncalibrated distributions have a lower standard deviation than their calibrated counterparts, we emphasize that the variance \textit{between} settings is much lower after calibration. While prior work \cite{beck-etal-2024-sensitivity, hu-collier-2024-quantifying} and our own results in Section \ref{sec:sd_vs_no_sd} showed that opinion alignment is dataset, LLM, and elicitation method dependent, these results indicate that calibration can improve the consistency between LLMs, datasets, and elicitation methods quite significantly. Differences between models become less pronounced following calibration. 

We do not find distinct patterns with model size within model family (e.g., Mistral small vs. Mistral large). Indeed, smaller models are sometimes more aligned than larger models, though across prompting methods, elicitation methods, and datasets, larger models are more aligned more consistently (see Appendix Table \ref{tab:regression_full}). However, we do find that chain of thought prompting (Appendix \ref{sec:cot}) appears to increase alignment for smaller models, but not for larger models.
The smaller Llama models in particular often refused to answer questions (i.e., questions would hit safety restrictions for attempting to role-play as a particular identity group), or would not follow instructions in formatting output distributions. We were thus unable to extract distributions for all questions, leading to higher variance in alignment. See Appendix Table \ref{tab:num_extracted_distributions} for percent distributions extracted in each setting.

\begin{table*}[h]
    \centering
    \footnotesize
    \addtolength{\tabcolsep}{-0.3em}
    \begin{tabular}{ll|llllllllll|ll}
    \toprule
    & \multirow{2}{*}{Demographic} & \multicolumn{2}{l}{Claude-3.5-v2} & \multicolumn{2}{l}{Llama-3.2-90B} & \multicolumn{2}{l}{Mistral-large} & \multicolumn{2}{l}{OLMo-2-7B-I} & \multicolumn{2}{l}{Qwen-2.5-72B} & \multicolumn{2}{|l}{\textbf{Average}} \\
 & & $V$ & $V_C$ & $V$ & $V_C$ & $V$ & $V_C$ & $V$ & $V_C$ & $V$ & $V_C$ & $V$ & $V_C$ \\
 \midrule

\multirow{18}{*}{\rotatebox[origin=c]{90}{World region (WGM)}} &
Aus/NZ & 82.3 & 79.7 & \textbf{94.7} & 83.5 & 88.2 & 86.6 & 73.5 & 79.1 & 89.1 & 84.2 & 85.6 & 82.6 \\
& Central Africa & 85.5 & 87.7 & 71.2 & 81.2 & 78.1 & 78.6 & 38.0 & \textbf{79.0} & 75.6 & 79.2 & 69.7 & \textbf{81.1} \\
& Cent. America \& Mex. & 89.2 & 90.0 & 74.8 & \textbf{85.1} & 81.7 & 82.8 & 46.4 & \textbf{84.7} & 78.5 & 81.8 & 74.1 & \textbf{84.9} \\
& Central Asia & 82.5 & 80.2 & 88.9 & 82.8 & 87.7 & 85.3 & 75.5 & 79.9 & 88.8 & 84.4 & 84.7 & 82.5 \\
& East Asia & 88.4 & 85.7 & 86.8 & 85.7 & 88.7 & 88.6 & 61.6 & 82.0 & 86.4 & 82.8 & 82.4 & 85.0 \\
& Eastern Africa & 88.2 & 88.4 & 80.0 & 87.7 & 88.0 & 87.3 & 63.3 & \textbf{87.7} & 86.0 & 86.2 & 81.1 & 87.5 \\
& Eastern Europe & 93.0 & 88.4 & 85.4 & 88.1 & 89.6 & 89.4 & 54.6 & 81.9 & 89.9 & 88.5 & 82.5 & 87.3 \\
& Middle East & 91.1 & 87.9 & 87.7 & 92.9 & 92.5 & 91.7 & 59.0 & 85.1 & 91.4 & 88.8 & 84.3 & 89.3 \\
& North Africa & 90.0 & 87.5 & 81.6 & \textbf{87.8} & 86.9 & 87.2 & 53.0 & 81.4 & 85.1 & 84.1 & 79.3 & 85.6 \\
& Northern America & 83.2 & 80.9 & 91.7 & 85.0 & 88.2 & 87.0 & 68.4 & 82.4 & 92.4 & 88.8 & 84.8 & 84.8 \\
& Northern Europe & 82.8 & 80.6 & \textbf{94.7} & 84.8 & 89.6 & 87.6 & 72.4 & 81.0 & 91.3 & 86.5 & 86.2 & 84.1 \\
& South America & 88.9 & 89.5 & 74.0 & \textbf{84.4} & 80.9 & 82.0 & 45.1 & \textbf{82.5} & 77.7 & 81.0 & 73.3 & \textbf{83.9} \\
& South Asia & 84.2 & 84.0 & 86.7 & 84.8 & 88.9 & 87.3 & 70.5 & 83.7 & 87.0 & 83.3 & 83.5 & 84.6 \\
& Southeast Asia & 82.1 & 82.5 & 79.1 & 80.2 & 86.3 & 84.6 & 70.6 & 81.9 & 84.2 & 81.7 & 80.5 & 82.2 \\
& Southern Africa & 86.4 & 90.1 & 74.2 & 84.4 & 82.4 & 82.8 & 49.1 & \textbf{88.8} & 79.7 & 83.3 & 74.4 & \textbf{85.9} \\
& Southern Europe & 93.2 & 89.1 & 84.5 & 88.7 & 90.0 & 89.7 & 53.4 & 81.7 & 88.5 & 88.2 & 81.9 & 87.5 \\
& Western Africa & 90.7 & 90.4 & 81.0 & \textbf{91.6} & 88.1 & 88.2 & 57.7 & \textbf{90.3} & 85.7 & 86.4 & 80.6 & \textbf{89.4} \\
& Western Europe & 83.2 & 80.6 & 93.0 & 86.4 & 90.3 & 88.5 & 71.6 & 84.2 & 93.5 & 88.7 & 86.3 & 85.7 \\
\midrule
\multicolumn{2}{l|}{All Demographics (WGM)} & 89.3 & 87.7 & 84.8 & \textbf{89.0} & 89.4 & 88.9 & 59.8 & \textbf{85.1} & 88.2 & 87.1 & 82.3 & 87.5\\
\midrule
\multirow{5}{*}{\rotatebox[origin=c]{90}{P.I.(OQA)}} &
Very conservative & 85.1 & 84.8 & 77.4 & 79.0 & 76.1 & 77.9 & 63.8 & 77.5 & 81.7 & 81.0 & 76.8 & 80.0 \\
& Conservative & 89.0 & 89.0 & 82.4 & 83.7 & 81.3 & 82.9 & 64.4 & \textbf{79.2} & 85.7 & 85.1 & 80.6 & 84.0 \\
& Moderate & 92.3 & 92.3 & 87.9 & 88.9 & 85.7 & 87.0 & 65.0 & \textbf{79.5} & 88.5 & 88.0 & 83.9 & 87.1 \\
& Liberal & 91.6 & 92.1 & 89.5 & 89.3 & 85.5 & 85.9 & 63.1 & \textbf{77.8} & 88.6 & 88.9 & 83.7 & 86.8 \\
& Very liberal & 87.8 & 88.3 & 86.5 & 85.9 & 83.3 & 83.3 & 61.1 & \textbf{76.1} & 84.8 & 85.0 & 80.7 & 83.7 \\
\midrule
\multicolumn{2}{l|}{All Demographics (OQA)} & 91.7 & 91.7 & 86.8 & 87.9 & 85.0 & 86.2 & 65.4 & \textbf{79.9} & 88.4 & 87.9 & 83.5 & 86.7\\
\midrule
\multirow{3}{*}{\rotatebox[origin=c]{90}{I.(WVS)}} &
High & 77.1 & 82.2 & 66.0 & 82.2 & 74.3 & 82.2 & 87.6 & 89.9 & 75.9 & 79.8 & 76.2 & 83.3 \\
& Middle & 75.2 & 80.2 & 64.7 & 80.2 & 72.7 & 80.2 & 86.8 & 86.6 & 73.6 & 77.8 & 74.6 & 81.0 \\
& Low & 77.4 & 82.4 & 66.8 & 82.4 & 75.0 & 82.4 & 87.8 & 88.5 & 76.1 & 80.0 & 76.6 & \textbf{83.1} \\
\midrule
\multicolumn{2}{l|}{All Demographics (WVS)} & 75.2 & \textbf{80.3} & 64.6 & \textbf{80.3} & 72.8 & \textbf{80.3} & 86.6 & 86.5 & 73.9 & 77.9 & 74.6 & 81.1\\
 \bottomrule
    \end{tabular}
    \caption{ Opinion alignment before ($V$) and after ($V_C$) calibration for three demographic categories (one from each dataset) using base-prompted, verbalized elicitation. ``P.I'' is political ideology and ``I.'' is income. Significantly higher values between each pair of columns (calculated via a paired t-test and Bonferroni correction) are \textbf{bolded}. The two ``Average'' columns on the right are averages across models, and the ``All Demographics'' rows are averages across the total set of SDs per dataset. See Appendix \ref{sec:appendix_individual_sd_results} for all SDs. \textbf{Calibrated distributions are more aligned with some SDs over others. Some models are better aligned with some datasets, e.g., Claude with OQA SDs and OLMo with WVS SDs.}}
    \vspace{-1em}
    \label{tab:individual_sociodemographics}
\end{table*}

\section{How does alignment vary across SDs?}

So far, we have studied alignment with SD groups \textit{aggregated across groups}, in order to compare different models and distribution elicitation methods. However, a particular model or method might be more aligned with some sociodemographics over others. With RQ3, we seek to understand the differences at the more granular demographic level.

We focus on three demographic categories, one from each dataset: world region (WGM), political ideology (OQA), and income (WVS). We only look at distributions elicited with the verbalized method, as it leads to the most aligned distributions with most models across datasets. We use the base-prompted distributions, as we find SD-prompted distributions do not show consistent improvement over base-prompted distributions for individual sociodemographics. Opinion alignment for world region (WGM), political ideology (OQA), and income (WVS) is shown in Table \ref{tab:individual_sociodemographics}. We show results for the same 5 LLMs used in Section \ref{sec:aligning_llms}. See Appendix \ref{sec:appendix_individual_sd_results} for alignment for other demographics.

\subsection{The effect of calibration}

We use the regression models trained on all SD groups to study how such aggregate models might calibrate individual SDs. We note few significant differences as our sample size for each demographic is low (4-6 examples). However, calibrated distributions are more absolutely aligned than their uncalibrated counterparts for 73.57\% of demographics in WGM, 68.41\% in OQA, and 78.17\% in WVS, across models for base verbalized distributions. This includes all demographic categories, including those in Table \ref{tab:individual_sociodemographics} and Appendix \ref{sec:appendix_individual_sd_results}. The delta improvement across demographics is most significant for OLMo on the WGM and OQA datasets.

As expected, calibration improves alignment for some demographics and reduces alignment for others. We find that it tends to improve alignment of SD groups that were less aligned with base-prompted distributions, such as Central America/Mexico, South America, Southern Africa, and Western Africa for world region. Alignment decreases for very high aligned demographic groups, such as Aus/NZ and Northern Europe.

\subsection{Alignment variation across LLMs}
For world region, alignment varies by model: Claude is most aligned with Africa and South/Central America, while Mistral and Llama are more aligned with Europe, Asia, and the Middle East. Qwen is most aligned with North America and Western Europe pre-calibration despite more extensive Chinese pretraining. For political ideology, Claude is most aligned with \textit{all} ideologies, followed by Qwen. On income (and most demographics in the WVS dataset), OLMo is most aligned with all levels, surprising given its lower alignment on the demographics in other datasets. See Appendix \ref{sec:post_training} for further discussion of OLMo alignment and post-training effects.

Across sociodemographics, Claude-generated distributions tend to be more highly aligned with the OQA dataset. 
All models show lowest alignment with middle income populations, possibly because low and high income groups have less varied opinions on these questions, i.e., income may be less predictive for middle income respondents.

\section{Does supervision work with less data?}
\label{sec:minimal_supervision}

Our results presented thus far used 80\% of each dataset for training our supervised regression models to calibrate the LLM-generated distributions. We next investigate whether smaller amounts of supervision can suffice for opinion alignment (RQ4). 

As discussed in Section \ref{sec:calibration}, we evaluate calibrated opinion alignment with 1, 5, 10, 50, 100, 200, ..., \textit{full examples} for each dataset, and average over 10 random samples for each size. 
We find that as few as 1-10 \textit{full examples} can suffice to closely match the minimal Mean Squared Error (MSE) for any particular dataset-LLM-elicitation setting. Plotting MSE over training data size shows that MSE usually converges between 1-10 full examples, though this is model and dataset dependent (see Appendix \ref{sec:appendix_minimal_supervision_plots}). We find that in comparison with few-shot prompting (Appendix \ref{sec:few_shot}), 
alignment improvement is more consistent with calibration.

\subsection{The effect on individual SD groups}
Although we see convergence with fewer examples, does using a small set of random examples affect alignment of individual demographics? We calibrate with a set of 5 full examples (in the middle of the 1-10 range described above for MSE convergence). While \textit{average} degradation is near zero, individual demographics are affected differently. 

Table~\ref{tab:min_supervision_degradation} shows the five most affected SDs and degradation for base and SD-prompted distributions. SD prompting changes which SDs show the highest alignment differences between full data and 5-full-example samples, though no differences are statistically significant. Using SD information increases degradation for the most affected demographics but shifts impact from historically underrepresented groups (SE Asia, less than high school education, Black, Hispanic) to overrepresented ones (Europe, Aus/NZ, tertiary education).

\begin{table}[]
    \centering
    \footnotesize
    \addtolength{\tabcolsep}{-0.2em}
    \begin{tabular}{lllrrr}
        \toprule
         & Category & Demographic & $V_{F}$ & $V_{5}$ & $\Delta$ \\
         \midrule
        \multirow{5}{*}{\rotatebox[origin=c]{90}{Base}} &
        Region & Southeast Asia & 81.0 & 79.4 & 1.6 \\
        & Education & Less than HS & 83.1 & 81.6 & 1.5 \\
        & Race & Black & 86.6 & 85.1 & 1.5 \\
        & Race & Hispanic & 87.2 & 85.9 & 1.4 \\
        & Pol. Party & Democrat & 84.7 & 83.3 & 1.4 \\
        \midrule
        \multirow{5}{*}{\rotatebox[origin=c]{90}{SD}} &
        Region & Northern Europe & 87.1 & 84.6 & 2.4 \\
        & Region & Western Europe & 87.1 & 84.9 & 2.3 \\
        & Region & Aus/NZ & 84.3 & 82.1 & 2.2 \\
        & Education & Tertiary & 88.6 & 86.5 & 2.1 \\
        & Employment & Unemployed & 86.9 & 84.9 & 2.0 \\
         \bottomrule
    \end{tabular}
    \caption{Demographics with largest opinion alignment \textit{degradation} from calibration models trained on the full dataset ($V_F$) vs.\ only five examples ($V_5$). $\Delta$ values shown differences. Distributions are verbally elicited, with base-prompted shown on top and SD-prompted on bottom. \textbf{Fewer training examples does not lead to any statistically significant degradation. SD prompting changes the groups with most absolute degradation.}}
    \vspace{-1em}
    \label{tab:min_supervision_degradation}
\end{table}

\section{Discussion}
\vspace{-0.5em}
\label{sec:discussion}

\textbf{The effect of SD prompting}. Given much prior work with conflicting findings, our extensive experiments show convincingly that prompting with SD information alone fails to consistently yield more aligned distributions. Future work might explore eliciting distributions with implicit demographic information or past opinions \cite{hwang-etal-2023-aligning, do-etal-2025-aligning}, personalization to individual group members rather than entire groups \cite{gordon2022jury, hwang-etal-2023-aligning}, or with {multimodal} prompts to incorporate implicit information.

\textbf{LLM distributions are uncalibrated}. We find that LLM-generated distributions (produced in answer to survey questions) are \textit{uncalibrated}; simple regression allows us to scale distributions to be more aligned with human opinion in aggregate, though this necessarily is more aligned with some populations over others. We also find that calibration reduces variance, both across examples and across settings. Since calibrated distributions generally have higher performance, this suggests LLMs exaggerate differences between sociodemographic groups, which calibration might help mitigate. 

\textbf{Calibration on individual SD groups}. 
We do not find many significant differences between calibrated/uncalibrated distributions at the individual SD group level. However, we hypothesize that calibrating responses from \textit{each individual} SD would likely improve its alignment, though we leave this to future work. However, efficacy may vary by demographic; some demographic traits are more relevant to the response distributions of certain questions (e.g., urban vs. rural areas might affect gun rights questions more than marital status).

\textbf{The choice of LLM and distribution elicitation method matters less with calibration}. 
We find that distributional alignment is dataset, LLM, and elicitation method dependent, consistent with prior work evaluating majority responses \cite{beck-etal-2024-sensitivity}. However, verbally-elicited distributions are most aligned with human distributions in the most settings.
In general, we find that calibration improves alignment \textit{more reliably} across models and methods; i.e., {the choice of model and elicitation method matters less after calibration}.

\section{Conclusion}
\vspace{-0.5em}
We investigate LLM \textit{distribution} alignment with human responses to subjective large-scale surveys, both on average and across diverse population groups. We show that using simple supervision can improve alignment with population groups consistently across datasets, models, and distribution elicitation techniques, as well as increase consistency between settings. Our benchmark serves to help enable and stimulate future research.

\section*{Limitations}
\label{sec:limitations}

\paragraph{Demographic modeling} 
Modeling groups with respect to their demographics assuredly provides only a narrow, incomplete view of people, and demographics alone do not determine our opinions. That said, such models may usefully approximate distributions of group opinions, provided care is exercised. We note that our methods for improving alignment should not be viewed as a way to fully and correctly model human groups, but rather a step towards more accurately representing groups. In addition, we note that not all demographics are necessarily predictive for all questions. While we do not study the predictiveness of demographics per question in this work, it might be an interesting analysis for future work.

\paragraph{Calibration model} 
Our calibration model predicts each answer choice individually and then normalizes the predicted answers, ignoring any correlations among answer choices. This might introduce risks of over-smoothing, loss of distributional structure, and potential distortions. Interestingly, though we tried constrained optimization that enforced proper distributions across all choices simultaneously, learning answer choices individually performed better. Future work might further analyze the effects of this individual answer choice prediction on alignment. 

\paragraph{Prompting methods} We only use a single temperature setting for our generations; future work might further analyze the effects of temperature on the quality of LLM-generated distributions. For elicitation methods, though a larger $n$ might improve generated distributions, our choice of $n$ is reasonable given inference costs and following prior work \cite{xiong2024can}. We use $n=3$ for verbalized elicitation (as opposed to $n=5$) since we found little variance between runs. 

\paragraph{Logit-based elicitation methods} 
We focus on methods that broadly work with open-source, open-model, and API-access LLMs. That said, we report a smaller scale study with logit-based distribution elicitation with four Llama models (Appendix \ref{sec:appendix_log_probs}). We found that log probabilities with calibration yielded the most aligned distributions in 10/24 settings considered. Future work might look more into comparisons between logit-based and verbalized distribution elicitation.

\paragraph{Additional \& intersectional demographics} 
We only studied a subset of the demographics available in the three datasets we use, as well as only single demographics due to cost limitations. Future work might study the effects of supervised calibration on distributional alignment with additional and/or intersectional demographics \cite{crenshaw-1991-intersectional}.

\paragraph{Populations studied}
While the datasets we use in this work do not cover all cultures and regions, we use surveys which cover a wide range: the WGM surveys over 140 countries, and the WVS (wave 7) covers 66 countries, both of which translate questionnaires to local languages while conducting the surveys \cite{Wellcome-Trust2019-dm, Haerpfer2020-do}.

\paragraph{Surveys as ground truth}
In this work, we use large scale surveys as reference distributions to evaluate and calibrate the LLM-generated distributions. We acknowledge that surveys are not perfect data sources, and do not fully capture human groups and their opinions. Survey data might include issues such as sampling bias, question wording effects, social desirability bias, and temporal instability (discussed more below). 
Despite these limitations, we believe that the surveys we use in this work are a reasonable source of ground truth data, surveying thousands of people across various world regions, beliefs, values, and opinions. We follow a body of prior work in this space \cite{santurkar2023whose, meister-etal-2025-benchmarking, durmus2024towards, alkhamissi-etal-2024-investigating, masoud-etal-2025-cultural, ma-etal-2024-potential}.

\paragraph{Temporal drift}
The supervised data used for calibration and for evaluation comes from large surveys conducted on human populations. As this data becomes outdated, calibrated distributions might not correctly reflect the opinions, beliefs, and values of current population groups. LLM-generated distributions may need to be re-calibrated and re-evaluated with new waves of survey data. Future work might look into when such re-calibration and re-evaluation needs to be done.

\section*{Ethical Considerations}
In this work, we study how we might align LLMs to the survey response distributions of various social groups. While this can be valuable for creating LLMs that more accurately portray and reason about population groups, this can also be used for adversarial purposes, such as chat bots attempting to simulate different user groups, targeted advertising, or targeted misinformation. We do not intend our datasets, models, and methods to be used for such adversarial purposes. Even for the intended purpose of furthering research in alignment of LLMs to population groups, we emphasize careful consideration of use-case and using caution to avoid propagating bias and stereotypes.

We note that by prompting LLMs with SD information, and evaluating their distributions against humans with those SDs, we are \textit{essentializing} the SD to the identities of the survey respondents and their opinions \cite{Wang2025-cx}. Although SDs are \textit{not} at all the only factor that influences people's opinions, they certainly play a part \cite{sap-etal-2022-annotators, biester-etal-2022-analyzing, pei-jurgens-2023-annotator}. By seeking to infer \textit{distributions} of human responses rather than majority responses for each SD group, we seek to align LLMs with groups in a distributionally pluralistic manner, rather than assume all members of the group would answer similarly. 

\section*{Acknowledgments}
We thank Sooyong Lee for early contributions that did not make it into the ultimate paper. This research was supported in part by Cisco, Good Systems\footnote{\url{https://goodsystems.utexas.edu/}} (a UT Austin Grand Challenge dedicated to developing responsible AI technologies), the Creating Connections program, and a grant from Open Philanthropy. We thank the Texas Advanced Computing Center (TACC) for use of its superb infrastructure. The statements made herein are solely the opinions of the authors and do not reflect the views of the sponsoring agencies.

\bibliography{anthology,bibtex}

\begin{thebibliography}{66}
\expandafter\ifx\csname natexlab\endcsname\relax\def\natexlab#1{#1}\fi

\bibitem[{Aher et~al.(2023)Aher, Arriaga, and Kalai}]{pmlr-v202-aher23a}
Gati~V Aher, Rosa~I. Arriaga, and Adam~Tauman Kalai. 2023.
\newblock \href {https://proceedings.mlr.press/v202/aher23a.html} {Using large language models to simulate multiple humans and replicate human subject studies}.
\newblock In \emph{Proceedings of the 40th International Conference on Machine Learning}, volume 202 of \emph{Proceedings of Machine Learning Research}, pages 337--371. PMLR.

\bibitem[{AlKhamissi et~al.(2024)AlKhamissi, ElNokrashy, Alkhamissi, and Diab}]{alkhamissi-etal-2024-investigating}
Badr AlKhamissi, Muhammad ElNokrashy, Mai Alkhamissi, and Mona Diab. 2024.
\newblock \href {https://aclanthology.org/2024.acl-long.671} {Investigating cultural alignment of large language models}.
\newblock In \emph{Proceedings of the 62nd Annual Meeting of the Association for Computational Linguistics (Volume 1: Long Papers)}, pages 12404--12422, Bangkok, Thailand. Association for Computational Linguistics.

\bibitem[{Argyle et~al.(2023)Argyle, Busby, Fulda, Gubler, Rytting, and Wingate}]{Argyle_Busby_Fulda_Gubler_Rytting_Wingate_2023}
Lisa~P. Argyle, Ethan~C. Busby, Nancy Fulda, Joshua~R. Gubler, Christopher Rytting, and David Wingate. 2023.
\newblock \href {https://doi.org/10.1017/pan.2023.2} {Out of one, many: Using language models to simulate human samples}.
\newblock \emph{Political Analysis}, 31(3):337–351.

\bibitem[{Beck et~al.(2024)Beck, Schuff, Lauscher, and Gurevych}]{beck-etal-2024-sensitivity}
Tilman Beck, Hendrik Schuff, Anne Lauscher, and Iryna Gurevych. 2024.
\newblock \href {https://aclanthology.org/2024.eacl-long.159} {Sensitivity, performance, robustness: Deconstructing the effect of sociodemographic prompting}.
\newblock In \emph{Proceedings of the 18th Conference of the European Chapter of the Association for Computational Linguistics (Volume 1: Long Papers)}, pages 2589--2615, St. Julian{'}s, Malta. Association for Computational Linguistics.

\bibitem[{Biester et~al.(2022)Biester, Sharma, Kazemi, Deng, Wilson, and Mihalcea}]{biester-etal-2022-analyzing}
Laura Biester, Vanita Sharma, Ashkan Kazemi, Naihao Deng, Steven Wilson, and Rada Mihalcea. 2022.
\newblock \href {https://aclanthology.org/2022.nlperspectives-1.2} {Analyzing the effects of annotator gender across {NLP} tasks}.
\newblock In \emph{Proceedings of the 1st Workshop on Perspectivist Approaches to NLP @LREC2022}, pages 10--19, Marseille, France. European Language Resources Association.

\bibitem[{Bisbee et~al.(2024)Bisbee, Clinton, Dorff, Kenkel, and Larson}]{Bisbee_Clinton_Dorff_Kenkel_Larson_2024}
James Bisbee, Joshua~D. Clinton, Cassy Dorff, Brenton Kenkel, and Jennifer~M. Larson. 2024.
\newblock \href {https://doi.org/10.1017/pan.2024.5} {Synthetic replacements for human survey data? the perils of large language models}.
\newblock \emph{Political Analysis}, 32(4):401–416.

\bibitem[{Cao et~al.(2025)Cao, Liu, Arora, Augenstein, R{\"o}ttger, and Hershcovich}]{cao-etal-2025-specializing}
Yong Cao, Haijiang Liu, Arnav Arora, Isabelle Augenstein, Paul R{\"o}ttger, and Daniel Hershcovich. 2025.
\newblock \href {https://doi.org/10.18653/v1/2025.naacl-long.162} {Specializing large language models to simulate survey response distributions for global populations}.
\newblock In \emph{Proceedings of the 2025 Conference of the Nations of the Americas Chapter of the Association for Computational Linguistics: Human Language Technologies (Volume 1: Long Papers)}, pages 3141--3154, Albuquerque, New Mexico. Association for Computational Linguistics.

\bibitem[{Cegin et~al.(2023)Cegin, Simko, and Brusilovsky}]{cegin-etal-2023-chatgpt}
Jan Cegin, Jakub Simko, and Peter Brusilovsky. 2023.
\newblock \href {https://doi.org/10.18653/v1/2023.emnlp-main.117} {{C}hat{GPT} to replace crowdsourcing of paraphrases for intent classification: Higher diversity and comparable model robustness}.
\newblock In \emph{Proceedings of the 2023 Conference on Empirical Methods in Natural Language Processing}, pages 1889--1905, Singapore. Association for Computational Linguistics.

\bibitem[{Chen et~al.(2025)Chen, Arditi, Sleight, Evans, and Lindsey}]{chen2025personavectorsmonitoringcontrolling}
Runjin Chen, Andy Arditi, Henry Sleight, Owain Evans, and Jack Lindsey. 2025.
\newblock \href {http://arxiv.org/abs/2507.21509} {Persona vectors: Monitoring and controlling character traits in language models}.
\newblock \emph{arXiv [cs.CL]}.

\bibitem[{Cheng et~al.(2023{\natexlab{a}})Cheng, Durmus, and Jurafsky}]{cheng-etal-2023-marked}
Myra Cheng, Esin Durmus, and Dan Jurafsky. 2023{\natexlab{a}}.
\newblock \href {https://doi.org/10.18653/v1/2023.acl-long.84} {Marked personas: Using natural language prompts to measure stereotypes in language models}.
\newblock In \emph{Proceedings of the 61st Annual Meeting of the Association for Computational Linguistics (Volume 1: Long Papers)}, pages 1504--1532, Toronto, Canada. Association for Computational Linguistics.

\bibitem[{Cheng et~al.(2023{\natexlab{b}})Cheng, Piccardi, and Yang}]{cheng-etal-2023-compost}
Myra Cheng, Tiziano Piccardi, and Diyi Yang. 2023{\natexlab{b}}.
\newblock \href {https://doi.org/10.18653/v1/2023.emnlp-main.669} {{C}o{MP}os{T}: Characterizing and evaluating caricature in {LLM} simulations}.
\newblock In \emph{Proceedings of the 2023 Conference on Empirical Methods in Natural Language Processing}, pages 10853--10875, Singapore. Association for Computational Linguistics.

\bibitem[{Choenni and Shutova(2025)}]{choenni2025selfalignmentimprovingalignmentcultural}
Rochelle Choenni and Ekaterina Shutova. 2025.
\newblock \href {http://arxiv.org/abs/2408.16482} {Self-alignment: Improving alignment of cultural values in llms via in-context learning}.
\newblock \emph{arXiv [cs.CL]}.

\bibitem[{Crenshaw(1991)}]{crenshaw-1991-intersectional}
Kimberle Crenshaw. 1991.
\newblock \href {http://www.jstor.org/stable/1229039} {Mapping the margins: Intersectionality, identity politics, and violence against women of color}.
\newblock \emph{Stanford Law Review}, 43(6):1241--1299.

\bibitem[{Dillion et~al.(2023)Dillion, Tandon, Gu, and Gray}]{DILLION2023597}
Danica Dillion, Niket Tandon, Yuling Gu, and Kurt Gray. 2023.
\newblock \href {https://doi.org/https://doi.org/10.1016/j.tics.2023.04.008} {Can ai language models replace human participants?}
\newblock \emph{Trends in Cognitive Sciences}, 27(7):597--600.

\bibitem[{Do et~al.(2025)Do, Kawaguchi, Kan, and Chen}]{do-etal-2025-aligning}
Xuan~Long Do, Kenji Kawaguchi, Min-Yen Kan, and Nancy Chen. 2025.
\newblock \href {https://aclanthology.org/2025.coling-main.172/} {Aligning large language models with human opinions through persona selection and value{--}belief{--}norm reasoning}.
\newblock In \emph{Proceedings of the 31st International Conference on Computational Linguistics}, pages 2526--2547, Abu Dhabi, UAE. Association for Computational Linguistics.

\bibitem[{Dominguez-Olmedo et~al.(2024)Dominguez-Olmedo, Hardt, and Mendler-D\"{u}nner}]{NEURIPS2024_515c6280}
Ricardo Dominguez-Olmedo, Moritz Hardt, and Celestine Mendler-D\"{u}nner. 2024.
\newblock \href {https://proceedings.neurips.cc/paper_files/paper/2024/file/515c62809e0a29729d7eec26e2916fc0-Paper-Conference.pdf} {Questioning the survey responses of large language models}.
\newblock In \emph{Advances in Neural Information Processing Systems}, volume~37, pages 45850--45878. Curran Associates, Inc.

\bibitem[{Dong et~al.(2024)Dong, Hu, and Collier}]{dong-etal-2024-llm}
Yijiang~River Dong, Tiancheng Hu, and Nigel Collier. 2024.
\newblock \href {https://doi.org/10.18653/v1/2024.findings-emnlp.592} {Can {LLM} be a personalized judge?}
\newblock In \emph{Findings of the Association for Computational Linguistics: EMNLP 2024}, pages 10126--10141, Miami, Florida, USA. Association for Computational Linguistics.

\bibitem[{Durmus et~al.(2024)Durmus, Nguyen, Liao, Schiefer, Askell, Bakhtin, Chen, Hatfield-Dodds, Hernandez, Joseph, Lovitt, McCandlish, Sikder, Tamkin, Thamkul, Kaplan, Clark, and Ganguli}]{durmus2024towards}
Esin Durmus, Karina Nguyen, Thomas Liao, Nicholas Schiefer, Amanda Askell, Anton Bakhtin, Carol Chen, Zac Hatfield-Dodds, Danny Hernandez, Nicholas Joseph, Liane Lovitt, Sam McCandlish, Orowa Sikder, Alex Tamkin, Janel Thamkul, Jared Kaplan, Jack Clark, and Deep Ganguli. 2024.
\newblock \href {https://openreview.net/forum?id=zl16jLb91v} {Towards measuring the representation of subjective global opinions in language models}.
\newblock In \emph{First Conference on Language Modeling}.

\bibitem[{Feng et~al.(2024)Feng, Sorensen, Liu, Fisher, Park, Choi, and Tsvetkov}]{feng-etal-2024-modular}
Shangbin Feng, Taylor Sorensen, Yuhan Liu, Jillian Fisher, Chan~Young Park, Yejin Choi, and Yulia Tsvetkov. 2024.
\newblock \href {https://doi.org/10.18653/v1/2024.emnlp-main.240} {Modular pluralism: Pluralistic alignment via multi-{LLM} collaboration}.
\newblock In \emph{Proceedings of the 2024 Conference on Empirical Methods in Natural Language Processing}, pages 4151--4171, Miami, Florida, USA. Association for Computational Linguistics.

\bibitem[{Gallegos et~al.(2024)Gallegos, Rossi, Barrow, Tanjim, Kim, Dernoncourt, Yu, Zhang, and Ahmed}]{gallegos2024bias}
Isabel~O Gallegos, Ryan~A Rossi, Joe Barrow, Md~Mehrab Tanjim, Sungchul Kim, Franck Dernoncourt, Tong Yu, Ruiyi Zhang, and Nesreen~K Ahmed. 2024.
\newblock Bias and fairness in large language models: A survey.
\newblock \emph{Computational Linguistics}, pages 1--79.

\bibitem[{Gao et~al.(2025)Gao, Lee, Burtch, and Fazelpour}]{gao2025cautionusingllmshuman}
Yuan Gao, Dokyun Lee, Gordon Burtch, and Sina Fazelpour. 2025.
\newblock \href {http://arxiv.org/abs/2410.19599} {Take caution in using llms as human surrogates: Scylla ex machina}.
\newblock \emph{arXiv [cs.CL]}.

\bibitem[{Geng et~al.(2024)Geng, Cai, Wang, Koeppl, Nakov, and Gurevych}]{geng-etal-2024-survey}
Jiahui Geng, Fengyu Cai, Yuxia Wang, Heinz Koeppl, Preslav Nakov, and Iryna Gurevych. 2024.
\newblock \href {https://doi.org/10.18653/v1/2024.naacl-long.366} {A survey of confidence estimation and calibration in large language models}.
\newblock In \emph{Proceedings of the 2024 Conference of the North American Chapter of the Association for Computational Linguistics: Human Language Technologies (Volume 1: Long Papers)}, pages 6577--6595, Mexico City, Mexico. Association for Computational Linguistics.

\bibitem[{Gordon et~al.(2022)Gordon, Lam, Park, Patel, Hancock, Hashimoto, and Bernstein}]{gordon2022jury}
Mitchell~L Gordon, Michelle~S Lam, Joon~Sung Park, Kayur Patel, Jeff Hancock, Tatsunori Hashimoto, and Michael~S Bernstein. 2022.
\newblock Jury learning: Integrating dissenting voices into machine learning models.
\newblock In \emph{Proceedings of the 2022 CHI Conference on Human Factors in Computing Systems}, pages 1--19.

\bibitem[{Gupta et~al.(2024)Gupta, Shrivastava, Deshpande, Kalyan, Clark, Sabharwal, and Khot}]{gupta2023bias}
Shashank Gupta, Vaishnavi Shrivastava, Ameet Deshpande, Ashwin Kalyan, Peter Clark, Ashish Sabharwal, and Tushar Khot. 2024.
\newblock Bias runs deep: Implicit reasoning biases in persona-assigned llms.
\newblock In \emph{The Twelfth International Conference on Learning Representations (ICLR)}.

\bibitem[{Haerpfer et~al.(2020)Haerpfer, Inglehart, Moreno, Welzel, Kizilova, Diez-Medrano, Lagos, Norris, Ponarin, and Puranen}]{Haerpfer2020-do}
Christian Haerpfer, Ronald Inglehart, Alejandro Moreno, Christian Welzel, Kseniya Kizilova, Jaime Diez-Medrano, Marta Lagos, Pippa Norris, Eduard Ponarin, and Bi~Puranen. 2020.
\newblock World values survey wave 7 (2017-2020) cross-national data-set.

\bibitem[{Hayati et~al.(2024)Hayati, Lee, Rajagopal, and Kang}]{hayati2024far}
Shirley Hayati, Minhwa Lee, Dheeraj Rajagopal, and Dongyeop Kang. 2024.
\newblock How far can we extract diverse perspectives from large language models?
\newblock In \emph{Proceedings of the 2024 Conference on Empirical Methods in Natural Language Processing}, pages 5336--5366.

\bibitem[{Hu and Collier(2024)}]{hu-collier-2024-quantifying}
Tiancheng Hu and Nigel Collier. 2024.
\newblock \href {https://aclanthology.org/2024.acl-long.554} {Quantifying the persona effect in {LLM} simulations}.
\newblock In \emph{Proceedings of the 62nd Annual Meeting of the Association for Computational Linguistics (Volume 1: Long Papers)}, pages 10289--10307, Bangkok, Thailand. Association for Computational Linguistics.

\bibitem[{Hwang et~al.(2023)Hwang, Majumder, and Tandon}]{hwang-etal-2023-aligning}
EunJeong Hwang, Bodhisattwa Majumder, and Niket Tandon. 2023.
\newblock \href {https://doi.org/10.18653/v1/2023.findings-emnlp.393} {Aligning language models to user opinions}.
\newblock In \emph{Findings of the Association for Computational Linguistics: EMNLP 2023}, pages 5906--5919, Singapore. Association for Computational Linguistics.

\bibitem[{Jansen et~al.(2023)Jansen, gyo Jung, and Salminen}]{JANSEN2023100020}
Bernard~J. Jansen, Soon gyo Jung, and Joni Salminen. 2023.
\newblock \href {https://doi.org/https://doi.org/10.1016/j.nlp.2023.100020} {Employing large language models in survey research}.
\newblock \emph{Natural Language Processing Journal}, 4:100020.

\bibitem[{Joshi et~al.(2025)Joshi, Ren, Swayamdipta, Koncel-Kedziorski, and Paek}]{joshi2025improving}
Brihi Joshi, Xiang Ren, Swabha Swayamdipta, Rik Koncel-Kedziorski, and Tim Paek. 2025.
\newblock Improving llm personas via rationalization with psychological scaffolds.
\newblock \emph{arXiv preprint arXiv:2504.17993}.

\bibitem[{Kaiser et~al.(2025)Kaiser, Kaiser, Manewitsch, Rau, and Schallner}]{kaiser-2025-simulating}
Carolin Kaiser, Jakob Kaiser, Vladimir Manewitsch, Lea Rau, and Rene Schallner. 2025.
\newblock \href {https://doi.org/10.1145/3708319.3733685} {Simulating human opinions with large language models: Opportunities and challenges for personalized survey data modeling}.
\newblock In \emph{Adjunct Proceedings of the 33rd ACM Conference on User Modeling, Adaptation and Personalization}, UMAP Adjunct '25, page 82–86, New York, NY, USA. Association for Computing Machinery.

\bibitem[{Kang et~al.(2025)Kang, Moon, Lee, Raj, Suh, and Chan}]{kang2025deep}
Minwoo Kang, Suhong Moon, Seung~Hyeong Lee, Ayush Raj, Joseph Suh, and David Chan. 2025.
\newblock \href {https://openreview.net/forum?id=zHdSCtNmM4} {Deep binding of language model virtual personas: a study on approximating political partisan misperceptions}.
\newblock In \emph{Second Conference on Language Modeling}.

\bibitem[{Khan et~al.(2025)Khan, Casper, and Hadfield-Menell}]{khan-2025-randomness}
Ariba Khan, Stephen Casper, and Dylan Hadfield-Menell. 2025.
\newblock \href {https://doi.org/10.1145/3715275.3732147} {Randomness, not representation: The unreliability of evaluating cultural alignment in llms}.
\newblock In \emph{Proceedings of the 2025 ACM Conference on Fairness, Accountability, and Transparency}, FAccT '25, page 2151–2165, New York, NY, USA. Association for Computing Machinery.

\bibitem[{Kirk et~al.(2024)Kirk, Whitefield, R{\"o}ttger, Bean, Margatina, Mosquera, Ciro, Bartolo, Williams, He, Vidgen, and Hale}]{kirk2024the}
Hannah~Rose Kirk, Alexander Whitefield, Paul R{\"o}ttger, Andrew~Michael Bean, Katerina Margatina, Rafael Mosquera, Juan~Manuel Ciro, Max Bartolo, Adina Williams, He~He, Bertie Vidgen, and Scott~A. Hale. 2024.
\newblock \href {https://openreview.net/forum?id=DFr5hteojx} {The {PRISM} alignment dataset: What participatory, representative and individualised human feedback reveals about the subjective and multicultural alignment of large language models}.
\newblock In \emph{The Thirty-eight Conference on Neural Information Processing Systems Datasets and Benchmarks Track}.

\bibitem[{Kitadai et~al.(2024)Kitadai, Ogawa, and Nishino}]{kitadai-2024-examining}
Ayato Kitadai, Kazuhito Ogawa, and Nariaki Nishino. 2024.
\newblock \href {https://doi.org/10.1109/BigData62323.2024.10825497} {Examining the feasibility of large language models as survey respondents}.
\newblock In \emph{2024 IEEE International Conference on Big Data (BigData)}, pages 3858--3864.

\bibitem[{Kojima et~al.(2022)Kojima, Gu, Reid, Matsuo, and Iwasawa}]{kojima-cot-2022}
Takeshi Kojima, Shixiang~Shane Gu, Machel Reid, Yutaka Matsuo, and Yusuke Iwasawa. 2022.
\newblock Large language models are zero-shot reasoners.
\newblock In \emph{Proceedings of the 36th International Conference on Neural Information Processing Systems}, NIPS '22, Red Hook, NY, USA. Curran Associates Inc.

\bibitem[{Krsteski et~al.(2025)Krsteski, Russo, Chang, West, and Gligorić}]{krsteski2025validsurveysimulationslimited}
Stefan Krsteski, Giuseppe Russo, Serina Chang, Robert West, and Kristina Gligorić. 2025.
\newblock \href {http://arxiv.org/abs/2510.11408} {Valid survey simulations with limited human data: The roles of prompting, fine-tuning, and rectification}.
\newblock \emph{arXiv [cs.CL]}.

\bibitem[{Kwok et~al.(2024)Kwok, Bravansky, and Griffin}]{kwok2024evaluating}
Louis Kwok, Michal Bravansky, and Lewis Griffin. 2024.
\newblock \href {https://openreview.net/forum?id=S4ZOkV1AHl} {Evaluating cultural adaptability of a large language model via simulation of synthetic personas}.
\newblock In \emph{First Conference on Language Modeling}.

\bibitem[{Lutz et~al.(2025)Lutz, Sen, Ahnert, Rogers, and Strohmaier}]{lutz2025promptmakespersonasystematic}
Marlene Lutz, Indira Sen, Georg Ahnert, Elisa Rogers, and Markus Strohmaier. 2025.
\newblock \href {http://arxiv.org/abs/2507.16076} {The prompt makes the person(a): A systematic evaluation of sociodemographic persona prompting for large language models}.

\bibitem[{Ma et~al.(2024)Ma, Wang, Hu, Haensch, Hedderich, Plank, and Kreuter}]{ma-etal-2024-potential}
Bolei Ma, Xinpeng Wang, Tiancheng Hu, Anna-Carolina Haensch, Michael~A. Hedderich, Barbara Plank, and Frauke Kreuter. 2024.
\newblock \href {https://doi.org/10.18653/v1/2024.findings-emnlp.513} {The potential and challenges of evaluating attitudes, opinions, and values in large language models}.
\newblock In \emph{Findings of the Association for Computational Linguistics: EMNLP 2024}, pages 8783--8805, Miami, Florida, USA. Association for Computational Linguistics.

\bibitem[{Masoud et~al.(2025)Masoud, Liu, Ferianc, Treleaven, and Rodrigues}]{masoud-etal-2025-cultural}
Reem Masoud, Ziquan Liu, Martin Ferianc, Philip~C. Treleaven, and Miguel~Rodrigues Rodrigues. 2025.
\newblock \href {https://aclanthology.org/2025.coling-main.567/} {Cultural alignment in large language models: An explanatory analysis based on hofstede{'}s cultural dimensions}.
\newblock In \emph{Proceedings of the 31st International Conference on Computational Linguistics}, pages 8474--8503, Abu Dhabi, UAE. Association for Computational Linguistics.

\bibitem[{Meister et~al.(2025)Meister, Guestrin, and Hashimoto}]{meister-etal-2025-benchmarking}
Nicole Meister, Carlos Guestrin, and Tatsunori Hashimoto. 2025.
\newblock \href {https://doi.org/10.18653/v1/2025.naacl-long.2} {Benchmarking distributional alignment of large language models}.
\newblock In \emph{Proceedings of the 2025 Conference of the Nations of the Americas Chapter of the Association for Computational Linguistics: Human Language Technologies (Volume 1: Long Papers)}, pages 24--49, Albuquerque, New Mexico. Association for Computational Linguistics.

\bibitem[{Mukherjee et~al.(2024)Mukherjee, Adilazuarda, Sitaram, Bali, Aji, and Choudhury}]{mukherjee-etal-2024-cultural}
Sagnik Mukherjee, Muhammad~Farid Adilazuarda, Sunayana Sitaram, Kalika Bali, Alham~Fikri Aji, and Monojit Choudhury. 2024.
\newblock \href {https://doi.org/10.18653/v1/2024.emnlp-main.884} {Cultural conditioning or placebo? on the effectiveness of socio-demographic prompting}.
\newblock In \emph{Proceedings of the 2024 Conference on Empirical Methods in Natural Language Processing}, pages 15811--15837, Miami, Florida, USA. Association for Computational Linguistics.

\bibitem[{Nadeem et~al.(2021)Nadeem, Bethke, and Reddy}]{nadeem-etal-2021-stereoset}
Moin Nadeem, Anna Bethke, and Siva Reddy. 2021.
\newblock \href {https://doi.org/10.18653/v1/2021.acl-long.416} {{S}tereo{S}et: Measuring stereotypical bias in pretrained language models}.
\newblock In \emph{Proceedings of the 59th Annual Meeting of the Association for Computational Linguistics and the 11th International Joint Conference on Natural Language Processing (Volume 1: Long Papers)}, pages 5356--5371, Online. Association for Computational Linguistics.

\bibitem[{OLMo et~al.(2025)OLMo, Walsh, Soldaini, Groeneveld, Lo, Arora, Bhagia, Gu, Huang, Jordan, Lambert, Schwenk, Tafjord, Anderson, Atkinson, Brahman, Clark, Dasigi, Dziri, Ettinger, Guerquin, Heineman, Ivison, Koh, Liu, Malik, Merrill, Miranda, Morrison, Murray, Nam, Poznanski, Pyatkin, Rangapur, Schmitz, Skjonsberg, Wadden, Wilhelm, Wilson, Zettlemoyer, Farhadi, Smith, and Hajishirzi}]{olmo20252olmo2furious}
Team OLMo, Pete Walsh, Luca Soldaini, Dirk Groeneveld, Kyle Lo, Shane Arora, Akshita Bhagia, Yuling Gu, Shengyi Huang, Matt Jordan, Nathan Lambert, Dustin Schwenk, Oyvind Tafjord, Taira Anderson, David Atkinson, Faeze Brahman, Christopher Clark, Pradeep Dasigi, Nouha Dziri, Allyson Ettinger, Michal Guerquin, David Heineman, Hamish Ivison, Pang~Wei Koh, Jiacheng Liu, Saumya Malik, William Merrill, Lester James~V. Miranda, Jacob Morrison, Tyler Murray, Crystal Nam, Jake Poznanski, Valentina Pyatkin, Aman Rangapur, Michael Schmitz, Sam Skjonsberg, David Wadden, Christopher Wilhelm, Michael Wilson, Luke Zettlemoyer, Ali Farhadi, Noah~A. Smith, and Hannaneh Hajishirzi. 2025.
\newblock \href {http://arxiv.org/abs/2501.00656} {2 olmo 2 furious}.
\newblock \emph{arXiv [cs.CL]}.

\bibitem[{Orlikowski et~al.(2025)Orlikowski, Pei, R{\"o}ttger, Cimiano, Jurgens, and Hovy}]{orlikowski-etal-2025-beyond}
Matthias Orlikowski, Jiaxin Pei, Paul R{\"o}ttger, Philipp Cimiano, David Jurgens, and Dirk Hovy. 2025.
\newblock \href {https://doi.org/10.18653/v1/2025.acl-long.104} {Beyond demographics: Fine-tuning large language models to predict individuals' subjective text perceptions}.
\newblock In \emph{Proceedings of the 63rd Annual Meeting of the Association for Computational Linguistics (Volume 1: Long Papers)}, pages 2092--2111, Vienna, Austria. Association for Computational Linguistics.

\bibitem[{Pang et~al.(2024)Pang, Tang, Ye, Xiong, Zhang, Wang, and Chen}]{pmlr-v235-pang24a}
Xianghe Pang, Shuo Tang, Rui Ye, Yuxin Xiong, Bolun Zhang, Yanfeng Wang, and Siheng Chen. 2024.
\newblock \href {https://proceedings.mlr.press/v235/pang24a.html} {Self-alignment of large language models via monopolylogue-based social scene simulation}.
\newblock In \emph{Proceedings of the 41st International Conference on Machine Learning}, volume 235 of \emph{Proceedings of Machine Learning Research}, pages 39416--39447. PMLR.

\bibitem[{Pedregosa et~al.(2011)Pedregosa, Varoquaux, Gramfort, Michel, Thirion, Grisel, and {Duchesnay}}]{Pedregosa2011-rk}
F~Pedregosa, G~Varoquaux, A~Gramfort, V~Michel, B~Thirion, O~Grisel, and {Duchesnay}. 2011.
\newblock Scikit-learn: Machine learning in python.
\newblock \emph{the Journal of machine Learning research}, 12:2825--2830.

\bibitem[{Pei and Jurgens(2023)}]{pei-jurgens-2023-annotator}
Jiaxin Pei and David Jurgens. 2023.
\newblock \href {https://doi.org/10.18653/v1/2023.law-1.25} {When do annotator demographics matter? measuring the influence of annotator demographics with the {POPQUORN} dataset}.
\newblock In \emph{Proceedings of the 17th Linguistic Annotation Workshop (LAW-XVII)}, pages 252--265, Toronto, Canada. Association for Computational Linguistics.

\bibitem[{Pu et~al.(2025)Pu, Saxon, Hua, and Wang}]{Pu2025-ls}
Xiao Pu, Michael Saxon, Wenyue Hua, and William~Yang Wang. 2025.
\newblock {ThoughtTerminator}: Benchmarking, calibrating, and mitigating overthinking in reasoning models.
\newblock In \emph{Second Conference on Language Modeling}.

\bibitem[{Santurkar et~al.(2023)Santurkar, Durmus, Ladhak, Lee, Liang, and Hashimoto}]{santurkar2023whose}
Shibani Santurkar, Esin Durmus, Faisal Ladhak, Cinoo Lee, Percy Liang, and Tatsunori Hashimoto. 2023.
\newblock Whose opinions do language models reflect?
\newblock In \emph{International Conference on Machine Learning}, pages 29971--30004. PMLR.

\bibitem[{Sap et~al.(2022)Sap, Swayamdipta, Vianna, Zhou, Choi, and Smith}]{sap-etal-2022-annotators}
Maarten Sap, Swabha Swayamdipta, Laura Vianna, Xuhui Zhou, Yejin Choi, and Noah~A. Smith. 2022.
\newblock \href {https://doi.org/10.18653/v1/2022.naacl-main.431} {Annotators with attitudes: How annotator beliefs and identities bias toxic language detection}.
\newblock In \emph{Proceedings of the 2022 Conference of the North American Chapter of the Association for Computational Linguistics: Human Language Technologies}, pages 5884--5906, Seattle, United States. Association for Computational Linguistics.

\bibitem[{Shi et~al.(2025)Shi, Jimenez, Dong, Seo, Yao, Kelch, and Narasimhan}]{shi2025impersona}
Quan Shi, Carlos~E Jimenez, Stephen Dong, Brian Seo, Caden Yao, Adam Kelch, and Karthik~R Narasimhan. 2025.
\newblock \href {https://openreview.net/forum?id=7qhBXq0NLN} {{IMP}ersona: Evaluating individual level {LLM} impersonation}.
\newblock In \emph{Second Conference on Language Modeling}.

\bibitem[{Smucker et~al.(2007)Smucker, Allan, and Carterette}]{smucker_comparison_2007}
Mark~D. Smucker, James Allan, and Ben Carterette. 2007.
\newblock \href {https://doi.org/10.1145/1321440.1321528} {A comparison of statistical significance tests for information retrieval evaluation}.
\newblock In \emph{Proceedings of the Sixteenth ACM Conference on Conference on Information and Knowledge Management}, CIKM '07, page 623–632, New York, NY, USA. Association for Computing Machinery.

\bibitem[{Sorensen et~al.(2025)Sorensen, Mishra, Patel, Tessler, Bakker, Evans, Gabriel, Goodman, and Rieser}]{sorensen-value-profiles-2025}
Taylor Sorensen, Pushkar Mishra, Roma Patel, Michael~Henry Tessler, Michiel~A. Bakker, Georgina Evans, Iason Gabriel, Noah~D. Goodman, and Verena Rieser. 2025.
\newblock \href {https://doi.org/10.48550/arXiv.2503.15484} {Value profiles for encoding human variation}.
\newblock \emph{CoRR}, abs/2503.15484.

\bibitem[{Sorensen et~al.(2024)Sorensen, Moore, Fisher, Gordon, Mireshghallah, Rytting, Ye, Jiang, Lu, Dziri, Althoff, and Choi}]{SorensenMFGMRYJ24}
Taylor Sorensen, Jared Moore, Jillian Fisher, Mitchell~L. Gordon, Niloofar Mireshghallah, Christopher~Michael Rytting, Andre Ye, Liwei Jiang, Ximing Lu, Nouha Dziri, Tim Althoff, and Yejin Choi. 2024.
\newblock \href {https://openreview.net/forum?id=gQpBnRHwxM} {Position: A roadmap to pluralistic alignment}.
\newblock In \emph{ICML}.

\bibitem[{Sun et~al.(2025{\natexlab{a}})Sun, Pei, Choi, and Jurgens}]{sun2023aligning}
Huaman Sun, Jiaxin Pei, Minje Choi, and David Jurgens. 2025{\natexlab{a}}.
\newblock \href {https://arxiv.org/abs/2311.09730} {{Aligning with Whom? Large Language Models Have Gender and Racial Biases in Subjective NLP Tasks}}.
\newblock In \emph{Proceedings of the 2025 Conference of the North American Chapter of the Association for Computational Linguistics (NAACL)}. Association for Computational Linguistics.

\bibitem[{Sun et~al.(2025{\natexlab{b}})Sun, Pei, Choi, and Jurgens}]{sun-etal-2025-sociodemographic}
Huaman Sun, Jiaxin Pei, Minje Choi, and David Jurgens. 2025{\natexlab{b}}.
\newblock \href {https://doi.org/10.18653/v1/2025.naacl-short.71} {Sociodemographic prompting is not yet an effective approach for simulating subjective judgments with {LLM}s}.
\newblock In \emph{Proceedings of the 2025 Conference of the Nations of the Americas Chapter of the Association for Computational Linguistics: Human Language Technologies (Volume 2: Short Papers)}, pages 845--854, Albuquerque, New Mexico. Association for Computational Linguistics.

\bibitem[{Sun et~al.(2024)Sun, Lee, Nan, Zhao, Lee, Jansen, and Kim}]{sun2024randomsiliconsamplingsimulating}
Seungjong Sun, Eungu Lee, Dongyan Nan, Xiangying Zhao, Wonbyung Lee, Bernard~J. Jansen, and Jang~Hyun Kim. 2024.
\newblock \href {http://arxiv.org/abs/2402.18144} {Random silicon sampling: Simulating human sub-population opinion using a large language model based on group-level demographic information}.
\newblock \emph{arXiv [cs.CL]}.

\bibitem[{Tian et~al.(2023)Tian, Mitchell, Zhou, Sharma, Rafailov, Yao, Finn, and Manning}]{tian-etal-2023-just}
Katherine Tian, Eric Mitchell, Allan Zhou, Archit Sharma, Rafael Rafailov, Huaxiu Yao, Chelsea Finn, and Christopher Manning. 2023.
\newblock \href {https://doi.org/10.18653/v1/2023.emnlp-main.330} {Just ask for calibration: Strategies for eliciting calibrated confidence scores from language models fine-tuned with human feedback}.
\newblock In \emph{Proceedings of the 2023 Conference on Empirical Methods in Natural Language Processing}, pages 5433--5442, Singapore. Association for Computational Linguistics.

\bibitem[{Wang et~al.(2025)Wang, Morgenstern, and Dickerson}]{Wang2025-cx}
Angelina Wang, Jamie Morgenstern, and John~P Dickerson. 2025.
\newblock Large language models that replace human participants can harmfully misportray and flatten identity groups.
\newblock \emph{Nat. Mach. Intell.}, 7(3):400--411.

\bibitem[{Wei et~al.(2022)Wei, Wang, Schuurmans, Bosma, Ichter, Xia, Chi, Le, and Zhou}]{wei-cot-2022}
Jason Wei, Xuezhi Wang, Dale Schuurmans, Maarten Bosma, Brian Ichter, Fei Xia, Ed~H. Chi, Quoc~V. Le, and Denny Zhou. 2022.
\newblock Chain-of-thought prompting elicits reasoning in large language models.
\newblock In \emph{Proceedings of the 36th International Conference on Neural Information Processing Systems}, NIPS '22, Red Hook, NY, USA. Curran Associates Inc.

\bibitem[{{Wellcome Trust} and {The Gallup Organization Ltd}(2019)}]{Wellcome-Trust2019-dm}
{Wellcome Trust} and {The Gallup Organization Ltd}. 2019.
\newblock \href {https://wellcome.org/insights/reports/wellcome-global-monitor/2018#downloads-4d1c} {Wellcome global monitor, 2018}.

\bibitem[{Xie et~al.(2025)Xie, Wu, Shen, Jain, Xia, Li, Chang, Rossi, Yu, Kumar, Majumder, Shang, Ammanabrolu, and McAuley}]{xie2025a}
Zhouhang Xie, Junda Wu, Yiran Shen, Raghav Jain, Yu~Xia, Xintong Li, Aaron Chang, Ryan~A. Rossi, Tong Yu, Sachin Kumar, Bodhisattwa~Prasad Majumder, Jingbo Shang, Prithviraj Ammanabrolu, and Julian McAuley. 2025.
\newblock \href {https://openreview.net/forum?id=lSWOMjonL7} {A survey on personalized and pluralistic preference alignment in large language models}.
\newblock In \emph{Second Conference on Language Modeling}.

\bibitem[{Xiong et~al.(2024)Xiong, Hu, Lu, LI, Fu, He, and Hooi}]{xiong2024can}
Miao Xiong, Zhiyuan Hu, Xinyang Lu, YIFEI LI, Jie Fu, Junxian He, and Bryan Hooi. 2024.
\newblock \href {https://openreview.net/forum?id=gjeQKFxFpZ} {Can {LLM}s express their uncertainty? an empirical evaluation of confidence elicitation in {LLM}s}.
\newblock In \emph{The Twelfth International Conference on Learning Representations}.

\bibitem[{Zheng et~al.(2024)Zheng, Pei, Logeswaran, Lee, and Jurgens}]{zheng-etal-2024-helpful}
Mingqian Zheng, Jiaxin Pei, Lajanugen Logeswaran, Moontae Lee, and David Jurgens. 2024.
\newblock \href {https://doi.org/10.18653/v1/2024.findings-emnlp.888} {When ``a helpful assistant'' is not really helpful: Personas in system prompts do not improve performances of large language models}.
\newblock In \emph{Findings of the Association for Computational Linguistics: EMNLP 2024}, pages 15126--15154, Miami, Florida, USA. Association for Computational Linguistics.

\end{thebibliography}

\clearpage

\appendix

\section{Impact of post-training on alignment}
\label{sec:post_training}
To study effects of post-training methods for RQ5, we compare differences in performance between four OLMo-2-7B models (the only completely open source model family we evaluate): the base model, base + supervised finetuning (SFT), base + SFT + direct preference optimization (DPO), and the instruct model (base + SFT + DPO + reinforcement learning with verifiable rewards (RLVR)). The SFT step involves finetuning on instruction datasets and scaled synthetic data, the DPO step involves preference finetuning with DPO, and the RLVR step involves finetuning targeted at domains where prompts with verifiable rewards can be constructed, such as math \cite{olmo20252olmo2furious}. We use these four models out-of-the-box via Huggingface
\footnote{\url{https://huggingface.co/allenai/models}} 
and study differences between base vs.\ SD and calibrated vs.\ uncalibrated with the verbalized elicitation method, since it is typically most aligned.

\begin{figure*}
    \centering
    \includegraphics[width=1\linewidth]{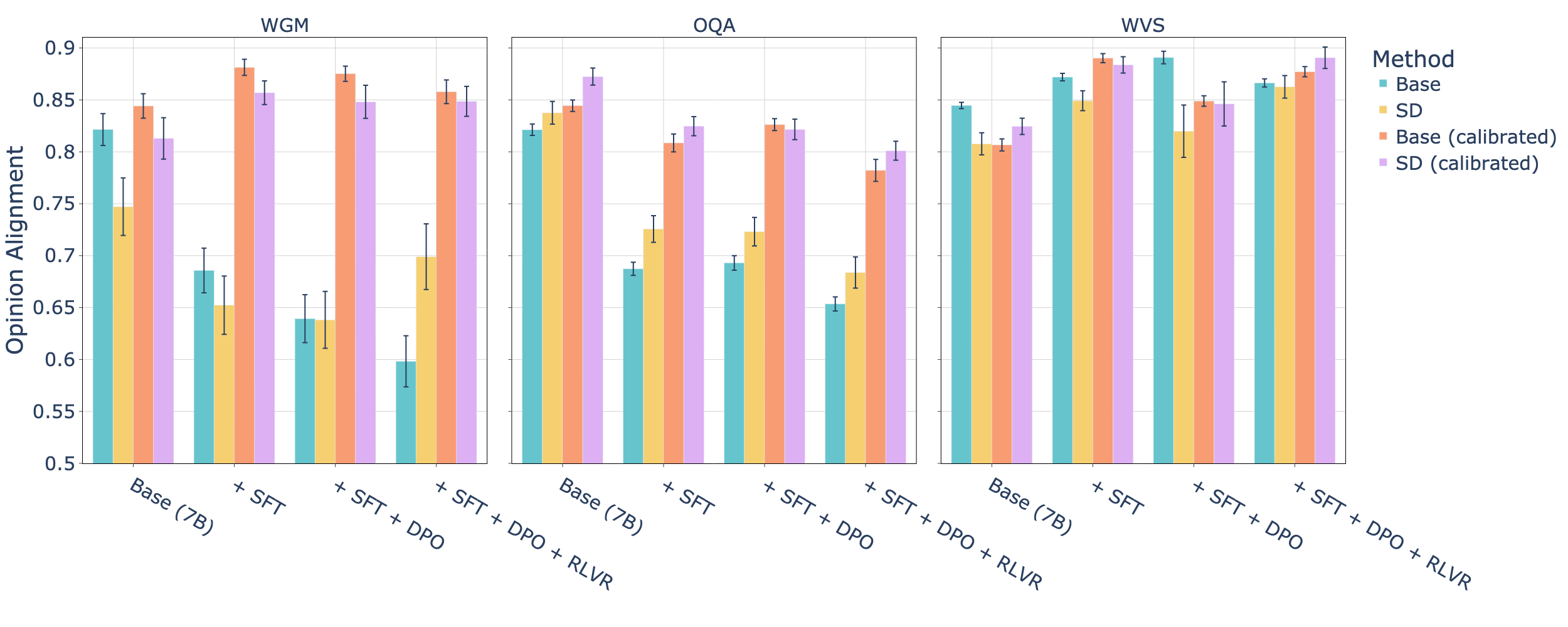}
    \vspace{-3em}
    \caption{Opinion alignment for OLMo-2-7B models with different post-training methods using the verbalized distribution elicitation method. \textbf{Calibration results in more consistent alignment across all post-training methods, both with and without SD prompting. Without calibration, alignment is much more dataset and post-training method dependent.}}
    \label{fig:olmo_verbalized}
\end{figure*}

Figure \ref{fig:olmo_verbalized} plots opinion alignment for all OLMo models. Interestingly, calibration appears to increase opinion alignment less for the base model, and much more so for post-trained models on the WGM and OQA datasets. OLMo models are also much more aligned with human responses on the WVS overall (both with and without calibration) compared to the other two datasets, and calibration appears to have less effect in this setting.

We suspect the increase in alignment on WVS when adding post-training indicates that fine-tuning data might align more with global populations surveyed in WVS. The drop in alignment for uncalibrated distributions on WGM and OQA might indicate that OLMo's post-training data is less aligned with some of the populations of these datasets. Overall, calibration leads to more consistent alignment, with and without SD prompting, regardless of post-training or dataset. This is in agreement with the rest of our results, where using calibration brings more consistency across settings.

\section{The effect of chain of thought prompting}
\label{sec:cot}
We compare CoT prompting with standard prompting + calibration for two models, the model that is most aligned on average (Claude 3.5 v2) and a small model (OLMo-2-7B Instruct). Our results are shown in Table \ref{tab:cot_results}.
We find that for Claude, CoT does not consistently improve alignment over standard prompting, though it does improve in some settings. For base prompting with verbalized elicitation, alignment with CoT is not significantly different than with calibration for 2/3 datasets. For SD prompting, calibration creates significantly more aligned distributions for all datasets. For OLMo, we find that CoT improves alignment much more consistently, often leading to distributions that are just as aligned as, or better aligned than, calibrated distributions for the verbalized setting. Calibration is consistently more aligned than CoT for distributions elicited via paraphrase or self-random.

Upon qualitative inspection, we found that SD + CoT prompting Claude led to an ``overthinking'' effect \cite{Pu2025-ls}, where the model sometimes generated a distribution that exaggerated a particular difference with the ground truth. For example, the knowledge that ``Men generally report higher confidence in scientific knowledge than women due to societal factors'' to generate a distribution for the question ``How much do you know about science?'' led to the model producing a distribution that was more skewed than the ground truth. We hypothesize that this led to lower alignment for Claude (and would be similar for other larger, more capable reasoning models), but not for smaller models like OLMo 7B. Future work might study this further.

\begin{table*}[h]
    \centering
    \footnotesize
    \addtolength{\tabcolsep}{-0.3em}
\begin{tabular}{cc|rrrrrrrrr|rrrrrrrrr}
\toprule
& \multirow{2}{*}{Model} &  \multicolumn{9}{c}{Base prompt} & \multicolumn{9}{c}{Sociodemographic prompt} \\
& & $P$ & \!$P_{cot}$ & \!$P_C$ & $S$ & \!$S_{cot}$ & \!$S_C$ & $V$ & \!$V_{cot}$ & \!$V_C$ & $P$ & \!$P_{cot}$ & \!$P_C$ & $S$ & \!$S_{cot}$ & \!$S_C$ & $V$ & \!$V_{cot}$ & \!$V_C$ \\
\midrule
\multirow{2}{*}{\rotatebox[origin=c]{90}{WGM}} & 
Claude & 66.2 & 65.6 & \textbf{85.2} & 59.5 & 60.7 & \textbf{85.1} & \cellcolor{yellow!30}{89.3} & 88.8 & 87.7 & 65.4 & 61.7 & \textbf{84.4} & 61.4 & 58.6 & \textbf{84.2} & \textbf{89.0} & 79.7 & \cellcolor{yellow!30}\textbf{89.3} \\
& OLMo & 59.6 & 62.9 & \textbf{81.5} & 67.7 & 62.7 & \textbf{80.4} & 59.8 & \cellcolor{yellow!30}\textbf{86.1} & \textbf{85.1} & 62.3 & 64.0 & \textbf{82.5} & 64.5 & 59.8 & \textbf{82.7} & 69.8 & 81.8 & \textbf{84.6} \\
\midrule
\multirow{2}{*}{\rotatebox[origin=c]{90}{OQA}} & 
Claude & 70.5 & 76.4 & \textbf{88.8} & 72.5 & 74.3 & \textbf{90.4} & \textbf{91.7} & 89.9 & \textbf{91.7} & 76.1 & 68.0 & \textbf{89.9} & 73.3 & 66.1 & \textbf{89.6} & \cellcolor{yellow!30}\textbf{91.9} & 80.3 & \textbf{91.6} \\
& OLMo & 72.6 & 79.3 & \cellcolor{yellow!30}\textbf{89.0} & 72.3 & 77.0 & \textbf{88.4} & 65.4 & \textbf{85.2} & 79.9 & 72.8 & 77.1 & \textbf{88.4} & 70.5 & 76.3 & \textbf{88.6} & 68.5 & \textbf{82.9} & \textbf{81.5} \\
\midrule
\multirow{2}{*}{\rotatebox[origin=c]{90}{WVS}} & 
Claude & 46.5 & 55.5 & \textbf{80.3} & 51.0 & 52.8 & \textbf{80.3} & 75.2 & \textbf{80.3} & \textbf{80.3} & 61.0 & 61.4 & \textbf{80.4} & 56.8 & 59.8 & \textbf{80.4} & 75.6 & {79.7} & \cellcolor{yellow!30}\textbf{81.7} \\
& OLMo & 75.3 & 76.9 & \textbf{80.3} & 74.9 & 73.5 & \textbf{80.3} & {86.6} & 86.5 & 86.5 & 58.3 & 72.7 & \textbf{79.2} & 60.0 & 69.4 & \textbf{77.2} & 86.0 & 85.7 & \cellcolor{yellow!30}\textbf{89.8} \\
\bottomrule
\end{tabular}
\caption{Opinion alignment for original, CoT prompted, and calibrated distributions for base and SD prompts. Each set of three columns compares the elicitation methods: paraphrase ($P$), self-random ($S$), and verbalized ($V$). \textbf{Bolded} values are significantly better than others in each triple (calculated via Anova followed by paired t-tests and Bonferroni correction). The \tcboxmath[colback=yellow!30]{$highlighted$} value is the highest alignment for each model. Claude refers to Claude-3.5-v2 and OLMo refers to OLMo-2-7B-Instruct. \textbf{CoT significantly improves alignment for verbalized elicitation for OLMo. For Claude, verbalized alignment is not significantly different or drops with CoT.}}
\label{tab:cot_results}
\vspace{-1em}
\end{table*}

\section{Comparison with few-shot prompting}
\label{sec:few_shot}
Given that the regression models appear to converge with so few examples, one might ask whether the regression is actually learning anything useful, or if we could achieve a similar improvement with \textit{few shot} prompting. To investigate this, we 5-shot prompt the LLMs with the \textit{same} examples given to the regression model in the 5 full example calibration setting. We use a 5-shot prompt since this lies in the middle of the number of full examples that were needed for convergence for the regression models. Whereas we averaged over 10 random samples in those experiments, here we use only 3 of the random samples because we found little variation. We average alignment over the 3 runs. 

Table \ref{tab:few_shot_results} compares few shot prompted ($F$) distributions to original and calibrated ($C$) distributions for base and SD prompts.
While 5-shot prompted distributions are not as aligned as calibrated distributions in most settings, they are significantly more aligned in a few cases. However, calibration leads to distributions that are more aligned more \textit{consistently}, across models, prompting methods, and distribution elicitation methods.

\begin{table*}[]
    \centering
    \footnotesize
    \addtolength{\tabcolsep}{-0.3em}
\begin{tabular}{cc|rrrrrrrrr|rrrrrrrrr}
\toprule
& \multirow{2}{*}{Model} &  \multicolumn{9}{c}{Base prompt} & \multicolumn{9}{c}{Sociodemographic prompt} \\
& & $P$ & \!$P_F$ & \!$P_C$ & $S$ & \!$S_F$ & \!$S_C$ & $V$ & \!$V_F$ & \!$V_C$ & $P$ & \!$P_F$ & \!$P_C$ & $S$ & \!$S_F$ & \!$S_C$ & $V$ & \!$V_F$ & \!$V_C$ \\
\midrule
\multirow{5}{*}{\rotatebox[origin=c]{90}{WGM}} & 
Claude & 66.2 & 63.5 & \textbf{86.4} & 59.5 & 61.7 & \textbf{83.7} & \cellcolor{yellow!30}{89.3} & 89.0 & 89.2 & 65.4 & 59.8 & \textbf{84.1} & 61.4 & 58.1 & \textbf{84.2} & \textbf{89.0} & 78.9 & \textbf{89.0} \\
& Llama & 68.1 & 72.5 & \textbf{82.8} & 73.0 & 66.5 & \textbf{86.4} & 84.8 & 86.0 & \textbf{89.2} & 70.6 & 66.7 & \textbf{86.0} & 67.6 & 64.9 & \textbf{85.6} & 85.0 & 78.7 & \cellcolor{yellow!30}\textbf{89.6} \\
& Mistral & 62.4 & 59.5 & \textbf{84.5} & 72.0 & 55.4 & \textbf{83.7} & \textbf{89.4} & 87.2 & \cellcolor{yellow!30}\textbf{89.6} & 68.4 & 58.4 & \textbf{85.8} & 63.0 & 55.3 & \textbf{84.9} & 87.3 & 78.1 & \textbf{88.5} \\
& OLMo & 59.6 & 70.5 & \textbf{81.1} & 67.7 & 70.9 & \textbf{81.0} & 59.8 & 81.1 & \cellcolor{yellow!30}\textbf{83.7} & 62.3 & 64.2 & \textbf{82.2} & 64.5 & 66.0 & \textbf{81.6} & 69.8 & \textbf{79.3} & \textbf{80.6} \\
& Qwen & 57.7 & 60.6 & \textbf{82.8} & 53.3 & 59.0 & \textbf{82.1} & \textbf{88.2} & \textbf{88.3} & 87.3 & 66.0 & 58.5 & \textbf{85.2} & 63.4 & 55.9 & \textbf{85.4} & \textbf{89.1} & 80.4 & \cellcolor{yellow!30}\textbf{89.4}\\
\cline{2-20}
& Average & 62.8 & 65.3 & \textbf{83.5} & 65.1 & 62.7 & \textbf{83.4} & 82.3 & 86.3 & \textbf{87.8} & 66.5 & 61.5 & \textbf{84.7} & 64.0 & 60.0 & \textbf{84.3} & 84.0 & 79.1 & \cellcolor{cyan!30}\textbf{87.4} \\
\bottomrule
\multirow{5}{*}{\rotatebox[origin=c]{90}{OQA}} & 
Claude & 70.5 & 77.9 & \textbf{88.4} & 72.5 & 75.7 & \textbf{89.6} & {91.7} & 91.5 & 91.5 & 76.1 & 68.1 & \textbf{89.7} & 73.3 & 66.9 & \textbf{89.7} & 91.9 & 79.7 & \cellcolor{yellow!30}\textbf{92.3} \\
& Llama & 79.3 & 77.6 & \textbf{89.7} & 75.3 & 76.6 & \textbf{89.6} & 86.8 & \textbf{88.8} & \textbf{88.6} & 79.2 & 75.0 & \cellcolor{yellow!30}\textbf{90.0} & 76.1 & 75.5 & \textbf{89.8} & 83.4 & 80.5 & \textbf{87.1} \\
& Mistral & 79.5 & 77.4 & \textbf{89.6} & 75.3 & 74.0 & \textbf{89.0} & 85.0 & \textbf{89.7} & 86.9 & 75.8 & 70.5 & \cellcolor{yellow!30}\textbf{89.8} & 72.4 & 69.9 & \textbf{87.5} & 83.8 & 80.7 & \textbf{87.8} \\
& OLMo & 72.6 & 77.0 & \cellcolor{yellow!30}\textbf{88.9} & 72.3 & 74.6 & \textbf{85.9} & 65.4 & \textbf{81.2} & 77.4 & 72.8 & 72.3 & \textbf{88.6} & 70.5 & 74.4 & \textbf{88.4} & 68.5 & 78.8 & \textbf{86.6} \\
& Qwen & 73.9 & 75.9 & \textbf{90.1} & 67.0 & 74.4 & \textbf{89.3} & \textbf{88.4} & \textbf{88.9} & 87.2 & 74.4 & 71.4 & \cellcolor{yellow!30}\textbf{90.2} & 71.3 & 69.3 & \textbf{89.3} & \textbf{89.2} & 81.0 & 87.8 \\
\cline{2-20}
& Average &  75.2 & 77.2 & \textbf{89.3} & 72.5 & 75.1 & \textbf{88.7} & 83.5 & \textbf{88.0} & 86.3 & 75.7 & 71.5 & \cellcolor{cyan!30}\textbf{89.7} & 72.7 & 71.2 & \textbf{88.9} & 83.4 & 80.1 & \textbf{88.3}\\
\bottomrule
\multirow{5}{*}{\rotatebox[origin=c]{90}{WVS}} & 
Claude & 46.5 & 62.4 & \textbf{81.7} & 51.0 & 62.1 & \textbf{80.8} & 75.2 & \cellcolor{yellow!30}\textbf{82.1} & \textbf{81.5} & 61.0 & 64.7 & \textbf{80.4} & 56.8 & 64.0 & \textbf{80.2} & 75.6 & \textbf{81.9} & 80.4 \\
& Llama & 61.6 & 70.1 & \textbf{79.4} & 59.1 & 68.9 & \textbf{80.9} & 64.6 & 79.7 & \textbf{83.5} & 62.1 & 70.3 & \textbf{80.0} & 59.5 & 72.2 & \textbf{80.9} & 67.7 & 78.8 & \cellcolor{yellow!30}\textbf{86.2} \\
& Mistral & 48.8 & 63.9 & \textbf{81.4} & 44.0 & 61.0 & \textbf{80.6} & 72.8 & \textbf{81.0} & \textbf{81.1} & 54.3 & 65.0 & \textbf{82.4} & 51.5 & 62.7 & \textbf{81.4} & 76.6 & 78.7 & \cellcolor{yellow!30}\textbf{82.5} \\
& OLMo & 75.3 & 72.9 & \textbf{80.0} & 74.9 & 74.2 & \textbf{80.2} & 86.6 & 84.2 & \textbf{87.8} & 58.3 & 68.8 & \textbf{78.8} & 60.0 & 72.3 & \textbf{77.5} & 86.0 & 78.0 & \cellcolor{yellow!30}\textbf{89.7} \\
& Qwen & 39.6 & 58.3 & \cellcolor{yellow!30}\textbf{82.1} & 42.4 & 56.8 & \textbf{81.3} & 74.0 & 78.4 & \textbf{81.0} & 49.1 & 59.0 & \textbf{81.5} & 49.4 & 56.8 & \textbf{81.2} & 73.4 & \textbf{78.3} & 76.6 \\
\cline{2-20}
& Average & 54.4 & 65.5 & \textbf{80.9} & 54.3 & 64.6 & \textbf{80.8} & 74.6 & 81.1 & \textbf{83.0}& 57.0 & 65.6 & \textbf{80.6} & 55.4 & 65.6 & \textbf{80.2} & 75.9 & 79.1 & \cellcolor{cyan!30}\textbf{83.1} \\
\bottomrule
\end{tabular}
\caption{Opinion alignment for original, few shot prompted$_F$, and calibrated$_C$ distributions for base and SD prompts. Each set of three columns compares the three elicitation methods: paraphrase (`P'), self-random (`S'), and verbalized (`V'). \textbf{Bolded} values are significant between each triple (calculated via Anova followed by paired t-tests and Bonferroni correction). The value highlighted in 
\tcboxmath[colback=yellow!30]{$yellow$} is the absolute best performing overall for each model. The values highlighted in \tcboxmath[colback=cyan!30]{$blue$} are the best averages across models. Each model refers to the largest model we study within the model family. \textbf{Few shot prompting does not consistently improve alignment across datasets, models, and elicitation methods.}}
\label{tab:few_shot_results}
\vspace{-1em}
\end{table*}

\section{Additional Results}
\label{sec:appendix_results}

\subsection{Alignment results for all models}
\label{sec:appendix_all_models}
Results are in Table \ref{tab:regression_full}.

\subsection{Log Probability Results}
\label{sec:appendix_log_probs}

We get the model's log probabilities for each answer choice \cite{geng-etal-2024-survey} and normalize to get the distribution over all answer choices, following prior work \cite{santurkar2023whose}. We obtain log probabilities for the smaller Llama models. Log probability results are shown in Table \ref{tab:log_prob_results}. We find that using log probabilities with calibration yielded the most aligned distributions in the 10/24 settings considered. Future work might look more into comparisons between logit-based and verbalized distribution elicitation.

\begin{table}[h]
\centering
\footnotesize
\begin{tabular}{cc|rr|rr}
\toprule
& \multirow{2}{*}{Model} &  \multicolumn{2}{c}{Base prompt} & \multicolumn{2}{c}{SD prompt} \\
& & $L$ & \!$L_C$ & $L$ & \!$L_C$ \\
\midrule
\multirow{4}{*}{\rotatebox[origin=c]{90}{WGM}} & Llama-3-70B & 73.1 & \textbf{86.2} & 67.5 & \textbf{86.9} \\
& Llama-3.1-70B & 74.0 & \textbf{83.2} & 70.4 & \textbf{88.1} \\
& Llama-3.2-1B & 83.6 & \textbf{88.6} & 83.0 & {85.1} \\
& Llama-3.2-11B & 84.0 & {85.0} & 82.2 & {84.4} \\
\midrule
& Average & 78.7 & \textbf{85.7} & 75.8 & \textbf{86.1}\\
& Std Dev & 5.9 & 2.3 & 8.0 & 1.7\\
\bottomrule
\multirow{4}{*}{\rotatebox[origin=c]{90}{OQA}} & Llama-3-70B & 75.4 & \textbf{85.2} & 72.3 & \textbf{88.1} \\
& Llama-3.1-70B & 78.9 & \textbf{87.3} & 75.7 & \textbf{87.4} \\
& Llama-3.2-1B  & 88.3 & {88.4} & {87.3} & 86.5 \\
& Llama-3.2-11B & 82.6 & \textbf{86.3} & 88.3 & {89.6} \\
\midrule
& Average & 81.3 & \textbf{86.8} & 80.9 & \textbf{87.9}\\
& Std Dev & 5.5 & 1.4 & 8.1 & 1.3\\
\bottomrule
\multirow{4}{*}{\rotatebox[origin=c]{90}{WVS}} & Llama-3-70B  & 54.1 & \textbf{82.8} & 51.1 & \textbf{83.5} \\
& Llama-3.1-70B   & 43.5 & \textbf{83.3} & 49.1 & \textbf{83.0} \\
& Llama-3.2-1B  & 83.9 & 86.4 & {85.1} & 84.8 \\
& Llama-3.2-11B   & 74.8 & \textbf{81.1} & 72.9 & \textbf{85.5} \\
\midrule
& Average & 64.1 & \textbf{83.4} & 64.5 & \textbf{84.2}\\
& Std Dev & 18.6 & 2.2 & 17.4 & 1.1\\
\bottomrule
\end{tabular}
\caption{Opinion alignment before and after calibration for each dataset and LLM, using log probability distributions. Each pair of columns compares the base-generated or SD-generated distributions to the calibrated distributions ($C$) for log probabilities (`L'). Bolded values are significant between each pair. The mean and standard deviation across models are shown in the bottom rows of each dataset section.}
\label{tab:log_prob_results}
\end{table}

\begin{table*}[h]
    \centering
    \footnotesize
    \addtolength{\tabcolsep}{-0.2em}
\begin{tabular}{cc|rrrrrr|rrrrrr}
\toprule
& \multirow{2}{*}{Model} &  \multicolumn{6}{c}{Base prompt} & \multicolumn{6}{c}{Sociodemographic prompt} \\
& & $P$ & \!$P_C$ & $S$ & \!$S_C$ & $V$ & \!$V_C$ & $P$ & \!$P_C$ & $S$ & \!$S_C$ & $V$ & \!$V_C$ \\
\midrule
\multirow{15}{*}{\rotatebox[origin=c]{90}{WGM}} & OLMo-2-7B-Base  & 65.6 & \textbf{80.1} & 79.2 & {81.8} & 82.2 & {83.9} & 74.8 & \textbf{81.9} & 71.0 & \textbf{81.8} & 74.9 & \textbf{82.9} \\
& OLMo-2-7B-SFT  & 80.6 & {81.8} & 70.3 & \textbf{81.8} & 68.6 & \textbf{86.3} & 75.6 & \textbf{82.6} & 70.2 & \textbf{82.1} & 65.4 & \textbf{85.0} \\
& OLMo-2-7B-DPO  & 71.1 & \textbf{77.4} & 63.2 & \textbf{82.1} & 63.9 & \textbf{86.2} & 72.2 & \textbf{83.0} & 68.7 & \textbf{82.4} & 63.8 & \textbf{84.9} \\
& OLMo-2-7B-Instruct  & 59.6 & \textbf{81.5} & 67.7 & \textbf{80.4} & 59.8 & \textbf{85.1} & 62.3 & \textbf{82.5} & 64.5 & \textbf{82.7} & 69.8 & \textbf{84.6} \\
& Llama-3-70B  & 64.9 & \textbf{85.3} & 73.7 & \textbf{83.8} & 84.5 & \textbf{88.8}  & 66.5 & \textbf{86.0} & 61.7 & \textbf{84.7} & 84.4 & \textbf{88.3} \\
& Llama-3.1-70B  & 73.2 & \textbf{86.7} & 70.3 & \textbf{86.5} & 80.6 & \textbf{86.0}  & 68.5 & \textbf{85.9} & 66.1 & \textbf{85.9} & 85.2 & \textbf{89.4} \\
& Llama-3.2-1B  & {81.8} & {81.8} & {81.8} & {81.8}  & -- & -- & 61.9 & \textbf{82.5} & 46.7 & \textbf{81.2} & 61.4 & {86.4} \\
& Llama-3.2-11B  & 73.7 & \textbf{87.7} & 73.2 & \textbf{83.5} & 73.6 & \textbf{84.5} & 71.8 & \textbf{85.3} & 72.0 & \textbf{84.7} & 71.7 & {79.2} \\
& Llama-3.2-90B & 68.1 & \textbf{84.6} & 73.0 & \textbf{86.2} & 84.8 & \textbf{89.0} & 70.6 & \textbf{86.0} & 67.6 & \textbf{86.1} & 85.0 & \textbf{89.8} \\
& Qwen-2.5-72B  & 57.7 & \textbf{83.1} & 53.3 & \textbf{83.2} & {88.2} & 87.1  & 66.0 & \textbf{84.1} & 63.4 & \textbf{85.2} & 89.1 & {89.4} \\
& Mistral-small & 61.6 & \textbf{83.9} & 61.6 & \textbf{84.6} & {88.9} & 88.8  & 67.5 & \textbf{86.6} & 64.4 & \textbf{85.8} & 89.0 & {89.1} \\
& Mistral-large  & 62.4 & \textbf{84.7} & 72.0 & \textbf{83.9} & {89.4} & 88.9 & 68.4 & \textbf{84.7} & 63.0 & \textbf{84.9} & 87.3 & {88.4} \\
& Claude-3  & 57.8 & \textbf{81.1} & 50.0 & \textbf{81.8} & 84.8 & {86.7} & 64.9 & \textbf{80.6} & 52.9 & \textbf{81.7} & {88.4} & 88.1 \\
& Claude-3.5-v1  & 64.5 & \textbf{82.5} & 62.0 & \textbf{84.3} & 86.1 & {87.4}  & 65.8 & \textbf{84.6} & 62.3 & \textbf{84.5} & 86.3 & {88.3} \\
& Claude-3.5-v2  & 66.2 & \textbf{85.2} & 59.5 & \textbf{85.1} & {89.3} & 87.7  & 65.4 & \textbf{84.4} & 61.4 & \textbf{84.2} & 89.0 & {89.3} \\
\midrule
& Average  & 67.3 & \textbf{83.2} & 67.4 & \textbf{83.4} & 80.3 & \textbf{86.9} & 68.1 & \textbf{84.0} & 63.7 & \textbf{83.9} & 79.4 & \textbf{86.9} \\
& Std Dev  & 7.5 & 2.7 & 9.0 & 1.8 & 9.9 & 1.7  & 4.1 & 1.8 & 6.7 & 1.7 & 10.3 & 3.0 \\
\bottomrule
\multirow{15}{*}{\rotatebox[origin=c]{90}{OQA}} & OLMo-2-7B-Base  & 85.0 & \textbf{89.2} & 82.6 & \textbf{88.4} & {82.1} & 81.8 & 80.8 & \textbf{88.3} & 77.2 & \textbf{88.1} & 83.8 & \textbf{87.6} \\
& OLMo-2-7B-SFT  & 78.8 & \textbf{88.0} & 79.4 & \textbf{86.8} & 68.7 & \textbf{81.6} & 80.7 & \textbf{87.9} & 77.8 & \textbf{87.2} & 72.6 & \textbf{83.0} \\
& OLMo-2-7B-DPO  & 79.3 & \textbf{88.6} & 81.5 & \textbf{86.7} & 69.3 & \textbf{82.8} & 82.1 & \textbf{88.4} & 78.7 & \textbf{87.7} & 72.3 & \textbf{82.7} \\
& OLMo-2-7B-Instruct  & 72.6 & \textbf{89.0} & 72.3 & \textbf{88.4} & 65.4 & \textbf{79.9} & 72.8 & \textbf{88.4} & 70.5 & \textbf{88.6} & 68.5 & \textbf{81.5} \\
& Llama-3-70B  & 72.2 & \textbf{87.5} & 76.8 & \textbf{89.1} & 81.8 & \textbf{85.0} & 74.4 & \textbf{88.9} & 70.1 & \textbf{88.9} & 79.5 & \textbf{83.7} \\
& Llama-3.1-70B  & 73.5 & \textbf{88.8} & 70.5 & \textbf{87.7} & 83.8 & \textbf{86.4} & 76.4 & \textbf{89.5} & 71.7 & \textbf{89.4} & 83.6 & \textbf{86.6} \\
& Llama-3.2-1B  & 82.3 & {88.4} & {88.4} & \textbf{88.4} & -- & -- & 72.6 & \textbf{89.0} & 70.2 & \textbf{88.6} & 83.6 & {84.9} \\
& Llama-3.2-11B  & 84.3 & \textbf{89.4} & 75.5 & \textbf{86.4} & 65.8 & \textbf{77.2} & 75.7 & \textbf{87.2} & 74.8 & \textbf{87.0} & 83.0 & {86.9} \\
& Llama-3.2-90B  & 79.3 & \textbf{89.6} & 75.3 & \textbf{89.4} & 86.8 & {87.9} & 79.2 & \textbf{90.1} & 76.1 & \textbf{90.0} & 83.4 & \textbf{85.9} \\
& Qwen-2.5-72B  & 73.9 & \textbf{89.7} & 67.0 & \textbf{88.8} & {88.4} & 87.9 & 74.4 & \textbf{90.0} & 71.3 & \textbf{89.5} & {89.2} & 88.6 \\
& Mistral-small  & 77.3 & \textbf{91.6} & 73.6 & \textbf{90.5} & 86.9 & {87.7} & 75.1 & \textbf{90.4} & 71.6 & \textbf{90.3} & 87.8 & {89.0} \\
& Mistral-large  & 79.5 & \textbf{89.9} & 75.3 & \textbf{88.3} & 85.0 & {86.2} & 75.8 & \textbf{89.2} & 72.4 & \textbf{89.5} & 83.8 & {84.7} \\
& Claude-3  & 73.5 & \textbf{89.9} & 66.4 & \textbf{90.3} & {87.6} & 86.5 & 76.4 & \textbf{90.6} & 72.6 & \textbf{90.3} & {89.3} & 88.6 \\
& Claude-3.5-v1  & 70.5 & \textbf{86.9} & 69.6 & \textbf{88.6} & {89.4} & 89.0  & 76.1 & \textbf{90.3} & 72.8 & \textbf{89.8} & {91.0} & 90.8 \\
& Claude-3.5-v2  & 70.5 & \textbf{88.8} & 72.5 & \textbf{90.4} & {91.7} & {91.7} & 76.1 & \textbf{89.9} & 73.3 & \textbf{89.6} & {91.9} & 91.6 \\
\midrule
& Average  & 76.8 & \textbf{89.0} & 75.1 & \textbf{88.6} & 80.9 & \textbf{85.1} & 76.6 & \textbf{89.2} & 73.4 & \textbf{89.0} & 82.9 & \textbf{86.4} \\
& Std Dev  & 4.8 & 1.1 & 6.0 & 1.3 & 9.4 & 4.0 & 2.9 & 1.0 & 2.8 & 1.1 & 7.0 & 3.0 \\
\bottomrule
\multirow{15}{*}{\rotatebox[origin=c]{90}{WVS}} & OLMo-2-7B-Base  & 74.5 & \textbf{80.3} & 70.7 & \textbf{80.1} & \textbf{84.5} & 80.3  & 78.2 & {81.4} & 78.2 & {81.5} & 80.8 & {81.9} \\
& OLMo-2-7B-SFT  & 69.7 & \textbf{79.2} & 68.1 & \textbf{79.1} & {87.2} & {87.2}& 72.9 & \textbf{80.6} & 73.1 & \textbf{80.7} & 84.9 & \textbf{89.0} \\
& OLMo-2-7B-DPO & 66.2 & \textbf{80.3} & 70.5 & \textbf{79.8} & \textbf{89.1} & 83.9  & 70.6 & \textbf{80.9} & 70.2 & \textbf{80.7} & 82.5 & {85.6} \\
& OLMo-2-7B-Instruct  & 75.3 & \textbf{80.3} & 74.9 & \textbf{80.3} & {86.6} & 86.5  & 58.3 & \textbf{79.2} & 60.0 & \textbf{77.2} & 86.0 & \textbf{89.8} \\
& Llama-3-70B  & 59.7 & \textbf{80.3} & 61.7 & \textbf{80.6} & 81.3 & {83.0}  & 67.9 & \textbf{80.8} & 62.8 & \textbf{80.5} & 78.1 & {82.1} \\
& Llama-3.1-70B  & 61.9 & \textbf{81.5} & 56.4 & \textbf{82.1} & 65.2 & \textbf{80.3}  & 63.6 & \textbf{80.8} & 57.6 & \textbf{81.3} & 68.3 & \textbf{82.8} \\
& Llama-3.2-1B & {80.3} & {80.3} & {80.3} & {80.3} & -- & -- & 68.7 & \textbf{82.8} & 66.6 & \textbf{78.9} & -- & -- \\
& Llama-3.2-11B  & 75.4 & \textbf{82.0} & 76.0 & \textbf{84.8} & 82.7 & {84.0} & 78.2 & \textbf{82.2} & 78.0 & {81.2} & 66.5 & {87.0} \\
& Llama-3.2-90B  & 61.6 & \textbf{79.9} & 59.1 & \textbf{80.3} & 64.6 & \textbf{80.3}  & 62.1 & \textbf{80.8} & 59.5 & \textbf{81.5} & 67.7 & \textbf{82.7} \\
& Qwen-2.5-72B  & 39.6 & \textbf{82.0} & 42.4 & \textbf{82.3} & 74.0 & {77.9} & 49.1 & \textbf{82.4} & 49.4 & \textbf{81.5} & 73.4 & \textbf{81.7} \\
& Mistral-small  & 42.9 & \textbf{80.3} & 46.3 & \textbf{80.3} & 75.0 & \textbf{80.3}  & 48.0 & \textbf{80.4} & 46.4 & \textbf{80.4} & 68.6 & \textbf{81.7} \\
& Mistral-large  & 48.8 & \textbf{80.3} & 44.0 & \textbf{82.2} & 72.8 & \textbf{80.3}  & 54.3 & \textbf{80.4} & 51.5 & \textbf{80.3} & 76.6 & \textbf{83.8} \\
& Claude-3 & 47.3 & \textbf{80.3} & 55.7 & \textbf{80.3} & 74.1 & \textbf{80.3} & 55.3 & \textbf{80.2} & 57.3 & \textbf{79.8} & 73.2 & \textbf{81.7} \\
& Claude-3.5-v1 & 44.3 & \textbf{80.3} & 49.6 & \textbf{80.3} & 75.7 & \textbf{80.3}  & 58.7 & \textbf{80.3} & 54.8 & \textbf{80.3} & 73.3 & {81.6} \\
& Claude-3.5-v2  & 46.5 & \textbf{80.3} & 51.0 & \textbf{80.3} & 75.2 & \textbf{80.3}  & 61.0 & \textbf{80.4} & 56.8 & \textbf{80.4} & 75.6 & {81.7} \\
\midrule
& Average  & 59.6 & \textbf{80.5} & 60.4 & \textbf{80.9} & 77.7 & \textbf{81.8} & 63.1 & \textbf{80.9} & 61.5 & \textbf{80.4} & 75.4 & \textbf{83.8} \\
& Std Dev & 13.8 & 0.7 & 12.4 & 1.4 & 7.8 & 2.7 & 9.5 & 0.9 & 9.9 & 1.1 & 6.4 & 2.9 \\
\bottomrule
\end{tabular}
\caption{Opinion alignment before and after calibration for each dataset, LLM, and elicitation method. Each pair of columns compares the base-generated or SD-generated distributions to the calibrated distributions ($C$) for each elicitation method: paraphrase (`P'), self-random (`S'), and verbalized (`V'). Bolded values are significant between each pair (calculated via paired t-test and Bonferroni correction, see Appendix \ref{sec:appendix_stat_sig}). The mean and standard deviation across models are shown in the bottom rows of each dataset section. We note that for some verbalized elicitation settings, Llama-3.2-1B was unable to follow instructions and either did not return any distributions in the format we required for parsing, or refused to produce a distribution (see Table \ref{tab:num_extracted_distributions}).}
\label{tab:regression_full}
\end{table*}

\subsection{Individual Sociodemographic Results}
\label{sec:appendix_individual_sd_results}

Results for all demographics for the WGM, OQA, and WVS datasets are in Tables \ref{tab:individual_dems_wgm_all}, \ref{tab:individual_dems_oqa_all}, and \ref{tab:individual_dems_wvs_all} respectively.
\begin{table*}[]
    \centering
    \footnotesize
\begin{tabular}{llllllllllll|ll}
\toprule
& \multirow{2}{*}{Demographic} & \multicolumn{2}{l}{Claude-3.5-v2} & \multicolumn{2}{l}{Llama-3.2-90B} & \multicolumn{2}{l}{Mistral-large} & \multicolumn{2}{r}{OLMo-2-7B-I} & \multicolumn{2}{l|}{Qwen-2.5-72B} & \multicolumn{2}{l}{Average} \\
 & & $V$ & $V_C$ & $V$ & $V_C$ & $V$ & $V_C$ & $V$ & $V_C$ & $V$ & $V_C$ & $V$ & $V_C$ \\
\midrule
\multirow{3}{*}{\rotatebox[origin=c]{90}{Age}} & 
15-29 & 95.2 & 92.7 & 85.0 & 94.1 & 92.9 & 92.5 & 56.6 & \textbf{88.6} & 90.2 & 90.7 & 84.0 & 91.7 \\
& 30-49 & 92.7 & 90.7 & 85.7 & \textbf{93.1} & 93.1 & 92.7 & 58.9 & \textbf{87.0} & 90.6 & 89.8 & 84.2 & 90.7 \\
& 50+ & 88.6 & 87.5 & 87.9 & 91.0 & 91.4 & 90.7 & 63.7 & 85.9 & 90.6 & 87.1 & 84.4 & 88.4 \\
\midrule
\multirow{3}{*}{\rotatebox[origin=c]{90}{Edu.}} &
primary & 83.9 & 83.8 & 78.0 & 86.3 & 84.6 & 84.0 & 62.1 & \textbf{88.6} & 82.5 & 82.6 & 78.2 & 85.1 \\
& secondary & 95.1 & 91.2 & 86.5 & 91.8 & 92.1 & 92.2 & 57.0 & \textbf{85.8} & 91.1 & 90.5 & 84.4 & 90.3 \\
& tertiary & 86.2 & 82.7 & 88.6 & 88.4 & 90.4 & 89.5 & 63.8 & 86.2 & 94.6 & 90.6 & 84.7 & 87.5 \\
\midrule
\multirow{6}{*}{\rotatebox[origin=c]{90}{Employment}} &
FT for em. & 92.3 & 88.5 & 89.1 & 91.1 & 93.4 & 92.7 & 60.6 & 86.0 & 93.9 & 90.8 & 85.9 & 89.8 \\
& FT for se. & 91.0 & 89.7 & 84.3 & \textbf{91.3} & 91.5 & 91.2 & 60.0 & \textbf{88.2} & 89.3 & 88.8 & 83.2 & 89.8 \\
& PT (no FT) & 91.2 & 89.9 & 86.0 & \textbf{93.2} & 92.4 & 92.1 & 60.9 & \textbf{88.1} & 90.2 & 89.2 & 84.1 & 90.5 \\
& PT (FT) & 92.1 & 90.6 & 83.1 & \textbf{93.0} & 90.3 & 90.4 & 58.5 & \textbf{89.7} & 88.0 & 87.9 & 82.4 & \textbf{90.3} \\
& OOWF & 91.1 & 89.6 & 85.6 & 92.1 & 91.5 & 91.1 & 60.5 & \textbf{86.7} & 89.7 & 88.6 & 83.7 & 89.6 \\
& unemployed & 92.9 & 92.7 & 80.3 & \textbf{91.4} & 88.1 & 88.8 & 51.6 & \textbf{87.4} & 85.0 & 87.5 & 79.6 & \textbf{89.6} \\
\midrule
\multirow{2}{*}{\rotatebox[origin=c]{90}{Sex}} &
female & 92.2 & 90.5 & 84.6 & 92.2 & 91.8 & 91.6 & 59.0 & \textbf{87.4} & 89.3 & 88.8 & 83.4 & 90.1 \\
& male & 91.9 & 89.5 & 87.8 & 93.4 & 93.8 & 92.6 & 60.6 & \textbf{86.8} & 92.5 & 90.5 & 85.3 & 90.6 \\
\midrule
\multirow{5}{*}{\rotatebox[origin=c]{90}{Income}} &
fourth 20\% & 92.9 & 90.7 & 87.5 & 93.3 & 94.3 & 93.2 & 60.1 & \textbf{86.9} & 92.0 & 90.3 & 85.4 & 90.9 \\
& middle 20\% & 91.9 & 90.3 & 86.2 & 92.5 & 92.7 & 92.4 & 60.2 & \textbf{87.1} & 90.7 & 89.4 & 84.3 & 90.3 \\
& poorest 20\% & 90.2 & 88.8 & 82.5 & 91.3 & 89.2 & 89.1 & 59.0 & \textbf{87.9} & 86.9 & 87.2 & 81.6 & 88.9 \\
& second 20\% & 91.0 & 89.5 & 85.0 & 91.7 & 91.4 & 91.1 & 60.4 & \textbf{87.3} & 89.4 & 88.4 & 83.4 & 89.6 \\
& top 20\% & 92.5 & 89.1 & 87.5 & 92.5 & 93.6 & 92.7 & 59.5 & \textbf{86.8} & 93.7 & 92.0 & 85.4 & 90.6 \\
\midrule
\multirow{2}{*}{\rotatebox[origin=c]{90}{Area}} &
city/suburb & 94.2 & 91.4 & 85.9 & 92.7 & 93.0 & 93.2 & 56.4 & \textbf{86.3} & 90.4 & 90.3 & 84.0 & 90.8 \\
& rural/town & 90.3 & 89.2 & 86.2 & 92.0 & 92.0 & 91.5 & 62.1 & \textbf{87.8} & 90.2 & 88.5 & 84.2 & 89.8 \\
\bottomrule
\end{tabular}
    \caption{Opinion alignment before ($V$) and after ($V_C$) calibration for WGM demographics using base-prompted, verbalized elicitation. Each pair of columns compares the base-generated distributions to the calibrated distributions (C), with significant differences between the two bolded. The two ``Average'' columns on the right are averages across models. 'FT' means 'full time', 'PT' means 'part time', 'OOWF' means 'out of work force', 'em.' means 'employer', 'se.' means 'self'.}
    \label{tab:individual_dems_wgm_all}
\end{table*}

\begin{table*}[]
    \centering
    \footnotesize
\begin{tabular}{llllllllllll|ll}
\toprule
& \multirow{2}{*}{Demographic} & \multicolumn{2}{l}{Claude-3.5-v2} & \multicolumn{2}{l}{Llama-3.2-90B} & \multicolumn{2}{l}{Mistral-large} & \multicolumn{2}{r}{OLMo-2-7B-I} & \multicolumn{2}{l|}{Qwen-2.5-72B} & \multicolumn{2}{l}{Average} \\
 & & $V$ & $V_C$ & $V$ & $V_C$ & $V$ & $V_C$ & $V$ & $V_C$ & $V$ & $V_C$ & $V$ & $V_C$ \\
\midrule
\multirow{4}{*}{\rotatebox[origin=c]{90}{Age}} & 
18-29 & 92.5 & 92.6 & 87.5 & 88.8 & 86.9 & 88.0 & 68.9 & \textbf{83.2} & 89.7 & 89.2 & 85.1 & 88.4 \\
& 30-49 & 92.6 & 92.5 & 86.5 & 87.8 & 85.9 & 87.1 & 66.6 & \textbf{81.5} & 89.2 & 88.7 & 84.2 & 87.5 \\
& 50-64 & 92.5 & 92.4 & 86.5 & 87.8 & 84.1 & 85.8 & 63.5 & \textbf{78.6} & 88.1 & 87.5 & 82.9 & 86.4 \\
& 65+ & 92.1 & 92.1 & 87.3 & 88.3 & 83.9 & 85.4 & 62.1 & \textbf{76.9} & 87.9 & 87.3 & 82.7 & 86.0 \\
\midrule
\multirow{6}{*}{\rotatebox[origin=c]{90}{Edu.}} & 
AD & 92.5 & 92.4 & 85.9 & 87.4 & 84.6 & 86.5 & 65.3 & \textbf{80.5} & 88.3 & 87.6 & 83.3 & 86.9 \\
& College grad & 92.1 & 92.0 & 87.6 & 88.6 & 85.7 & 87.0 & 65.0 & \textbf{79.8} & 88.6 & 88.0 & 83.8 & 87.1 \\
& HS grad & 91.5 & 91.4 & 85.2 & 86.5 & 83.2 & 84.8 & 63.6 & \textbf{78.5} & 87.8 & 87.2 & 82.3 & 85.7 \\
& < high school & 88.8 & 88.3 & 84.4 & 85.5 & 82.7 & 84.0 & 69.2 & 81.3 & 87.0 & 86.5 & 82.4 & 85.1 \\
& Postgrad & 92.2 & 92.1 & 87.3 & 88.4 & 85.4 & 86.3 & 64.0 & \textbf{78.7} & 88.2 & 87.6 & 83.4 & 86.6 \\
& Some college & 93.7 & 93.7 & 87.0 & 88.4 & 84.9 & 86.5 & 65.2 & \textbf{80.2} & 89.4 & 88.8 & 84.0 & 87.5 \\
\midrule
\multirow{4}{*}{\rotatebox[origin=c]{90}{Region}} & 
Midwest & 92.3 & 92.3 & 86.2 & 87.6 & 84.7 & 86.2 & 64.2 & \textbf{79.3} & 88.3 & 87.7 & 83.1 & 86.6 \\
& Northeast & 92.9 & 92.8 & 88.2 & 89.1 & 85.6 & 87.1 & 64.0 & \textbf{78.7} & 89.1 & 88.6 & 84.0 & 87.3 \\
& South & 92.9 & 92.8 & 87.1 & 88.3 & 84.9 & 86.6 & 65.1 & \textbf{79.9} & 88.9 & 88.4 & 83.8 & 87.2 \\
& West & 92.8 & 92.7 & 87.9 & 89.1 & 86.0 & 87.3 & 65.0 & \textbf{80.1} & 89.1 & 88.5 & 84.2 & 87.5 \\
\midrule
\multirow{5}{*}{\rotatebox[origin=c]{90}{Income}} &
\$100k + & 90.7 & 90.6 & 85.7 & 86.9 & 84.1 & 85.3 & 64.0 & \textbf{78.6} & 86.9 & 86.2 & 82.3 & 85.5 \\
& \$30k-\$50k & 93.2 & 93.1 & 86.3 & 87.6 & 84.9 & 86.5 & 65.1 & \textbf{80.0} & 89.4 & 88.8 & 83.8 & 87.2 \\
& \$50k-\$75k & 93.2 & 93.2 & 87.1 & 88.3 & 85.4 & 86.8 & 64.9 & \textbf{79.7} & 89.1 & 88.5 & 83.9 & 87.3 \\
& \$75k-\$100k & 93.0 & 92.8 & 87.9 & 89.2 & 86.1 & 87.5 & 65.6 & \textbf{80.4} & 89.4 & 88.8 & 84.4 & 87.7 \\
& < \$30k & 93.9 & 93.7 & 88.4 & 89.8 & 86.1 & 87.7 & 65.7 & \textbf{80.7} & 90.2 & 89.5 & 84.9 & 88.3 \\
\midrule
\multirow{5}{*}{\rotatebox[origin=c]{90}{Marital}} &
Divorced & 94.3 & 94.1 & 86.8 & 88.3 & 85.0 & 86.8 & 63.7 & \textbf{78.8} & 89.5 & 88.7 & 83.9 & 87.3 \\
& Married & 92.0 & 91.9 & 86.5 & 87.8 & 84.7 & 86.1 & 64.6 & \textbf{79.5} & 88.3 & 87.7 & 83.2 & 86.6 \\
& Never married & 93.3 & 93.2 & 88.5 & 89.9 & 87.5 & 88.5 & 66.5 & \textbf{81.4} & 89.9 & 89.3 & 85.1 & 88.5 \\
& Separated & 93.8 & 93.6 & 89.2 & 90.3 & 86.4 & 87.6 & 65.4 & \textbf{79.4} & 90.1 & 89.7 & 85.0 & 88.1 \\
& Widowed & 90.1 & 90.0 & 87.5 & 88.4 & 82.9 & 84.1 & 61.5 & \textbf{75.8} & 86.1 & 85.5 & 81.6 & 84.8 \\
\midrule
\multirow{4}{*}{\rotatebox[origin=c]{90}{Pol. Party}} &
Democrat & 90.5 & 91.0 & 89.5 & 89.2 & 85.3 & 85.8 & 62.5 & \textbf{77.3} & 87.9 & 88.4 & 83.1 & 86.3 \\
& Independent & 92.1 & 92.0 & 86.3 & 87.4 & 84.3 & 85.7 & 64.6 & \textbf{79.6} & 88.2 & 87.6 & 83.1 & 86.5 \\
& Other & 91.2 & 91.0 & 84.5 & 86.1 & 84.0 & 85.6 & 65.8 & \textbf{80.9} & 87.5 & 86.7 & 82.6 & 86.1 \\
& Republican & 88.3 & 88.2 & 81.9 & 83.2 & 80.6 & 82.2 & 64.7 & \textbf{79.5} & 85.4 & 84.9 & 80.2 & 83.6 \\
\midrule
\multirow{5}{*}{\rotatebox[origin=c]{90}{Race}} &
Asian & 93.9 & 93.8 & 88.3 & 89.8 & 89.8 & 90.8 & 73.1 & \textbf{86.9} & 92.6 & 92.0 & 87.5 & 90.7 \\
& Black & 93.4 & 93.6 & 91.4 & 91.8 & 87.4 & 88.1 & 65.7 & \textbf{80.3} & 90.5 & 90.0 & 85.7 & 88.8 \\
& Hispanic & 94.3 & 94.3 & 90.3 & 91.2 & 87.6 & 88.6 & 67.1 & \textbf{81.9} & 91.6 & 91.1 & 86.2 & 89.4 \\
& Other & 93.1 & 92.5 & 86.5 & 88.3 & 85.9 & 87.7 & 67.2 & \textbf{81.6} & 90.4 & 89.5 & 84.6 & 87.9 \\
& White & 92.2 & 92.1 & 86.4 & 87.7 & 84.5 & 85.9 & 63.8 & \textbf{78.8} & 88.2 & 87.6 & 83.0 & 86.4 \\
\midrule
\multirow{12}{*}{\rotatebox[origin=c]{90}{Religion}} &
Agnostic & 89.5 & 89.5 & 85.1 & 86.4 & 82.9 & 84.1 & 63.3 & \textbf{77.4} & 85.7 & 84.9 & 81.3 & 84.5 \\
& Atheist & 87.6 & 88.1 & 85.3 & 85.9 & 82.4 & 83.2 & 62.4 & 76.8 & 84.6 & 84.1 & 80.5 & 83.6 \\
& Buddhist & 90.1 & 90.2 & 90.2 & 89.9 & 86.4 & 86.5 & 66.4 & \textbf{81.0} & 88.5 & 88.5 & 84.3 & 87.2 \\
& Hindu & 88.3 & 88.6 & 86.3 & 85.9 & 86.9 & 86.4 & 73.0 & \textbf{85.2} & 87.2 & 87.5 & 84.3 & 86.7 \\
& Jewish & 93.3 & 93.6 & 87.7 & 88.7 & 87.2 & 88.2 & 66.8 & \textbf{80.7} & 89.9 & 89.5 & 85.0 & 88.1 \\
& Mormon & 87.5 & 87.2 & 82.6 & 84.0 & 82.9 & 84.2 & 67.8 & 81.4 & 85.8 & 85.2 & 81.3 & 84.4 \\
& Muslim & 91.9 & 91.6 & 91.2 & 92.3 & 89.4 & 90.7 & 69.9 & \textbf{83.9} & 91.4 & 90.7 & 86.8 & 89.8 \\
& Nothing & 93.2 & 93.1 & 89.2 & 90.3 & 86.6 & 88.0 & 65.1 & \textbf{80.2} & 89.4 & 89.0 & 84.7 & 88.1 \\
& Orthodox & 93.4 & 93.1 & 88.5 & 89.9 & 88.2 & 89.3 & 69.9 & \textbf{84.1} & 91.8 & 91.2 & 86.4 & 89.5 \\
& Other & 92.7 & 92.5 & 86.1 & 87.6 & 84.0 & 85.6 & 63.4 & \textbf{78.5} & 89.5 & 88.9 & 83.1 & 86.6 \\
& Protestant & 91.4 & 91.4 & 85.1 & 86.3 & 83.3 & 85.0 & 63.9 & \textbf{78.8} & 87.3 & 86.8 & 82.2 & 85.7 \\
& Roman Cath. & 93.1 & 93.0 & 87.5 & 88.7 & 85.7 & 87.2 & 66.3 & \textbf{80.8} & 89.3 & 88.7 & 84.4 & 87.7 \\
\midrule
\multirow{2}{*}{\rotatebox[origin=c]{90}{Sex}} &
Female & 93.4 & 93.3 & 87.8 & 89.1 & 85.1 & 86.5 & 63.5 & \textbf{78.4} & 89.2 & 88.7 & 83.8 & 87.2 \\
& Male & 91.5 & 91.4 & 86.1 & 87.6 & 85.4 & 86.6 & 66.2 & \textbf{81.1} & 88.1 & 87.4 & 83.5 & 86.8 \\
\bottomrule
\end{tabular}
    \caption{Opinion alignment before ($V$) and after ($V_C$) calibration for OQA demographics using base-prompted, verbalized elicitation. Each pair of columns compares the base-generated distributions to the calibrated distributions (C), with significant differences between the two bolded. The two ``Average'' columns on the right are averages across models. 'AD' means 'Associate's degree', 'HS' means 'high school'.}
    \label{tab:individual_dems_oqa_all}
\end{table*}

\begin{table*}[]
    \centering
    \footnotesize
\begin{tabular}{llllllllllll|ll}
\toprule
& \multirow{2}{*}{Demographic} & \multicolumn{2}{l}{Claude-3.5-v2} & \multicolumn{2}{l}{Llama-3.2-90B} & \multicolumn{2}{l}{Mistral-large} & \multicolumn{2}{r}{OLMo-2-7B-I} & \multicolumn{2}{l|}{Qwen-2.5-72B} & \multicolumn{2}{l}{Average} \\
 & & $V$ & $V_C$ & $V$ & $V_C$ & $V$ & $V_C$ & $V$ & $V_C$ & $V$ & $V_C$ & $V$ & $V_C$ \\
\midrule
\multirow{6}{*}{\rotatebox[origin=c]{90}{Age}} & 
16-24  & 76.4 & 81.6 & 65.5 & 81.6 & 73.9 & 81.6 & 86.2 & 88.3 & 75.3 & 79.2 & 75.5 & \textbf{82.5} \\
& 25-34  & 76.8 & 81.8 & 65.9 & 81.8 & 74.3 & 81.8 & 86.6 & 88.8 & 75.6 & 79.4 & 75.8 & 82.7 \\
& 35-44  & 76.7 & 81.7 & 65.9 & 81.7 & 74.1 & 81.7 & 87.1 & 88.2 & 75.2 & 79.2 & 75.8 & 82.5 \\
& 45-54  & 76.0 & 81.0 & 65.5 & 81.0 & 73.6 & 81.0 & 87.5 & 87.6 & 74.6 & 78.5 & 75.4 & 81.8 \\
& 55-64  & 74.6 & 79.5 & 64.3 & 79.5 & 72.3 & 79.5 & 87.7 & 85.8 & 73.2 & 77.1 & 74.4 & 80.3 \\
& 65+ & 72.9 & 77.9 & 62.8 & 77.9 & 70.8 & 77.9 & 87.5 & 84.6 & 71.6 & 75.5 & 73.1 & 78.8 \\
\midrule
\multirow{9}{*}{\rotatebox[origin=c]{90}{Education}} &
bachelor & 74.1 & 79.1 & 63.7 & 79.1 & 71.8 & 79.1 & 86.9 & 85.3 & 72.8 & 76.7 & 73.9 & 79.9 \\
& doctoral & 73.3 & 78.8 & 62.6 & 78.8 & 71.0 & 78.8 & 86.0 & 85.7 & 72.5 & 76.4 & 73.1 & 79.7 \\
& early child. & 78.7 & 83.7 & 68.1 & 83.7 & 76.4 & 83.7 & 90.2 & 91.3 & 77.4 & 81.2 & 78.2 & \textbf{84.7} \\
& lower sec. & 75.6 & 80.6 & 65.3 & 80.6 & 73.4 & 80.6 & 87.6 & 87.1 & 74.3 & 78.2 & 75.2 & 81.4 \\
& master & 74.3 & 79.5 & 63.6 & 79.5 & 72.0 & 79.5 & 86.8 & 87.0 & 73.2 & 77.1 & 74.0 & 80.5 \\
& post-sec.& 75.4 & 80.4 & 64.6 & 80.4 & 72.8 & 80.4 & 86.2 & 86.2 & 74.0 & 78.0 & 74.6 & 81.1 \\
& primary & 76.1 & 81.1 & 65.7 & 81.1 & 73.9 & 81.1 & 88.8 & 87.7 & 74.8 & 78.7 & 75.9 & \textbf{81.9} \\
& SC tert. & 73.7 & 78.7 & 63.2 & 78.7 & 71.4 & 78.7 & 85.9 & 85.2 & 72.3 & 76.2 & 73.3 & 79.5 \\
& upper sec. & 75.4 & 80.3 & 64.6 & 80.3 & 72.9 & 80.3 & 86.2 & 86.7 & 74.0 & 77.9 & 74.6 & 81.1 \\
\midrule
\multirow{8}{*}{\rotatebox[origin=c]{90}{Employment}} &
full time & 75.1 & 80.1 & 64.7 & 80.1 & 72.8 & 80.1 & 87.1 & 86.5 & 73.7 & 77.7 & 74.7 & 80.9 \\
& housewife & 75.7 & 80.7 & 65.4 & 80.7 & 73.5 & 80.7 & 87.5 & 87.9 & 74.4 & 78.3 & 75.3 & \textbf{81.7} \\
& other & 74.8 & 79.8 & 64.7 & 79.8 & 72.7 & 79.8 & 87.1 & 83.9 & 73.6 & 77.4 & 74.6 & 80.1 \\
& part time & 76.8 & 81.8 & 66.2 & 81.8 & 74.3 & 81.8 & 87.1 & 88.2 & 75.3 & 79.4 & 75.9 & 82.6 \\
& ret./pen. & 72.9 & 77.9 & 62.6 & 77.9 & 70.7 & 77.9 & 87.1 & 84.8 & 71.6 & 75.4 & 73.0 & 78.8 \\
& self-empl. & 77.0 & 82.0 & 66.0 & 82.0 & 74.4 & 82.0 & 87.4 & 89.6 & 75.7 & 79.6 & 76.1 & \textbf{83.0} \\
& student & 75.6 & 80.8 & 64.9 & 80.8 & 73.1 & 80.8 & 86.1 & 87.0 & 74.4 & 78.4 & 74.8 & 81.6 \\
& unempl. & 78.0 & 83.0 & 67.0 & 83.0 & 75.4 & 83.0 & 86.9 & 89.0 & 76.8 & 80.6 & 76.8 & \textbf{83.7} \\
\midrule
\multirow{7}{*}{\rotatebox[origin=c]{90}{Household size}} &
1 & 74.1 & 79.1 & 63.8 & 79.1 & 72.0 & 79.1 & 87.2 & 85.0 & 72.9 & 76.7 & 74.0 & 79.8 \\
& 2 & 73.5 & 78.5 & 63.3 & 78.5 & 71.4 & 78.5 & 87.0 & 84.6 & 72.2 & 76.1 & 73.5 & 79.2 \\
& 3 & 75.0 & 80.0 & 64.5 & 80.0 & 72.6 & 80.0 & 87.1 & 85.9 & 73.4 & 77.5 & 74.5 & 80.7 \\
& 4 & 75.6 & 80.5 & 65.1 & 80.5 & 73.1 & 80.5 & 86.9 & 86.7 & 74.0 & 78.1 & 74.9 & 81.3 \\
& 5 & 76.5 & 81.4 & 65.5 & 81.4 & 73.9 & 81.4 & 87.0 & 88.8 & 75.2 & 79.0 & 75.6 & \textbf{82.4} \\
& 6 & 76.8 & 81.8 & 65.7 & 81.8 & 74.1 & 81.8 & 87.2 & 89.5 & 75.6 & 79.4 & 75.9 & \textbf{82.9} \\
& 7+ & 78.9 & 86.5 & 66.1 & \textbf{86.5} & 74.6 & \textbf{86.5} & 80.6 & 86.0 & 78.4 & 84.8 & 75.7 & \textbf{86.1} \\
\midrule
\multirow{2}{*}{\rotatebox[origin=c]{90}{Imm.}} &
Immigrant & 71.7 & 76.7 & 61.3 & 76.7 & 69.5 & 76.7 & 86.0 & 83.8 & 70.4 & 74.2 & 71.8 & 77.6 \\
& Native & 76.1 & 81.1 & 65.5 & 81.1 & 73.5 & 81.1 & 87.0 & 87.6 & 74.6 & 78.7 & 75.3 & 81.9 \\
\midrule
\multirow{6}{*}{\rotatebox[origin=c]{90}{Marital}} &
Divorced & 74.2 & 79.2 & 63.9 & 79.2 & 72.0 & 79.2 & 86.7 & 84.8 & 72.9 & 76.8 & 73.9 & 79.8 \\
& Live together& 74.1 & 79.2 & 63.5 & 79.2 & 71.7 & 79.2 & 83.8 & 82.7 & 72.6 & 76.7 & 73.1 & 79.4 \\
& Married & 76.2 & 81.2 & 65.6 & 81.2 & 73.6 & 81.2 & 87.4 & 88.3 & 74.7 & 78.8 & 75.5 & 82.1 \\
& Separated & 74.5 & 79.4 & 63.9 & 79.4 & 72.1 & 79.4 & 84.9 & 83.6 & 73.1 & 77.0 & 73.7 & 79.8 \\
& Single & 75.9 & 81.0 & 65.3 & 81.0 & 73.5 & 81.0 & 86.6 & 87.1 & 74.5 & 78.6 & 75.2 & 81.7 \\
& Widowed & 75.0 & 80.0 & 64.2 & 80.0 & 72.5 & 80.0 & 87.0 & 86.5 & 73.7 & 77.5 & 74.5 & 80.8 \\
\midrule
\multirow{10}{*}{\rotatebox[origin=c]{90}{Religion}} &
Buddhist & 74.8 & 79.8 & 63.8 & 79.8 & 72.1 & 79.8 & 85.8 & 86.1 & 73.5 & 77.3 & 74.0 & 80.6 \\
& Hindu & 79.7 & 86.0 & 68.4 & 86.0 & 76.9 & 86.0 & 85.3 & 93.4 & 79.6 & 83.6 & 78.0 & \textbf{87.0} \\
& Jew & 76.3 & 81.5 & 65.2 & 81.5 & 73.6 & 81.5 & 88.0 & 87.9 & 75.2 & 79.0 & 75.7 & \textbf{82.3} \\
& Muslim & 76.9 & 81.8 & 66.1 & 81.8 & 74.4 & 81.8 & 89.5 & 90.5 & 75.6 & 79.4 & 76.5 & \textbf{83.1} \\
& Orthodox & 73.1 & 78.6 & 62.3 & 78.6 & 70.7 & 78.6 & 85.5 & 85.8 & 72.4 & 76.2 & 72.8 & 79.6 \\
& Other & 70.8 & 75.7 & 61.4 & 75.7 & 68.3 & 75.7 & 85.7 & 80.1 & 69.2 & 73.3 & 71.1 & 76.1 \\
& Other Christ. & 71.6 & 76.6 & 61.5 & 76.6 & 69.5 & 76.6 & 84.0 & 80.6 & 70.3 & 74.2 & 71.4 & 76.9 \\
& Protestant & 74.3 & 79.3 & 63.7 & 79.3 & 71.9 & 79.3 & 85.3 & 83.9 & 72.9 & 76.9 & 73.6 & 79.7 \\
& Roman Cath. & 73.9 & 78.9 & 63.4 & 78.9 & 71.6 & 78.9 & 85.2 & 83.8 & 72.6 & 76.5 & 73.3 & 79.4 \\
& none & 70.6 & 75.6 & 61.0 & 75.6 & 68.5 & 75.6 & 84.9 & 81.1 & 69.3 & 73.2 & 70.9 & 76.2 \\
\midrule
\multirow{2}{*}{\rotatebox[origin=c]{90}{Sex}} &
Female & 75.2 & 80.2 & 64.3 & 80.2 & 72.5 & 80.2 & 86.1 & 86.5 & 73.8 & 77.7 & 74.4 & 81.0 \\
& Male & 76.3 & 81.2 & 65.9 & 81.2 & 73.9 & 81.2 & 87.8 & 87.7 & 74.8 & 78.8 & 75.7 & 82.0 \\
\bottomrule
\end{tabular}
    \caption{Opinion alignment before ($V$) and after ($V_C$) calibration for WVS demographics using base-prompted, verbalized elicitation. Each pair of columns compares the base-generated distributions to the calibrated distributions (C), with significant differences between the two bolded. The two ``Average'' columns on the right are averages across models. 'child.' means 'childhood', 'sec.' means 'secondary', 'SC tert.' means 'short-cycle tertiary', 'ret./pen.' means 'retired/pensioned', 'empl' means 'employed'.}
    \label{tab:individual_dems_wvs_all}
\end{table*}

\subsection{Out-of-domain generalization}
\label{sec:appendix_generalization_results}
We study whether our regression models for calibration generalize to unseen \textit{datasets}, and whether calibrated distributions are more aligned than the original LLM-generated distributions for the unseen dataset. We train regression models on two of our three datasets and evaluate on the held-out dataset. 
Results are shown in Tables \ref{tab:regression_full_generalization} and \ref{tab:log_prob_generalization_results}. 
As expected, opinion alignment is lower on  unseen dataset, but the distributions calibrated on out-of-distribution data are typically more aligned with human responses than the original LLM-generations of that dataset. Alignment is higher in 92.8\% of settings for the unseen OQA dataset, 90.6\% of settings for WVS, and 85.5\% for WGM. This suggests that regression models trained on these datasets could generalize to data, though it would be better to collect a few supervised examples in-domain to train regression models.

\begin{table*}[]
    \centering
    \footnotesize
    \addtolength{\tabcolsep}{-0.2em}
\begin{tabular}{cc|rrrrrr|rrrrrr}
\toprule
& \multirow{2}{*}{Model} &  \multicolumn{6}{c}{Base prompt} & \multicolumn{6}{c}{Sociodemographic prompt} \\
& & $P$ & \!$P_C$ & $S$ & \!$S_C$ & $V$ & \!$V_C$ & $P$ & \!$P_C$ & $S$ & \!$S_C$ & $V$ & \!$V_C$ \\
\midrule
\multirow{15}{*}{\rotatebox[origin=c]{90}{WGM}}
& OLMo-2-7B-Base & 65.6 & \textbf{82.4} & 79.2 & {81.8} & 82.2 & {82.6} & 74.8 & \textbf{81.9} & 71.0 & \textbf{82.1} & 74.9 & \textbf{82.6} \\
& OLMo-2-7B-SFT & 80.6 & {81.8} & 70.3 & \textbf{81.3} & 68.6 & \textbf{83.9} & 75.6 & \textbf{82.4} & 70.2 & \textbf{82.2} & 65.4 & \textbf{82.7} \\
& OLMo-2-7B-DPO & 71.1 & \textbf{81.7} & 63.2 & \textbf{82.1} & 63.9 & \textbf{83.1} & 72.2 & \textbf{82.7} & 68.7 & \textbf{82.4} & 63.8 & \textbf{83.0} \\
& OLMo-2-7B-Instruct & 59.6 & \textbf{82.0} & 67.7 & \textbf{81.8} & 59.8 & \textbf{82.8} & 62.3 & \textbf{81.7} & 64.5 & \textbf{82.2} & 69.8 & \textbf{84.4} \\
& Llama-3-70B & 64.9 & \textbf{83.6} & 73.7 & \textbf{81.8} & 84.5 & {86.5}  & 66.5 & \textbf{84.6} & 61.7 & \textbf{84.1} & 84.4 & {87.1} \\
& Llama-3.1-70B & 73.2 & \textbf{82.7} & 70.3 & \textbf{84.4} & 80.6 & {81.8} & 68.5 & \textbf{85.2} & 66.1 & \textbf{81.8} & 85.2 & {87.9} \\
& Llama-3.2-1B & {81.8} & {81.8} & {81.8} & {81.8} & -- & -- & 61.9 & \textbf{82.3} & 46.7 & \textbf{80.7} & 61.4 & {86.4} \\
& Llama-3.2-11B  & 73.7 & \textbf{85.9} & 73.2 & \textbf{81.4} & 73.6 & \textbf{82.1} & 71.8 & \textbf{84.5} & 72.0 & \textbf{84.1} & 71.7 & {81.1} \\
& Llama-3.2-90B & 68.1 & \textbf{84.1} & 73.0 & \textbf{84.4} & 84.8 & {86.2} & 70.6 & \textbf{84.7} & 67.6 & \textbf{84.4} & 85.0 & {86.4} \\
& Qwen-2.5-72B & 57.7 & \textbf{83.5} & 53.3 & \textbf{83.0} & {88.2} & 86.7 & 66.0 & \textbf{81.8} & 63.4 & \textbf{83.7} & \textbf{89.1} & 85.9 \\
& Mistral-small & 61.6 & \textbf{84.1} & 61.6 & \textbf{83.7} & \textbf{88.9} & 85.7 & 67.5 & \textbf{84.6} & 64.4 & \textbf{83.9} & \textbf{89.0} & 85.0 \\
& Mistral-large & 62.4 & \textbf{84.4} & 72.0 & \textbf{83.5} & {89.4} & 86.8 & 68.4 & \textbf{84.6} & 63.0 & \textbf{84.2} & {87.3} & 86.2 \\
& Claude-3 & 57.8 & \textbf{82.2} & 50.0 & \textbf{81.8} & {84.8} & 84.3 & 64.9 & \textbf{82.2} & 52.9 & \textbf{81.6} & {88.4} & 85.6 \\
& Claude-3.5-v1 & 64.5 & \textbf{83.9} & 62.0 & \textbf{83.5} & {86.1} & 84.9 & 65.8 & \textbf{84.7} & 62.3 & \textbf{84.5} & 86.3 & {88.1} \\
& Claude-3.5-v2 & 66.2 & \textbf{84.3} & 59.5 & \textbf{83.1} & {89.3} & 86.5 & 65.4 & \textbf{84.3} & 61.4 & \textbf{84.1} & {89.0} & 87.7 \\
\midrule
& Average & 67.3 & \textbf{83.2} & 67.4 & \textbf{82.6} & 80.3 & \textbf{84.6} & 68.1 & \textbf{83.5} & 63.7 & \textbf{83.1} & 79.4 & \textbf{85.3}\\
& Std Dev & 7.5 & 1.2 & 9.0 & 1.1 & 9.9 & 1.8 & 4.1 & 1.3 & 6.7 & 1.2 & 10.3 & 2.2 \\
\bottomrule

\multirow{15}{*}{\rotatebox[origin=c]{90}{OQA}}
& OLMo-2-7B-Base & 85.0 & \textbf{89.1} & 82.6 & \textbf{88.4} & 82.1 & \textbf{88.0} & 80.8 & \textbf{88.3} & 77.2 & \textbf{88.3} & 83.8 & \textbf{88.4} \\
& OLMo-2-7B-SFT & 78.8 & \textbf{88.4} & 79.4 & \textbf{88.4} & 68.7 & \textbf{83.4} & 80.7 & \textbf{88.3} & 77.8 & \textbf{88.4} & 72.6 & \textbf{84.2} \\
& OLMo-2-7B-DPO & 79.3 & \textbf{88.4} & 81.5 & \textbf{88.3} & 69.3 & \textbf{83.6} & 82.1 & \textbf{88.2} & 78.7 & \textbf{88.3} & 72.3 & \textbf{85.4} \\
& OLMo-2-7B-Instruct & 72.6 & \textbf{88.6} & 72.3 & \textbf{88.4} & 65.4 & \textbf{82.5} & 72.8 & \textbf{88.5} & 70.5 & \textbf{88.6} & 68.5 & \textbf{81.4} \\
& Llama-3-70B  & 72.2 & \textbf{88.4} & 76.8 & \textbf{88.4} & 81.8 & \textbf{88.4} & 74.4 & \textbf{88.7} & 70.1 & \textbf{89.0} & 79.5 & \textbf{88.8} \\
& Llama-3.1-70B & 73.5 & \textbf{88.9} & 70.5 & \textbf{88.9} & 83.8 & \textbf{88.4}  & 76.4 & \textbf{89.1} & 71.7 & \textbf{90.4} & 83.6 & \textbf{89.6} \\
& Llama-3.2-1B  & 82.3 & \textbf{88.4} & {88.4} & {88.4} & -- & -- & 72.6 & \textbf{89.1} & 70.2 & \textbf{88.5} & 83.6 & {87.6} \\
& Llama-3.2-11B  & 84.3 & \textbf{88.4} & 75.5 & \textbf{87.7} & 65.8 & \textbf{88.2}  & 75.7 & \textbf{87.8} & 74.8 & \textbf{87.7} & 83.0 & {88.7} \\
& Llama-3.2-90B & 79.3 & \textbf{89.4} & 75.3 & \textbf{88.4} & 86.8 & {88.4} & 79.2 & \textbf{88.9} & 76.1 & \textbf{89.0} & 83.4 & \textbf{88.8} \\
& Qwen-2.5-72B & 73.9 & \textbf{88.4} & 67.0 & \textbf{88.5} & 88.4 & {88.9} & 74.4 & \textbf{88.4} & 71.3 & \textbf{88.4} & {89.2} & 89.1 \\
& Mistral-small & 77.3 & \textbf{88.4} & 73.6 & \textbf{88.4} & 86.9 & {88.4} & 75.1 & \textbf{88.4} & 71.6 & \textbf{88.4} & 87.8 & {89.5} \\
& Mistral-large & 79.5 & \textbf{88.8} & 75.3 & \textbf{88.4} & 85.0 & \textbf{88.8} & 75.8 & \textbf{88.9} & 72.4 & \textbf{88.4} & 83.8 & \textbf{87.5} \\
& Claude-3 & 73.5 & \textbf{88.4} & 66.4 & \textbf{88.4} & 87.6 & {88.9} & 76.4 & \textbf{88.4} & 72.6 & \textbf{88.5} & {89.3} & {89.3} \\
& Claude-3.5-v1 & 70.5 & \textbf{88.4} & 69.6 & \textbf{88.4} &{89.4} & 88.4 & 76.1 & \textbf{89.5} & 72.8 & \textbf{89.7} & 91.0 & {91.5} \\
& Claude-3.5-v2 & 70.5 & \textbf{88.4} & 72.5 & \textbf{88.4} & \textbf{91.7} & 89.2 & 76.1 & \textbf{89.2} & 73.3 & \textbf{89.4} & {91.9} & 90.8 \\
\midrule
& Average & 76.8 & \textbf{88.6} & 75.1 & \textbf{88.4} & 80.9 & \textbf{87.4} & 76.6 & \textbf{88.6} & 73.4 & \textbf{88.7} & 82.9 & \textbf{88.0}\\
& Std Dev & 4.8 & 0.3 & 6.0 & 0.2 & 9.4 & 2.3 & 2.9 & 0.5 & 2.8 & 0.7 & 7.0 & 2.6\\
\bottomrule

\multirow{15}{*}{\rotatebox[origin=c]{90}{WVS}}
& OLMo-2-7B-Base & 74.5 & \textbf{81.2} & 70.7 & \textbf{80.3} & {84.5} & 84.4 & 78.2 & {81.1} & 78.2 & {81.0} & 80.8 & {83.0} \\
& OLMo-2-7B-SFT & 69.7 & \textbf{79.1} & 68.1 & \textbf{79.2} & {87.2} & 85.6 & 72.9 & \textbf{80.3} & 73.1 & \textbf{80.8} & 84.9 & {87.0} \\
& OLMo-2-7B-DPO & 66.2 & \textbf{78.3} & 70.5 & \textbf{79.7} & \textbf{89.1} & 82.8 & 70.6 & \textbf{80.3} & 70.2 & \textbf{80.8} & 82.5 & {84.3} \\
& OLMo-2-7B-Instruct & 75.3 & \textbf{80.9} & 74.9 & \textbf{80.5} & 86.6 & {87.6} & 58.3 & \textbf{78.1} & 60.0 & \textbf{76.9} & 86.0 & {88.6} \\
& Llama-3-70B  & 59.7 & \textbf{76.3} & 61.7 & \textbf{79.3} & 81.3 & {82.9}  & 67.9 & \textbf{79.2} & 62.8 & \textbf{79.6} & 78.1 & {79.4} \\
& Llama-3.1-70B  & 61.9 & \textbf{76.8} & 56.4 & \textbf{75.8} & 65.2 & {70.7}  & 63.6 & \textbf{78.0} & 57.6 & \textbf{76.7} & 68.3 & {72.1} \\
& Llama-3.2-1B  & {80.3} & {80.3} & {80.3} & {80.3} & -- & -- & 68.7 & \textbf{82.5} & 66.6 & \textbf{79.0} & -- & -- \\
& Llama-3.2-11B  & 75.4 & \textbf{81.9} & 76.0 & \textbf{84.4} & 82.7 & {83.6}  & 78.2 & \textbf{83.2} & 78.0 & {81.9} & 66.5 & {77.0} \\
& Llama-3.2-90B & 61.6 & \textbf{76.8} & 59.1 & \textbf{75.4} & 64.6 & \textbf{70.2} & 62.1 & \textbf{77.7} & 59.5 & \textbf{76.5} & 67.7 & {71.5} \\
& Qwen-2.5-72B & 39.6 & \textbf{71.7} & 42.4 & \textbf{74.3} & 74.0 & {74.3} & 49.1 & \textbf{71.8} & 49.4 & \textbf{74.7} & 73.4 & {73.8} \\
& Mistral-small & 42.9 & \textbf{73.0} & 46.3 & \textbf{74.4} & 75.0 & {75.8} & 48.0 & \textbf{72.5} & 46.4 & \textbf{74.6} & 68.6 & {71.0} \\
& Mistral-large & 48.8 & \textbf{74.2} & 44.0 & \textbf{76.4} & 72.8 & {74.8} & 54.3 & \textbf{74.4} & 51.5 & \textbf{76.4} & 76.6 & {78.3} \\
& Claude-3 & 47.3 & \textbf{71.4} & 55.7 & \textbf{76.3} & {74.1} & 74.0 & 55.3 & \textbf{74.8} & 57.3 & \textbf{77.1} & 73.2 & {73.3} \\
& Claude-3.5-v1 & 44.3 & \textbf{71.2} & 49.6 & \textbf{74.6} & 75.7 & {77.7} & 58.7 & \textbf{76.7} & 54.8 & \textbf{78.1} & 73.3 & {75.5} \\
& Claude-3.5-v2 & 46.5 & \textbf{72.1} & 51.0 & \textbf{75.5} & {75.2} & {75.2} & 61.0 & \textbf{78.2} & 56.8 & \textbf{78.4} & {75.6} & {75.6} \\
\midrule
& Average & 59.6 & \textbf{76.3} & 60.4 & \textbf{77.8} & 77.7 & \textbf{78.5} & 63.1 & \textbf{77.9} & 61.5 & \textbf{78.2} & 75.4 & \textbf{77.9}\\
& Std Dev & 13.8 & 3.9 & 12.4 & 3.0 & 7.7 & 5.8 & 9.5 & 3.4 & 9.9 & 2.3 & 6.4 & 5.8 \\
\bottomrule
\end{tabular}
\caption{Opinion alignment before and after calibration for each dataset, LLM, and elicitation method, \textbf{when training on two datasets and evaluating on the out-of-domain dataset}. Each pair of columns compares the base-generated or SD-generated distributions to the calibrated distributions ($C$) for each elicitation method: paraphrase (`P'), self-random (`S'), and verbalized (`V'). Bolded values are significant between each pair. The mean and standard deviation across models are shown in the bottom rows of each dataset section.}
\label{tab:regression_full_generalization}
\end{table*}

\begin{table}[h]
\centering
\footnotesize
\begin{tabular}{cc|rr|rr}
\toprule
& \multirow{2}{*}{Model} &  \multicolumn{2}{c}{Base prompt} & \multicolumn{2}{c}{SD prompt} \\
& & $L$ & \!$L_C$ & $L$ & \!$L_C$ \\
\midrule
\multirow{4}{*}{\rotatebox[origin=c]{90}{WGM}} 
& Llama-3-70B & 73.1 & \textbf{84.6} & 67.5 & \textbf{86.1}\\
& Llama-3.1-70B & 74.0 & \textbf{81.8} & 70.4 & \textbf{86.1}\\
& Llama-3.2-1B & {83.6} & 83.5 & 83.0 & {83.1}\\
& Llama-3.2-11B & 84.0 & {84.3} & {82.2} & 81.8\\
\midrule
& Average & 78.7 & \textbf{83.5} & 75.8 & \textbf{84.3}\\
& Std Dev & 5.9 & 1.3 & 8.0 & 2.2\\
\bottomrule
\multirow{4}{*}{\rotatebox[origin=c]{90}{OQA}} 
& Llama-3-70B & 75.4 & \textbf{88.4} & 72.3 & \textbf{88.5}\\
& Llama-3.1-70B & 78.9 & \textbf{88.4} & 75.7 & \textbf{88.4}\\
& Llama-3.2-1B  & \textbf{88.3} & 85.2 & {87.3} & 86.6\\
& Llama-3.2-11B &  82.6 & \textbf{88.3} & 88.3 & {88.4}\\
\midrule
& Average & 81.3 & \textbf{87.6} & 80.9 & \textbf{88.0}\\
& Std Dev & 5.5 & 1.6 & 8.1 & 0.9 \\
\bottomrule
\multirow{4}{*}{\rotatebox[origin=c]{90}{WVS}} 
& Llama-3-70B  & 54.1 & \textbf{74.8} & 51.1 & \textbf{72.1} \\
& Llama-3.1-70B   & 43.5 & \textbf{71.3} & 49.1 & \textbf{70.8}\\
& Llama-3.2-1B  & {83.9} & 82.1 & {85.1} & 84.6\\
& Llama-3.2-11B   &  74.8 & \textbf{78.3} & 72.9 & {75.9}\\
\midrule
& Average & 64.1 & \textbf{76.6} & 64.5 & \textbf{75.8}\\
& Std Dev & 18.5 & 4.6 & 17.4 & 6.2\\
\bottomrule
\end{tabular}
\caption{Opinion alignment before and after calibration for each dataset and LLM using log probability distributions, \textbf{when training on two datasets and evaluating on the out-of-domain dataset}. Each pair of columns compares the base-generated or SD-generated distributions to the calibrated distributions ($C$) for log probabilities (`L'). Bolded values are significant between each pair. The mean and standard deviation across models are shown in the bottom rows of each dataset section.}
\label{tab:log_prob_generalization_results}
\end{table}

\section{Extracted distributions}
The percent of distributions extracted from each LLM for each dataset and elicitation method is shown in Table \ref{tab:num_extracted_distributions}.

\begin{table}[h]
    \centering
    \footnotesize
    \addtolength{\tabcolsep}{-0.3em}
    \begin{tabular}{cl|rrr|rrr}
\toprule
& \multirow{2}{*}{Model} &  \multicolumn{3}{c}{Base prompt} & \multicolumn{3}{c}{SD prompt} \\
& & $P$  & $S$  & $V$  & $P$  & $S$ & $V$ \\
\midrule
\multirow{15}{*}{\rotatebox[origin=c]{90}{WGM}}
& Claude-3 & 100 & 100 & 100 & 100 & 100 & 100 \\
& Claude-3.5-v1 & 100 & 100 & 100 & 100 & 100 & 100 \\
& Claude-3.5-v2 & 100 & 100 & 100 & 100 & 100 & 100 \\
& Llama-3-70B & 100 & 100 & 100 & 100 & 100 & 100 \\
& Llama-3.1-70B & 100 & 100 & 82.4 & 100 & 100 & 92.8 \\
& Llama-3.2-1B & 100 & 100 & 0 & 100 & 100 & 1.3 \\
& Llama-3.2-11B & 100 & 100 & 70.6 & 100 & 100 & 27.1 \\
& Llama-3.2-90B & 100 & 100 & 100 & 100 & 100 & 94.9 \\
& Mistral-small & 100 & 100 & 100 & 100 & 100 & 99.7 \\
& Mistral-large & 100 & 100 & 100 & 100 & 100 & 100 \\
& OLMo-2-7B-B & 100 & 100 & 47.1 & 100 & 100 & 69.3 \\
& OLMo-2-7B-S & 100 & 100 & 100 & 100 & 100 & 97.6 \\
& OLMo-2-7B-D & 100 & 100 & 100 & 100 & 100 & 92.9 \\
& OLMo-2-7B-I & 100 & 100 & 70.6 & 100 & 100 & 89.1 \\
& Qwen-2.5-72B & 100 & 100 & 100 & 100 & 100 & 100 \\
\midrule
\multirow{15}{*}{\rotatebox[origin=c]{90}{OQA}}
& Claude-3 & 100 & 100 & 100 & 100 & 100 & 100 \\
& Claude-3.5-v1 & 100 & 100 & 100 & 100 & 100 & 100 \\
& Claude-3.5-v2 & 100 & 100 & 100 & 100 & 100 & 100 \\
& Llama-3-70B & 100 & 100 & 91.9 & 100 & 100 & 99.8 \\
& Llama-3.1-70B & 100 & 100 & 100 & 100 & 12.1 & 93.4 \\
& Llama-3.2-1B & 100 & 100 & 0 & 100 & 100 & 0.6 \\
& Llama-3.2-11B & 100 & 100 & 32.4 & 100 & 100 & 7.2 \\
& Llama-3.2-90B & 100 & 100 & 100 & 100 & 100 & 99.7 \\
& Mistral-small & 100 & 100 & 100 & 100 & 100 & 100 \\
& Mistral-large & 100 & 100 & 100 & 100 & 100 & 99.7 \\
& OLMo-2-7B-B & 100 & 100 & 100 & 100 & 100 & 99.9 \\
& OLMo-2-7B-S & 100 & 100 & 100 & 100 & 100 & 99.5 \\
& OLMo-2-7B-D & 100 & 100 & 94.6 & 100 & 100 & 79.8 \\
& OLMo-2-7B-I & 100 & 100 & 75.7 & 100 & 100 & 79.5 \\
& Qwen-2.5-72B & 100 & 100 & 100 & 100 & 100 & 99.4 \\
\midrule
\multirow{15}{*}{\rotatebox[origin=c]{90}{WVS}}
& Claude-3 & 100 & 100 & 100 & 100 & 100 & 100 \\
& Claude-3.5-v1 & 100 & 100 & 100 & 19.6 & 19.6 & 19.6 \\
& Claude-3.5-v2 & 100 & 100 & 100 & 49 & 49 & 49 \\
& Llama-3-70B & 100 & 100 & 97.9 & 99.4 & 99 & 87.2 \\
& Llama-3.1-70B & 100 & 100 & 100 & 49 & 49 & 47.2 \\
& Llama-3.2-1B & 100 & 100 & 0 & 49 & 49 & 0.4 \\
& Llama-3.2-11B & 100 & 100 & 39.7 & 49 & 49 & 3.7 \\
& Llama-3.2-90B & 100 & 100 & 100 & 100 & 100 & 98.3 \\
& Mistral-small & 100 & 100 & 100 & 100 & 100 & 100 \\
& Mistral-large & 100 & 100 & 100 & 100 & 100 & 99.9 \\
& OLMo-2-7B-B & 100 & 100 & 100 & 100 & 100 & 99.6 \\
& OLMo-2-7B-S & 100 & 100 & 100 & 100 & 100 & 99.3 \\
& OLMo-2-7B-D & 100 & 100 & 76.8 & 100 & 100 & 72.2 \\
& OLMo-2-7B-I & 100 & 100 & 82.1 & 100 & 100 & 79.3 \\
& Qwen-2.5-72B & 100 & 100 & 100 & 100 & 100 & 98.6 \\
\bottomrule
\end{tabular}
    \caption{The percent of extracted distributions in each setting. We found that smaller models (especially Llama 3.2 1B and 11B) would often refuse to answer questions (question would hit safety restrictions for role-playing as a particular identity group), or did not follow instructions in formatting output distributions.}
    \label{tab:num_extracted_distributions}
\end{table}

\section{Minimal supervision training plots}
\label{sec:appendix_minimal_supervision_plots}
\begin{figure}[]
    \centering
    \vspace{-1em}
    \subfigure[]{\includegraphics[width=0.4\textwidth]{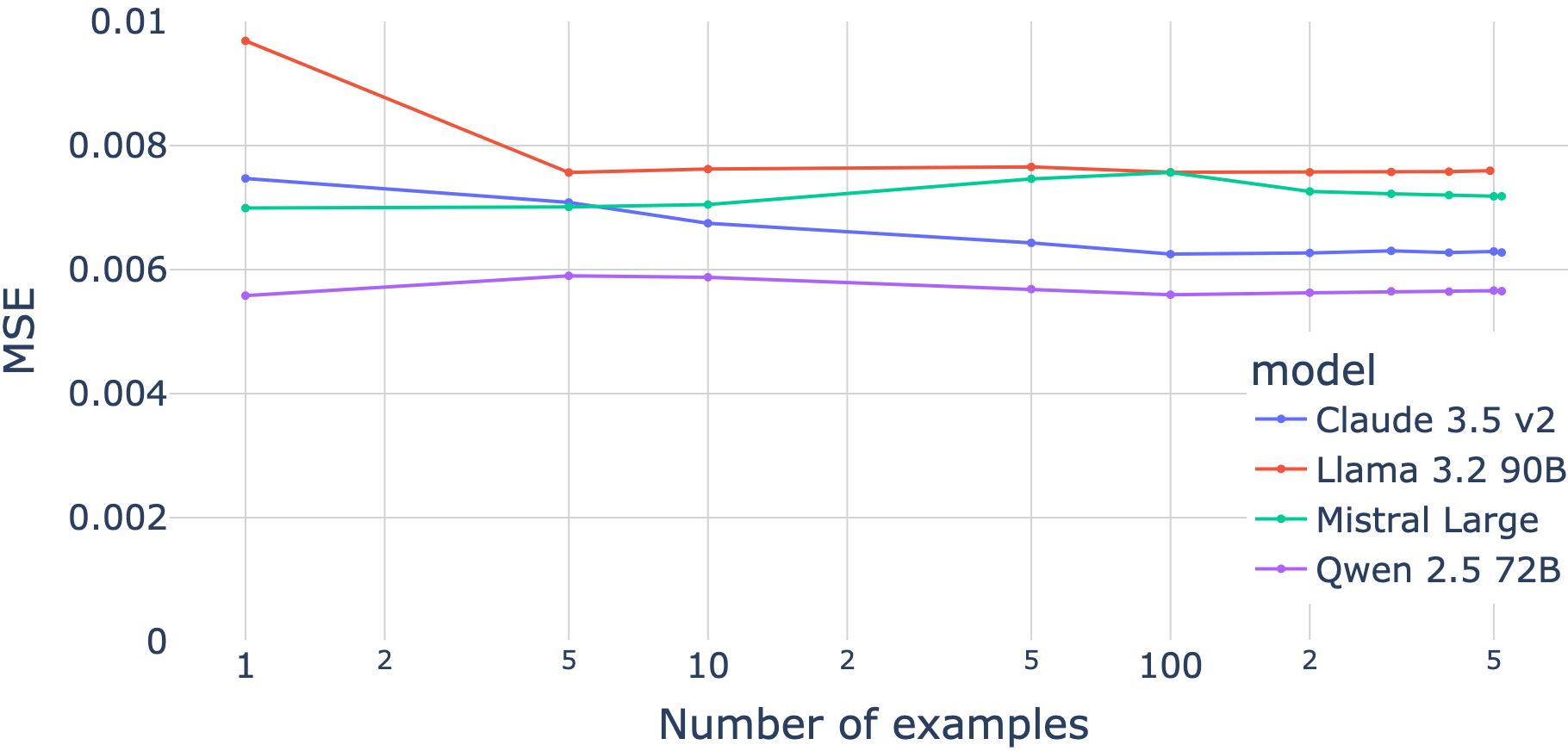}} 
    \subfigure[]{\includegraphics[width=0.4\textwidth]{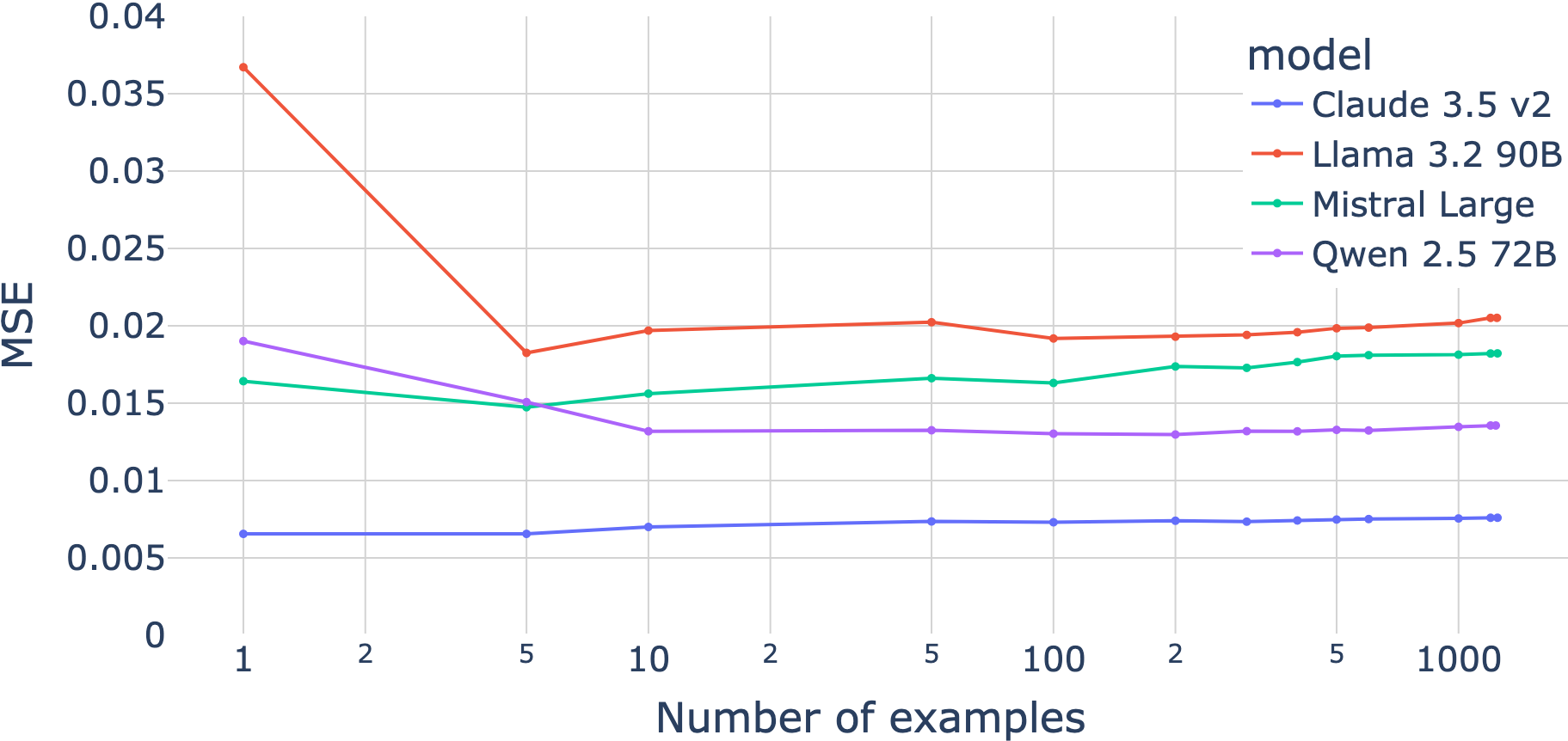}} 
    \subfigure[]{\includegraphics[width=0.4\textwidth]{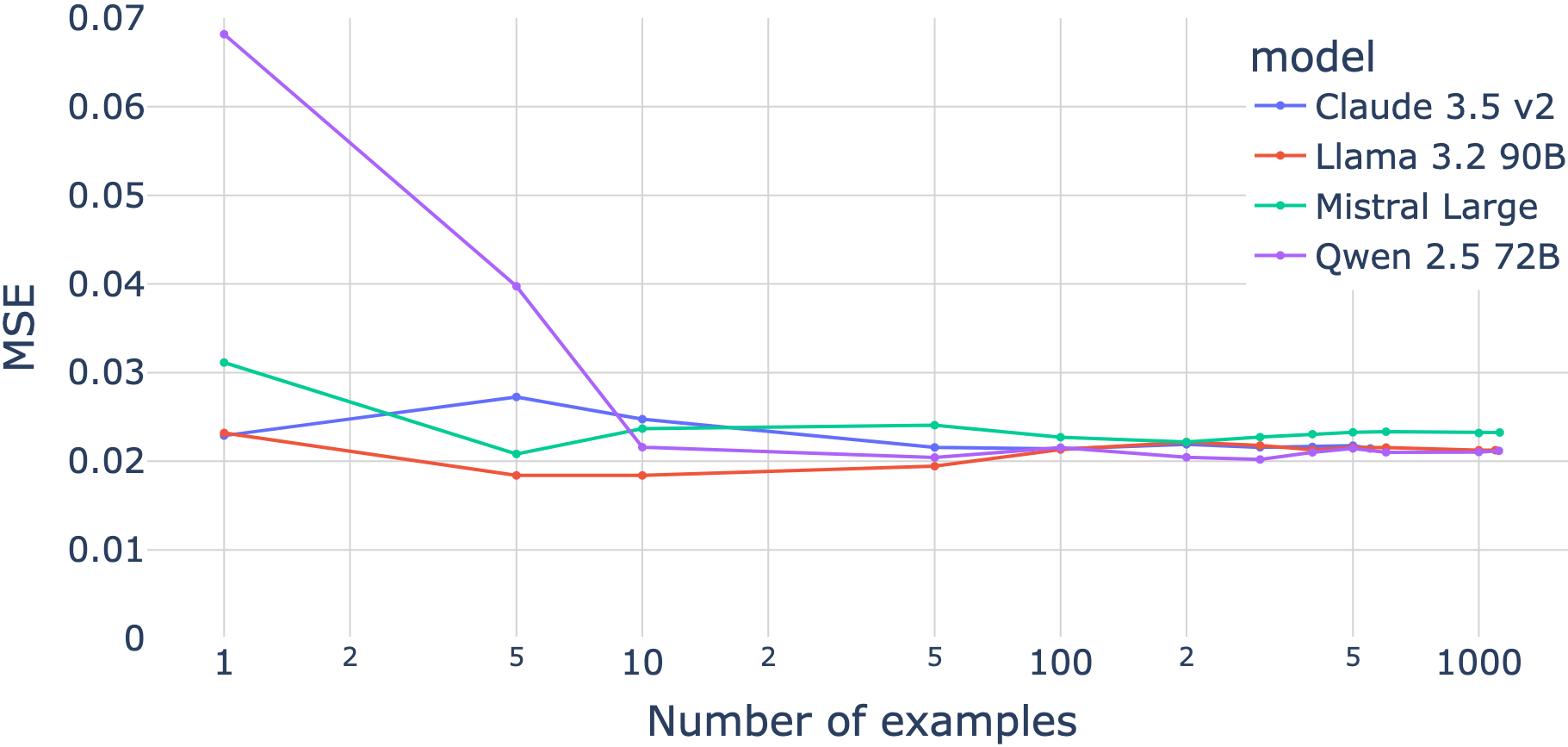}}
    \vspace{-1em}
    \caption{Mean Squared Error (MSE) of regression models on various training data sizes, using SD prompted and verbally elicited distributions. Plots are shown for each dataset: (a) WGM, (b) OQA, and (c) WVS. \textbf{Although model and dataset dependent, MSE converges between 1 and 10 examples.}}
    \label{fig:minimal_supervision_plots}
    \vspace{-1em}
\end{figure}

We plot Mean Squared Error (MSE) over training data size in Figure \ref{fig:minimal_supervision_plots}, showing that MSE usually converges between 1 and 10 examples, though this is model and dataset dependent.

\section{Statistical significance details}
\label{sec:appendix_stat_sig}
We list the t-test statistic and p-value after Bonferroni correction in Table \ref{tab:stat_sig_details} for our main results (Tables \ref{tab:regression_all_datasets} and \ref{tab:regression_full}). Our Bonferroni alpha is 0.000172, calculated using the Python Statsmodels\footnote{\url{https://www.statsmodels.org/stable/index.html}} package. Our results are calculated on the 10\% test split discussed in Section \ref{sec:calibration}. While the number of examples is different for each of the 270 settings, the average number of examples is approximately 237. While we do not conduct explicit tests for normality for using the t-test, the t-test is remarkably robust to violations in sample normality \cite{smucker_comparison_2007}, particularly given the number of examples.

\begin{table*}[]
    \centering
    \footnotesize
    \addtolength{\tabcolsep}{-0.35em}
    \begin{tabular}{ll|rrrrrr|rrrrrr}
\toprule
& \multirow{2}{*}{Model} &  \multicolumn{6}{c}{t-test statistic} & \multicolumn{6}{c}{Corrected p-value}\\
& & $P_B$ & \!$P_S$ & $S_B$ & \!$S_S$ & $V_B$ & \!$V_S$ & $P_B$ & \!$P_S$ & $S_B$ & \!$S_S$ & $V_B$ & \!$V_S$\\
\midrule
\multirow{15}{*}{\rotatebox[origin=c]{90}{WGM}} 
& O-2-7B-B & 11.76 & 5.75 & 3.07 & 7.76 & 1.21 & 4.99 & 2.70e-24 & 6.04e-06 & 6.74e-01 & 3.33e-11 & 1.00 & 3.59e-04 \\
& O-2-7B-S & 1.18 & 6.01 & 11.11 & 9.34 & 15.54 & 13.24 & 1.00 & 1.50e-06 & 5.57e-22 & 5.00e-16 & 1.91e-38 & 1.36e-29 \\
& O-2-7B-D & 4.20 & 8.99 & 18.04 & 10.32 & 16.17 & 14.32 & 1.01e-02 & 6.41e-15 & 3.94e-48 & 2.84e-19 & 6.95e-41 & 2.33e-33 \\
& O-2-7B-I & 16.07 & 12.93 & 13.79 & 13.32 & 17.65 & 9.12 & 1.79e-40 & 1.67e-28 & 9.17e-32 & 5.70e-30 & 1.14e-42 & 4.05e-15 \\
& L-3-70B & 20.48 & 19.53 & 8.47 & 22.37 & 5.37 & 5.07 & 1.39e-57 & 6.64e-54 & 2.66e-13 & 1.76e-64 & 4.51e-05 & 1.99e-04 \\
& L-3.1-70B & 16.44 & 18.58 & 16.76 & 21.40 & 5.08 & 5.79 & 6.50e-42 & 3.15e-50 & 3.56e-43 & 4.20e-61 & 1.90e-04 & 5.20e-06 \\
& L-3.2-1B & 0.00 & 12.97 & 0.00 & 11.94 & -- & 3.25 & 1.00 & 1.48e-28 & 1.00 & 1.14e-19 & -- & 1.00 \\
& L-3.2-11B & 15.13 & 13.25 & 8.32 & 12.94 & 7.41 & 2.52 & 7.45e-37 & 9.27e-30 & 7.71e-13 & 1.34e-28 & 6.34e-10 & 1.00 \\
& L-3.2-90B & 16.71 & 17.13 & 15.74 & 19.81 & 5.38 & 6.32 & 5.59e-43 & 1.38e-44 & 3.23e-39 & 5.30e-55 & 4.13e-05 & 2.63e-07 \\
& Q-2.5-72B & 12.96 & 17.09 & 16.79 & 20.88 & -1.12 & 0.42 & 1.16e-28 & 1.90e-44 & 2.70e-43 & 3.97e-59 & 1.00 & 1.00 \\
& M-small & 25.53 & 21.12 & 27.01 & 22.72 & -0.08 & 0.02 & 1.65e-76 & 4.81e-60 & 7.12e-82 & 4.28e-66 & 1.00 & 1.00 \\
& M-large & 24.04 & 16.25 & 10.43 & 23.27 & -0.71 & 1.33 & 4.76e-71 & 3.48e-41 & 1.27e-19 & 7.47e-68 & 1.00 & 1.00 \\
& C-3 & 17.07 & 11.82 & 17.69 & 19.29 & 1.52 & -0.44 & 2.31e-44 & 1.65e-24 & 8.75e-47 & 5.68e-53 & 1.00 & 1.00 \\
& C-3.5-v1 & 18.09 & 18.54 & 22.78 & 21.61 & 1.57 & 2.54 & 2.47e-48 & 4.67e-50 & 2.43e-66 & 6.55e-62 & 1.00 & 1.00 \\
& C-3.5-v2 & 25.11 & 17.45 & 20.90 & 20.97 & -2.23 & 0.40 & 5.34e-75 & 7.43e-46 & 3.40e-59 & 1.93e-59 & 1.00 & 1.00 \\
\midrule
\multirow{15}{*}{\rotatebox[origin=c]{90}{OQA}} 
& O-2-7B-B & 8.75 & 11.18 & 13.23 & 15.47 & -0.46 & 5.71 & 5.85e-15 & 4.04e-24 & 6.33e-33 & 1.89e-43 & 1.00 & 5.03e-06 \\
& O-2-7B-S & 13.85 & 11.11 & 12.47 & 12.70 & 15.24 & 12.01 & 9.37e-36 & 7.87e-24 & 1.54e-29 & 1.52e-30 & 2.38e-42 & 1.51e-27 \\
& O-2-7B-D & 15.60 & 10.48 & 7.80 & 13.52 & 17.40 & 12.41 & 4.52e-44 & 2.50e-21 & 7.52e-12 & 3.16e-34 & 4.37e-53 & 3.09e-29 \\
& O-2-7B-I & 28.64 & 20.04 & 15.73 & 18.59 & 23.56 & 14.21 & 3.07e-113 & 2.64e-65 & 9.58e-45 & 4.99e-55 & 1.68e-81 & 5.79e-36 \\
& L-3-70B & 17.00 & 24.45 & 21.35 & 32.48 & 5.46 & 6.98 & 4.83e-51 & 1.37e-90 & 7.73e-74 & 1.32e-133 & 2.00e-05 & 2.16e-09 \\
& L-3.1-70B & 36.79 & 24.62 & 27.93 & 13.35 & 4.98 & 4.88 & 9.31e-156 & 1.63e-91 & 2.17e-109 & 2.21e-19 & 2.42e-04 & 4.03e-04 \\
& L-3.2-1B & 6.98 & 20.50 & 0.00 & 13.20 & -- & 0.12 & 2.17e-09 & 1.05e-66 & 1.00 & 1.52e-27 & -- & 1.00 \\
& L-3.2-11B & 9.69 & 14.83 & 13.04 & 14.61 & 9.04 & 1.26 & 2.66e-18 & 2.20e-40 & 4.33e-32 & 2.82e-39 & 3.13e-12 & 1.00 \\
& L-3.2-90B & 23.61 & 20.52 & 21.04 & 26.79 & 2.05 & 4.21 & 5.28e-86 & 2.23e-69 & 3.85e-72 & 2.91e-103 & 1.00 & 8.59e-03 \\
& Q-2.5-72B & 23.40 & 25.33 & 34.58 & 29.76 & -0.93 & -1.09 & 7.32e-85 & 2.52e-95 & 1.59e-144 & 3.04e-119 & 1.00 & 1.00 \\
& M-small & 25.22 & 25.02 & 24.07 & 28.47 & 1.40 & 1.91 & 9.29e-95 & 1.20e-93 & 1.74e-88 & 2.50e-112 & 1.00 & 1.00 \\
& M-large & 19.53 & 23.38 & 16.79 & 31.89 & 1.78 & 1.19 & 3.66e-64 & 9.49e-85 & 5.54e-50 & 8.59e-129 & 1.00 & 1.00 \\
& C-3 & 20.60 & 20.89 & 36.32 & 25.30 & -1.69 & -1.16 & 8.62e-70 & 5.03e-71 & 2.16e-153 & 8.09e-95 & 1.00 & 1.00 \\
& C-3.5-v1 & 22.84 & 26.23 & 26.93 & 32.96 & -0.88 & -0.48 & 7.12e-82 & 3.15e-100 & 5.45e-104 & 3.96e-136 & 1.00 & 1.00 \\
& C-3.5-v2 & 23.75 & 21.06 & 23.16 & 28.23 & -0.12 & -0.76 & 8.61e-87 & 2.77e-72 & 1.33e-83 & 5.29e-111 & 1.00 & 1.00 \\
\midrule
\multirow{15}{*}{\rotatebox[origin=c]{90}{WVS}} 
& O-2-7B-B & 7.51 & 3.25 & 10.70 & 3.59 & -5.39 & 1.06 & 5.90e-11 & 3.50e-01 & 3.11e-22 & 1.04e-01 & 2.89e-05 & 1.00 \\
& O-2-7B-S & 8.88 & 7.02 & 9.70 & 7.42 & 0.05 & 3.82 & 2.05e-15 & 1.67e-09 & 2.16e-18 & 1.12e-10 & 1.00 & 4.35e-02 \\
& O-2-7B-D & 14.79 & 9.23 & 8.49 & 10.15 & -5.32 & 2.01 & 2.75e-40 & 1.32e-16 & 4.16e-14 & 6.37e-20 & 5.68e-05 & 1.00 \\
& O-2-7B-I & 6.14 & 16.59 & 5.36 & 12.43 & -0.13 & 4.43 & 4.12e-07 & 1.05e-45 & 3.42e-05 & 1.02e-26 & 1.00 & 3.39e-03 \\
& L-3-70B & 20.86 & 14.21 & 21.30 & 15.21 & 2.34 & 2.89 & 1.68e-71 & 2.45e-37 & 6.87e-74 & 5.51e-42 & 1.00 & 1.00 \\
& L-3.1-70B & 22.28 & 11.97 & 17.83 & 14.92 & 12.25 & 6.48 & 3.50e-79 & 9.47e-25 & 1.67e-55 & 1.55e-35 & 1.23e-28 & 1.53e-07 \\
& L-3.2-1B & 0.00 & 8.56 & 0.00 & 5.52 & -- & -- & 1.00 & 1.25e-12 & 1.00 & 6.97e-05 & -- & -- \\
& L-3.2-11B & 9.13 & 3.82 & 9.60 & 2.80 & 2.05 & 2.06 & 2.59e-16 & 4.76e-02 & 5.43e-18 & 1.00 & 1.00 & 1.00 \\
& L-3.2-90B & 16.67 & 16.70 & 18.73 & 18.95 & 13.14 & 9.84 & 1.46e-49 & 2.28e-49 & 3.48e-60 & 7.55e-61 & 1.44e-32 & 1.50e-18 \\
& Q-2.5-72B & 38.70 & 28.16 & 35.38 & 27.84 & 3.54 & 6.38 & 1.28e-166 & 5.87e-110 & 9.59e-150 & 3.15e-108 & 1.27e-01 & 1.18e-07 \\
& M-small & 31.25 & 24.70 & 28.92 & 27.31 & 6.01 & 9.69 & 5.40e-128 & 2.01e-91 & 1.82e-115 & 2.05e-105 & 9.30e-07 & 5.10e-18 \\
& M-large & 22.67 & 19.72 & 29.96 & 24.61 & 6.85 & 4.80 & 2.60e-81 & 8.13e-65 & 4.52e-121 & 6.80e-90 & 5.23e-09 & 5.94e-04 \\
& C-3 & 22.23 & 13.11 & 15.77 & 12.80 & 5.56 & 6.99 & 6.10e-79 & 3.65e-28 & 4.98e-45 & 9.07e-27 & 1.17e-05 & 2.47e-09 \\
& C-3.5-v1 & 28.00 & 6.57 & 25.12 & 8.89 & 5.72 & 2.96 & 1.90e-110 & 4.23e-07 & 9.96e-95 & 2.37e-12 & 4.72e-06 & 1.00 \\
& C-3.5-v2 & 26.18 & 10.74 & 25.44 & 13.76 & 5.56 & 3.24 & 1.62e-100 & 1.64e-20 & 1.82e-96 & 3.11e-31 & 1.15e-05 & 3.96e-01 \\
\bottomrule
\end{tabular}
    \caption{The t-test statistic and corrected p-value (after Bonferroni correction with alpha 0.000172) between the uncalibrated and calibrated pair for each dataset-model-elicitation method setting. These values are for the results in Appendix Table \ref{tab:regression_full} (subset in the main paper Table \ref{tab:regression_all_datasets}). For model families, 'O' stands for 'OLMo', 'L' stands for 'Llama', 'Q' stands for 'Qwen', 'M' stands for 'Mistral', 'C' stands for 'Claude'. For the OLMo models, 'B' stands for 'Base', 'S' stands for 'SFT', 'D' stands for 'DPO', 'I' stands for 'Instruct'.}
    \label{tab:stat_sig_details}
\end{table*}

\section{Calibration hyperparameters}
\label{sec:appendix_hyperparameters}
We perform grid search over the following hyperparameters, tuned on our development set: 
\begin{itemize}
    \item model: \{random forest regression, ridge linear regression, lasso linear regression\}
    \item random forest
    \begin{itemize}
        \item num estimators: \{100, 125, 150, 175, 200, 225, 250\}
        \item max depth: \{1, 2, 3, 4, 5, 6, None\} 
    \end{itemize}
    \item linear models
    \begin{itemize}
        \item alpha: \{0.1, 0.5, 1, 2, 5, 10, 20, 30, 100, 200\}
        \item max iterations: \{50, 100, 200, 300, 1000, 2000\}
    \end{itemize}
\end{itemize}

\section{Prompts}
\label{sec:prompts}

\subsection{Demographics used}
Demographics are shown in Table \ref{tab:example_dataset_questions}.
\begin{table*}[h]
    \footnotesize
    \centering
    \begin{center}
    \begin{tabular}{c p{0.05\linewidth} p{0.27\linewidth} p{0.45\linewidth}}
    \toprule
    Dataset & Qs & Example question & Demographics\\
    \midrule
    \rowcolor{SeaGreen!25}\textbf{WGM} & 14 & \textit{How much do you, personally, know about science? Do you know a lot, some, not much, or nothing at all?} \newline \textit{How much do you trust doctors and nurses in your country?} & \textbf{age} (15-29, 30-49, 50+), \textbf{education}(primary, secondary, tertiary), \textbf{employment status} (employed full time for an employer, employed full time for self, employed part time but do not want full time, unemployed, employed part time want full time, out of work force), \textbf{income quintile} (poorest 20\%, second 20\%, middle 20\%, fourth 20\%, top 20\%), \textbf{living area} (city, rural), \textbf{sex} (male, female), \textbf{world region} (Eastern Africa, Central Africa, North Africa, Southern Africa, Western Africa, Central America and Mexico, Northern America, South America, Central Asia, East Asia, Southeast Asia, South Asia, Middle East, Eastern Europe, Northern Europe, Southern Europe, Western Europe, Aus/NZ)\\
    \rowcolor{Emerald!25}{\textbf{OQA}} & 38 & \textit{How important, if at all, is being a gun owner to your overall identity?} \newline \textit{How much confidence, if any, do you have in elected officials to act in the best interests of the public?} & {\textbf{age} (18-29, 30-49, 50-64, 65+), \textbf{education} (Less than high school, High school graduate, Some college, no degree, Associates degree, College graduate/some postgrad, Postgraduate), \textbf{income} (Less than \$30,000, \$30,000-\$50,000, \$50,000-\$75,000, \$75,000-\$100,000, \$100,000 or more), \textbf{marital status} (Married, Divorced, Separated, Widowed, Never been married), \textbf{political ideology} (Very conservative, Conservative, Moderate, Liberal, Very liberal), \textbf{political party} (Republican, Democrat, Independent, Other), \textbf{race} (White, Black, Asian, Hispanic, Other), \textbf{US region} (Northeast, Midwest, South, West), \textbf{religion} (Protestant, Roman Catholic, Mormon, Orthodox, Jewish, Muslim, Buddhist, Hindu, Atheist, Agnostic, Other, Nothing in particular), \textbf{sex} (Male, Female)}\\
    \rowcolor{PineGreen!25}{\textbf{WVS}} & 40 & \textit{Do you agree strongly, agree, disagree, or disagree strongly with the following statement? One of my main goals in life has been to make my parents proud.} \newline \textit{Do you agree, disagree or neither agree nor disagree with the following statement? When jobs are scarce, employers should give priority to people of this country over immigrants.} & {\textbf{age} (16-24 years, 25-34 years, 35-44 years, 45-54 years, 55-64 years, 65+ years), \textbf{education} (early childhood, primary, lower secondary, upper secondary, post-secondary, short-cycle tertiary, bachelor or equivalent, master or equivalent, doctoral or equivalent), \textbf{employment status} (full time, part time, self-employed, retired/pensioned, housewife not otherwise employed, student, unemployed), \textbf{household size} (1, 2, 3, 4, 5, 6, 7 persons or more), \textbf{immigrant status} (I am born in this country, I am an immigrant to this country), \textbf{income} (low, middle, high), \textbf{marital status} (Married, Living together as married, Divorced, Separated, Widowed, Single), \textbf{religion} (none, Roman Catholic, Protestant, Orthodox (Russian/Greek/etc.), Jew, Muslim, Hindu, Buddhist, Other Christian), \textbf{sex} (Male, Female)}\\
    \bottomrule
    \end{tabular}
    \end{center}
    \caption{The number of questions (Qs) and two example questions from each dataset, along with the demographics we use for each dataset. We do not use all the questions or demographics available due to cost for generations for each prompt, model, and elicitation method (270 settings, see Section \ref{sec:datasets}).}
    \label{tab:example_dataset_questions}
\end{table*}

\subsection{Base prompts}
Base prompts are shown in Table \ref{tab:prompts_base}.

\begin{table*}[]
    \centering
    \small
    \begin{tabular}{l p{0.8\linewidth}}
    \toprule
    Prompt type & Prompt template \\
    \midrule
    Verbalized & \makecell[l]{What percentage of people would respond with \{answer choices\} to the following question? \\
    Question: `\{q\}' \\
    Respond with a list of \{l\} numbers which represent the percentages for \{answer choices\} \\in that order.\\
    Expected response format: [, , , ,]\\
    Response:}\\
    
    Log probabilities & \makecell[l]{Would you respond with \{answer choices\} to the following question? \\
Question: `\{q\}' \\
Respond with only a single integer number in the inclusive range [1,\{l\}], representing \\the index of your selected answer choice in \{answer choices\}. \\
Response: \{answer\}}\\
    
    Self-random & \makecell[l]{Would you respond with \{answer choices\} to the following question? \\
Question: `\{q\}' \\
Respond with only a single integer number in the inclusive range [1,\{l\}], representing \\the index of your selected answer choice in \{answer choices\}. \\
Response:}\\
    
    Paraphrase 1 & \makecell[l]{Would you respond with \{answer choices\} to the following question? \\
Question: `\{q\}' \\
Respond with only a single integer number in the inclusive range [1,\{l\}], representing \\the index of your selected answer choice in \{answer choices\}. \\
Response:}\\
    Paraphrase 2 & \makecell[l]{How might someone respond to the following question? \\
Question: `\{q\}' \\
Possible responses are \{answer choices\}.\\
Respond with only a single integer number in the inclusive range [1,\{l\}], representing \\the index of your selected answer choice in \{answer choices\}. \\
Response:}\\
    Paraphrase 3 & \makecell[l]{
Question: `\{q\}' \\
Response options: \{answer choices\}.\\
Respond with only a single integer number in the inclusive range [1,\{l\}], representing \\the index of your selected answer choice in \{answer choices\}. \\
Response:}\\
    Paraphrase 4 & \makecell[l]{ Below you will be asked to answer some questions.\\
Question: `\{q\}' \\
Response options: \{answer choices\}.\\
Respond with only a single integer number in the inclusive range [1,\{l\}], representing \\the index of your selected answer choice in \{answer choices\}. \\
Response: }\\
    Paraphrase 5 & \makecell[l]{ Answer the following question.\\
Question: `\{q\}' \\
Response options: \{answer choices\}.\\
Respond with only a single integer number in the inclusive range [1,\{l\}], representing \\the index of your selected answer choice in \{answer choices\}. \\
Response: }\\
    \bottomrule
    \end{tabular}
    \caption{Prompt templates used for base prompts (without any sociodemographic information). \{q\} is the question from the original dataset with minor modifications to be suitable for LLMs. \{answer choices\} are the likert-scale answers to the questions from the original dataset.}
    \label{tab:prompts_base}
\end{table*}

\subsection{Sociodemographic prompts}
Sociodemographic prompts are shown in Table \ref{tab:prompts_sd}.
\begin{table*}[]
    \centering
    \small
    \begin{tabular}{l p{0.8\linewidth}}
    \toprule
    Prompt type & Prompt template \\
    \midrule
    Verbalized & \makecell[l]{What percentage of people whose \{att\} is \{c\} would respond with \{answer choices\} to \\the following question?\\
Question: '\{q\}'\\
Respond with a list of \{l\} numbers which represent the percentages for \{answer choices\}\\ in that order.\\
Expected response format: [, , , ,]\\
Response:}\\
    
Log probabilities & \makecell[l]{Imagine your \{att\} is \{c\}. Would you respond with \{answer choices\} to the following question?\\
Question: `\{q\}'\\
Respond with only a single integer number in the inclusive range [1,\{l\}], representing\\ the index of your selected answer choice in \{answer choices\}.\\
Response: \{answer\}}\\
    
Self-random & \makecell[l]{Imagine your \{att\} is \{c\}. Would you respond with \{answer choices\} to the following question?\\
Question: `\{q\}'\\
Respond with only a single integer number in the inclusive range [1,\{l\}], representing\\ the index of your selected answer choice in \{answer choices\}.\\
Response:}\\
    
Paraphrase 1 & \makecell[l]{Imagine your \{att\} is \{c\}. Would you respond with \{answer choices\} to the following question?\\
Question: `\{q\}'\\
Respond with only a single integer number in the inclusive range [1,\{l\}], representing\\ the index of your selected answer choice in \{answer choices\}.\\
Response:}\\

Paraphrase 2 & \makecell[l]{How would a person whose \{att\} is \{c\} respond to the following question?\\
Question: `\{q\}'\\
Possible responses are \{answer choices \}\\
Respond with only a single integer number in the inclusive range [1,\{l\}], representing\\ the index of your selected answer choice in \{answer choices\}.\\
Response:}\\

Paraphrase 3 & \makecell[l]{Person: \{c\} \{att\}\\
Question: `\{q\}'\\
Response options: \{answer choices \}\\
Respond with only a single integer number in the inclusive range [1,\{l\}], representing\\ the index of your selected answer choice in \{answer choices\}.\\
Response:}\\

Paraphrase 4 & \makecell[l]{Below you will be asked to provide a short description of your \{att\} and then answer some questions.\\
Description: I am \{c\}\\
Question: `\{q\}'\\
Respond with only a single integer number in the inclusive range [1,\{l\}], representing\\ the index of your selected answer choice in \{answer choices\}.\\
Response:}\\

Paraphrase 5 & \makecell[l]{Answer the following question as if your \{att\} is \{c\}. Would you respond with \{answer choices\}\\ to the following question?\\
Question: `\{q\}'\\
Respond with only a single integer number in the inclusive range [1,\{l\}], representing\\ the index of your selected answer choice in \{answer choices\}.\\
Response:}\\
    \bottomrule
    \end{tabular}
    \caption{Prompt templates used for sociodemographic prompts (with sociodemographic information). \{q\} is the question from the original dataset with minor modifications to be suitable for LLMs. \{answer choices\} are the likert-scale answers to the questions from the original dataset. \{att\} and \{c\} correspond to the demographic attribute and class respectively (e.g., ``age'' and ``15-24 years''). We note that the placement of \{att\} and \{c\} in the prompt might be slightly different/inverted depending on the demographic for correct grammar.}
    \label{tab:prompts_sd}
\end{table*}

\subsection{Prompt variations}
Prompt variations are shown in Table \ref{tab:prompts_variations}.

\begin{table*}[h]
    \centering
    \small
    \begin{tabular}{l p{0.8\linewidth}}
    \toprule
    Prompt variation & Prompt template \\
    \midrule
    Few shot & \makecell[l]{Please complete the task below. Follow the examples given.\\
    Examples: \{5 examples\} \\
    Task: \{Original prompt\}}\\
    \midrule
    CoT & \makecell[l]{\{Original question\}\\
    Explain your reasoning step-by-step before answering.\\
    \{Response instructions\}\\
    Put your final answer at the end of your response between <ANS\_START> and </ANS\_END>. \\
    Use a maximum of 500 words for your reasoning and final answer combined.}\\
    
    \bottomrule
    \end{tabular}
    \caption{Prompt templates for the different prompt variants. This text is combined with the templates for both base prompts (Table \ref{tab:prompts_base}) and SD prompts (Table \ref{tab:prompts_sd}) to get the results shown in Tables \ref{tab:cot_results} and \ref{tab:few_shot_results}. All other results use the standard prompts without variation.}
    \label{tab:prompts_variations}
\end{table*}

\end{document}